%% file: iclr2027_conference.tex
\documentclass{article}

\usepackage{iclr2027_conference,times}
\input{math_commands.tex}

\usepackage{hyperref}
\usepackage{tikz}
\usepackage{adjustbox}
\usetikzlibrary{arrows.meta,positioning,calc,fit}
\usepackage{url}
\usepackage{graphicx}
\usepackage{amssymb}
\usepackage{booktabs}
\usepackage{multirow}
\usepackage{tabularx}
\usepackage{makecell}
\usepackage{subcaption}
\usepackage{float}
\usepackage{placeins}

\usepackage{xcolor}
\usepackage{microtype}
\usepackage{enumitem}
\usepackage{longtable}
\hypersetup{hidelinks}

\definecolor{generation}{HTML}{3B82F6}
\definecolor{imputation}{HTML}{10B981}
\definecolor{prediction}{HTML}{F59E0B}

\newcommand{\ours}{\textsc{OTBench}}

\title{One Table, Many Queries: Benchmarking Generalist Tabular Models\\
Across Generation, Imputation, and Prediction}

\author{%
\begin{tabular*}{\dimexpr\textwidth-2\tabcolsep\relax}[t]{@{}l@{\extracolsep{\fill}\hspace{4pt}}l*{4}{@{\hspace{4pt}}l}@{}}
Yuanrui Wang\textsuperscript{1,$*$} &
Kaifang Li\textsuperscript{1,3,$*$} &
Xuanyue Li\textsuperscript{1,4} &
Ziyun Li\textsuperscript{1} &
Yan Lu\textsuperscript{1,2} &
Xingxuan Zhang\textsuperscript{1,2,$\dagger$} \\
Peng Cui\textsuperscript{1,2,$\dagger$} & & & & &
\end{tabular*} \\
\textsuperscript{1}StableAI,
\textsuperscript{2}Tsinghua University,
\textsuperscript{3}Beijing University of Posts and Telecommunications\\
\textsuperscript{4}Southeast University \\
\texttt{wangyuanrui@stable-ai.cn}
}

\iclrfinalcopy

\begin{document}

\maketitle
\pagestyle{plain}
{\renewcommand{\thefootnote}{\fnsymbol{footnote}}%
\footnotetext[1]{Equal contribution.}%
\footnotetext[2]{Corresponding authors.}}

\begin{abstract}
Generation, imputation, and prediction query different aspects of the same
tabular distribution, but are usually evaluated separately. This leaves
unclear which capabilities they share, how their training objectives
interact, and whether one model can perform all three well. We introduce the
One Table Benchmark (\ours{}), spanning 56 datasets, 46 registered methods,
and 20 capability axes under aligned data splits and explicit task
contracts. We compare task ranks on aligned support, train five backbones under all
seven non-empty objective combinations on five datasets with three seeds,
and contrast fixed-model profiles with specialist portfolios. Rank comparisons connect imputation and
prediction through numerical recovery and dependence preservation, and link
generation to both through predictive utility; distributional fidelity
follows different, sometimes opposing, rankings. Objective interventions
show the largest pooled gains from including the recipient's own objective. Generation has
the broadest coverage of positive pooled auxiliary effects, while prediction shows
potential to improve imputation and synthetic-data utility, with responses
that vary across backbones. Cross-addition effects reveal that objective
benefits need not add and depend on the objectives already present.
Fixed-model comparisons expose corresponding utility--fidelity trade-offs:
some generalists surpass specialist portfolios on utility, while fidelity
continues to favor specialization. Frozen foundation models show strong
prediction and numerical point imputation, together with promising synthetic-data
utility, highlighting their potential across all three tasks. These findings
motivate unified tabular models that share useful conditional information
while improving categorical reconstruction and faithful joint sampling.
Code is available at
\href{https://github.com/limix-ldm-ai/OTBench}{%
\textcolor{blue}{\nolinkurl{https://github.com/limix-ldm-ai/OTBench}}}.
\end{abstract}

\input{sections/01_introduction}
\input{sections/02_preliminaries}

\input{sections/03_evaluation_metrics}

\input{sections/04_task_rank_analysis}
\input{sections/05_causal_transfer}
\FloatBarrier
\input{sections/06_method_landscape}
\FloatBarrier
\input{sections/07_conclusion}
\input{sections/08_limitations_and_future_work}

\clearpage
\input{sections/statements}

\clearpage
\bibliography{iclr2027_conference}
\bibliographystyle{iclr2027_conference}

\clearpage
\appendix
\input{sections/appendix}

\end{document}

%% file: math_commands.tex
\usepackage{amsmath,amsfonts,bm}

\def\eqref#1{equation~\ref{#1}}

\def\1{\bm{1}}

\DeclareMathAlphabet{\mathsfit}{\encodingdefault}{\sfdefault}{m}{sl}
\SetMathAlphabet{\mathsfit}{bold}{\encodingdefault}{\sfdefault}{bx}{n}



%% file: sections/01_introduction.tex
\section{Introduction}
\label{sec:introduction}

Generation, imputation, and prediction use different aspects of the same
tabular distribution. Generation produces complete records, imputation
recovers missing features from observed information, and prediction infers
a designated target. Each task has its own benchmarks and modeling
traditions~\citep{structuredeval,shadesnull,tabarena}, yet their common
statistical foundation suggests opportunities for sharing learned
capabilities. Understanding these relationships matters for building models
that can serve several uses of a table, and for deciding when a task should
benefit from training on another.

The conditional-query view makes this connection explicit: generation
queries the joint distribution, while imputation and prediction query
feature and target conditionals. Models increasingly support several of
these interfaces~\citep{vaeac,forestdiffusion,limix}, but existing task-wise
evaluations provide limited insight into which capabilities are shared.
Appendix~\ref{app:related-work} reviews related benchmarks and generalist
models.

Similar rankings may reflect useful common information without showing
whether one training objective helps another. Conversely, objectives that
help separately may provide overlapping or competing benefits when combined.
We therefore study three questions: which capabilities co-occur across tasks,
how objective additions and combinations affect a recipient task, and
whether one fixed model can offer a competitive three-task profile.

\begin{figure}[t]
    \centering
    \includegraphics[width=\textwidth]{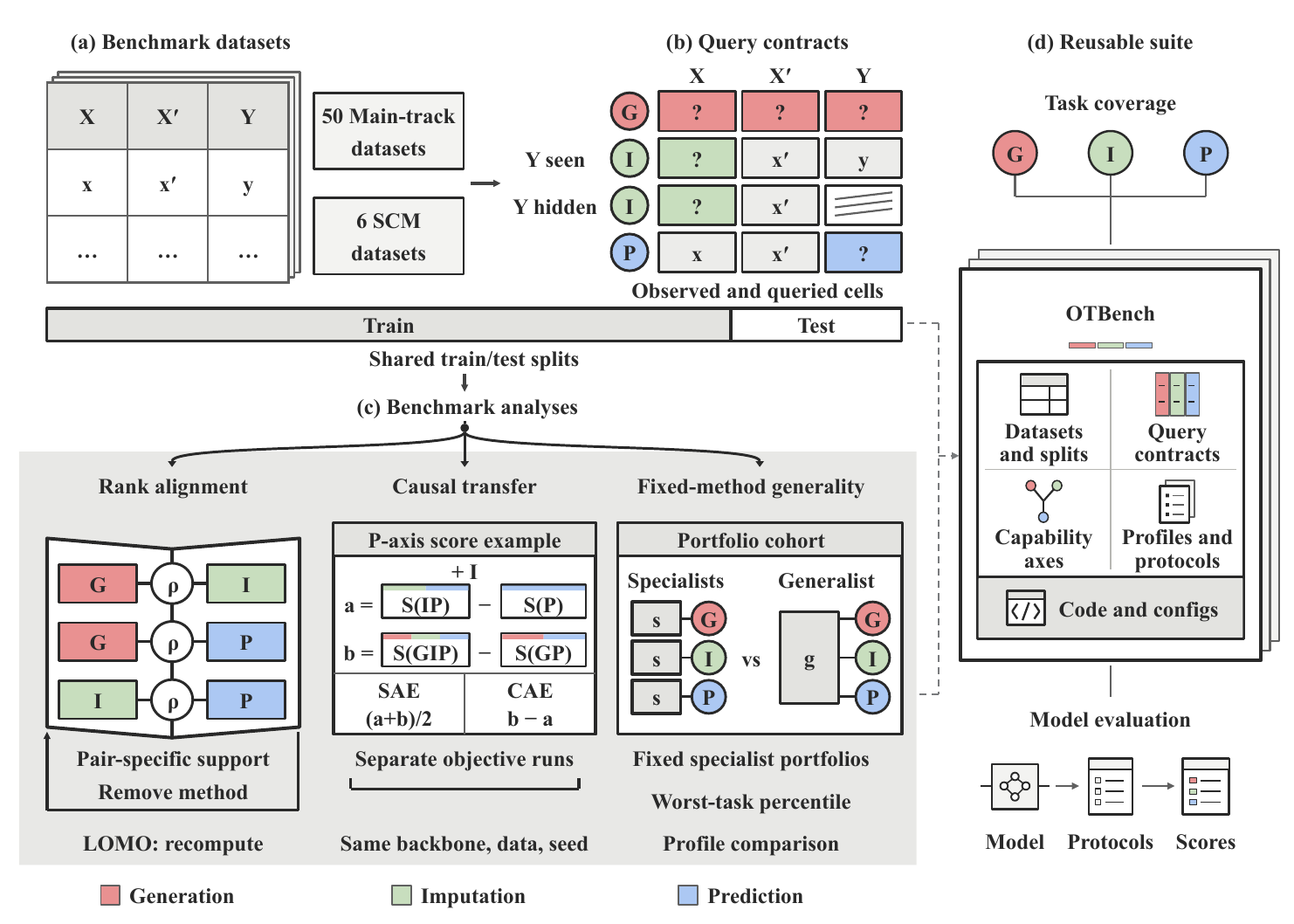}
    \caption{\textbf{OTBench overview.}
    (a) Benchmark datasets and fixed train--test splits shared across tasks.
    (b) Query contracts specify the observed and queried cells for generation,
    imputation, and prediction.
    (c) Complementary analyses assess rank alignment, causal transfer under
    matched objective interventions, and fixed-method generality.
    Imputation is target-observed for rank alignment (MCAR, 30\% missingness)
    and fixed-method generality (averaged over 12 settings), and target-hidden
    for causal transfer.
    (d) The reusable suite bundles benchmark definitions, evaluation protocols,
    code, and configurations.}
    \label{fig:otbench-overview}
\end{figure}

We introduce the One Table Benchmark (\ours{}), a unified platform for these
comparisons. It covers 56 datasets, 46 registered methods, and 20 semantic
capability axes. A common outer
train--test split aligns the data populations across tasks, while explicit
information-access and missingness contracts preserve each task's meaning.
Metric analysis separates point accuracy, distributional fidelity,
dependence preservation, predictive quality, and downstream utility, so that
cross-task relationships can be located at the level of capabilities.
Figure~\ref{fig:otbench-overview} summarizes the benchmark workflow.

We first compare task ranks on aligned method--dataset support
(Section~\ref{sec:task_rank_analysis}). Imputation and prediction agree most
consistently through numerical recovery, dependence preservation, and
predictive utility. Generation connects to both through the predictive
utility of synthetic data, whereas distributional fidelity has different,
sometimes opposing, rank relationships. This suggests that the information
needed for useful conditional queries can be shared even when broader
distributional criteria favor different methods.

To examine objective contributions under controlled changes, we train five
backbones under all seven non-empty combinations of generation, imputation,
and prediction on five datasets with three seeds and declared inference
policies (Section~\ref{sec:transfer}). Single-addition effects (SAE) measure the gain
from enabling one objective; cross-addition effects (CAE) measure whether
two objectives contribute additively. Recipient self-inclusion yields
the largest pooled gains. Generation has broadly positive pooled auxiliary
effects, while prediction shows potential to improve imputation and
synthetic-data utility. These pooled auxiliary responses are descriptive and backbone-dependent.
CAE reveals sub-additivity and context dependence on the benchmark rank scale: useful individual
objectives need not yield additive joint gains.

Finally, fixed-model profiles and comparisons with declared specialist
portfolios connect these relationships to practical model choice
(Section~\ref{sec:methods}). Fidelity and utility favor different profiles;
no fixed identity leads throughout. Recent tabular foundation
models~\citep{limix2,tabiclv2,tabpfn3} achieve strong prediction and numerical
point imputation without single-dataset fine-tuning, and their generation
is competitive on downstream utility. These results reveal potential for
unified foundation models, while motivating better categorical
reconstruction and faithful joint sampling.

\begin{samepage}
Our contributions are:
\begin{itemize}[leftmargin=*,nosep]
    \item A conditional-query benchmark that aligns datasets and capability
    axes across generation, imputation, and prediction while preserving
    task-specific information and evaluation contracts;
    \item Complementary rank comparisons, matched objective interventions,
    and fixed-model profile experiments that distinguish shared
    capabilities, objective interactions, and practical generality;
    \item Empirical findings on utility-centered task relationships,
    non-additive objective effects, and the potential of foundation models
    to support all three tasks.
\end{itemize}
\end{samepage}

%% file: sections/02_preliminaries.tex
\section{Preliminaries}
\label{sec:prelimilary_queries}

\input{sections/02_1_definition.tex}

\input{sections/02_2_dataset_construction.tex}

%% file: sections/02_1_definition.tex
\subsection{Definition}
\label{sec:prelimilary_definition}

\begin{table}[h]
    \centering
    \footnotesize
    \setlength{\tabcolsep}{4.5pt}
    \renewcommand{\arraystretch}{1.05}
    \caption{The three tabular tasks expressed as probabilistic queries.}
    \label{tab:unified-task-definition}
    \begin{tabular}{@{}lccc@{}}
        \toprule
        Task & Observed \(O\) & Queried \(Q\) & Target distribution \\
        \midrule
        Generation
        & \(\varnothing\)
        & \(\mathcal V\)
        & \(\mathbb P^\ast(Z)\) \\
        Imputation
        & \((\mathcal I\setminus M)\cup T_C\)
        & \(M\)
        & \(\mathbb P^\ast(Z_M\mid Z_O;C)\) \\
        Prediction
        & \(\mathcal I\)
        & \(\{p+1\}\)
        & \(\mathbb P^\ast(Y\mid X;C)\) \\
        \bottomrule
    \end{tabular}
\end{table}

To compare methods within a common probabilistic framework, we formulate generation, imputation, and prediction as queries to the same joint distribution over a table. Let a complete row be \(Z=(X_1,\ldots,X_p,Y)\in\mathcal Z:=\bigl(\prod_{j=1}^{p}\mathcal X_j\bigr)\times\mathcal Y\), where \(X=(X_1,\ldots,X_p)\) denotes the features, \(Y\) denotes the supervised target, and \(\mathcal Z\) is the Cartesian product of their domains. Each variable may be continuous or discrete. We denote the unknown joint distribution of \(Z\) by \(\mathbb P^\ast\).
Let \(\mathcal I=\{1,\ldots,p\}\) and \(\mathcal V=\mathcal I\cup\{p+1\}\), with \(Z_{p+1}=Y\). We represent each task as \(q=(O,Q,C)\), where \(O,Q\subseteq\mathcal V\), \(O\cap Q=\varnothing\), and \(C\) specifies the task protocol. Given fitted parameters \(\hat{\theta}\), the model approximates \(\mathbb P^\ast(Z_Q\mid Z_O;C)\) by \(\widehat{\mathbb P}_{\hat{\theta}}(Z_Q\mid Z_O;C)\). Table~\ref{tab:unified-task-definition} summarizes the observed variables, queried variables, and target distribution for each task.
For imputation, \(M\subseteq\mathcal I\) denotes the masked feature set, with \(T_C=\{p+1\}\) when the target is visible under protocol \(C\) and \(T_C=\varnothing\) otherwise.

%% file: sections/02_2_dataset_construction.tex
\subsection{Dataset Construction}
\label{sec:prelimilary_dataset_construction}

Before evaluating any method, we construct a 56-dataset benchmark from 492
OpenML datasets~\citep{openml} reported across 24 studies.  We apply prespecified feasibility filters,
then use method-independent descriptors for PCA and $k$-means~\citep{jolliffe2002pca,lloyd1982least}.
The nearest nonduplicate candidate to each of 50 centroids forms the OpenML track, while six
TabStruct structural-causal-model datasets~\citep{tabstruct} form a separate structural track.
Dataset selection does not use results from any evaluated method.
All tasks share a fixed train–test split, and model fitting, preprocessing, schema estimation, and model selection use the training split only.
Appendix~\ref{app:dataset-construction} details sources, matching, filtering,
descriptors, clustering, duplicate suppression, aggregate benchmark
composition, the fixed split, and the separate SCM track.

%% file: sections/03_evaluation_metrics.tex
\section{Evaluation Metrics}
\label{sec:metrics}

\subsection{Common-support rank protocol}
\label{sec:common-support-rank-protocol}

Let $x_{s,m,k}$ be the replicate-averaged value of metric $k$ for method $m$ in stratum
$s$.  We orient it using $\phi_k(x)\in\{x,-x,-|x-a_k|\}$ for higher-is-better,
error, or target-proximity metrics.  For collection $c$ with members
$\mathcal{K}_c$, let $\mathcal{M}_{s,c}$ contain methods with observations for every member:
\begin{equation}
 r_{s,m,k}=\operatorname{prank}_{m'\in\mathcal{M}_{s,c}}
 \!\left(\phi_k(x_{s,m,k})\right), \qquad
 S_{s,m,c}=\frac{1}{|\mathcal{K}_c|}
 \sum_{k\in\mathcal{K}_c} r_{s,m,k}.
 \label{eq:metric-composite}
\end{equation}
Ties receive the average percentile rank, and $S_{s,m,c}$ requires all declared members.
A metric pair $c=\{i,j\}$ is used to audit redundancy; a retained axis $c$ defines an axis score.
Both use the same orientation and common support within separate protocol
strata; full details are in Appendix~\ref{app:metric-audit}.
All subsequent analyses keep the same fixed member set $\mathcal K_c$ and the
same member-complete, equal-weight aggregation rule.  Their ranking reference
populations remain analysis-specific: for example, Task Rank Analysis ranks
within a task pair's aligned common-method support, whereas Controlled
Transfer ranks within the seven objective arms.

\subsection{Redundancy and retained axes}

\input{tables/metric_redundancy_summary}

Redundancy occurs in every task but varies in extent: generation retains the richest
local structure, whereas prediction shows the greatest compression
(Table~\ref{tab:metric-redundancy-summary}).  PCA shows stronger global
compression, but a shared latent direction does not imply that two metrics are
substitutable.

The audit therefore informs axis construction without mechanically determining it.
We merge only aligned metrics that share an estimand, preserve distinct or opposing
quantities, and keep the components of diagnostic bundles available for separate
reporting.  This yields 20 semantic
axes; protocol strata remain separate rather than being counted as new
constructs.  Table~\ref{tab:metric-groups} records
their exact membership; Appendix~\ref{app:metric-audit} gives the task-specific
rationales alongside Figures~\ref{fig:metric-correlations-generation}--
\ref{fig:metric-correlations-imputation-prediction}.  Axis scores remain
separate rather than forming a universal score.

Where applicable, our metric definitions follow established constructions.
Marginal-shape errors combine per-column two-sample KS statistics for
numerical variables~\citep{smirnov1948table} with categorical total-variation
distances.  We also use C2ST~\citep{lopezpaz2017revisiting}; $\alpha$-precision, $\beta$-recall, and
authenticity~\citep{alaa2022faithful}; a mixed-type $k$-way query workload
adapted from discrete marginal queries~\citep{mckenna2019graphical}; and a TSTR
gap adapted from the TSTR protocol~\citep{esteban2017realvalued}.  The privacy
axis combines benchmark-defined closest-record diagnostics based on Gower
distance~\citep{gower1971general}, a distance-based membership diagnostic
motivated by membership-inference attacks and low-FPR reporting
~\citep{shokri2017membership,carlini2022membership}, and a legacy
$\delta$-presence-inspired diagnostic~\citep{nergiz2007hiding}; none constitutes
a privacy guarantee.  The imputation distribution axes use Jensen--Shannon
divergence~\citep{lin1991divergence} and the 1-Wasserstein
distance~\citep{villani2009optimal}.  Classification evaluation includes
balanced accuracy and MCC~\citep{brodersen2010balanced,matthews1975comparison},
ROC-AUC and average precision~\citep{hanley1982meaning,davis2006relationship},
proper scores~\citep{brier1950verification,good1952rational}, and calibration
and selective-prediction measures~\citep{naeini2015obtaining,guo2017calibration,geifman2019bias}.

\input{tables/metric_groups}

%% file: tables/metric_redundancy_summary.tex
\begin{table}[H]
\centering
\caption{Metric-space redundancy by task.}
\label{tab:metric-redundancy-summary}
\small
\setlength{\tabcolsep}{8pt}
\renewcommand{\arraystretch}{1.00}
\begin{tabular}{@{}l r r r c@{}}
\toprule
Cohort & Metrics & Clusters & PCs & PCA mass (PC1) \\
\midrule
Generation & 29 & 17 & 5 & 80\% (53\%) \\
Imputation & 10 & 8 & 3 & 74\% (41\%) \\
Prediction (cls.) & 10 & 4 & 2 & 88\% (75\%) \\
Prediction (reg.) & 5 & 2 & 1 & 86\% (86\%) \\
\bottomrule
\end{tabular}
\end{table}

%% file: tables/metric_groups.tex
\begin{table}[H]
\centering
\caption{\textbf{Retained capability axes and exact metric membership.}
Within an axis, direction-aligned metrics are equally averaged after
within-protocol ranking (Equation~\ref{eq:metric-composite}); diagnostic bundle
components remain reportable.}
\label{tab:metric-groups}
\scriptsize
\setlength{\tabcolsep}{3pt}
\renewcommand{\arraystretch}{1.14}
\begin{tabularx}{\textwidth}{@{}l >{\raggedright\arraybackslash}X@{}}
\toprule
Task & Retained axes (member metrics) \\
\midrule
G & \textbf{Fidelity}: \texttt{numerical\_ks\_error},
\texttt{shape\_error}, \texttt{one\_way\_error},
\texttt{two\_way\_error}, \texttt{three\_way\_error};
\textbf{C2ST}: \texttt{gbdt\_c2st\_auroc},
\texttt{logistic\_c2st\_auroc};
\textbf{$\alpha$-precision}: \texttt{alpha\_precision};
\textbf{association}: \texttt{mixed\_association\_error},
\texttt{mixed\_association\_similarity};
\textbf{TSTR gap}: \texttt{tstr\_gap};
\textbf{global-utility compromise}: \texttt{global\_utility\_ratio},
\texttt{beta\_recall\_coverage}, \texttt{authenticity};
\textbf{privacy-risk diagnostics}:
\texttt{mean\_min\_euclidean\_dcr},
\texttt{normalized\_median\_gower\_dcr},
\texttt{distance\_mia\_auroc},
\texttt{distance\_mia\_tpr\_at\_1pct\_fpr},
\texttt{delta\_presence}. \\
\addlinespace
I & \textbf{numerical point accuracy}: \texttt{macro\_nmae},
\texttt{macro\_nrmse};
\textbf{categorical point accuracy}:
\texttt{categorical\_balanced\_accuracy},
\texttt{categorical\_macro\_f1};
\textbf{categorical distribution}: \texttt{categorical\_jsd};
\textbf{numerical distribution}: \texttt{numeric\_wasserstein1};
\textbf{association}: \texttt{association\_matrix\_error};
\textbf{downstream utility gap}: \texttt{utility\_gap}. \\
\addlinespace
P (cls.) & \textbf{hard-decision performance}: \texttt{accuracy},
\texttt{balanced\_accuracy}, \texttt{macro\_f1}, \texttt{mcc};
\textbf{proper scores}: \texttt{brier\_score}, \texttt{log\_loss};
\textbf{ranking/selective risk}: \texttt{roc\_auc},
\texttt{average\_precision}, \texttt{aurc};
\textbf{calibration}: \texttt{ece}. \\
\addlinespace
P (reg.) & \textbf{squared-error family}: \texttt{r2},
\texttt{nrmse}, \texttt{rmse};
\textbf{mean absolute error}: \texttt{mae};
\textbf{median error}: \texttt{median\_absolute\_error}. \\
\bottomrule
\end{tabularx}
\end{table}

%% file: sections/04_task_rank_analysis.tex
\section{Task Rank Analysis}
\label{sec:task_rank_analysis}

\subsection{Objective and Method}
\label{sec:tra_objective_and_method}

To characterize cross-task alignment among existing methods, we compare their
relative rankings across generation (G), imputation (I), and prediction (P).
Consistent rankings would indicate that the evaluated capabilities covary
across tasks, whereas axis-dependent or conflicting rankings would show that
cross-task performance cannot be adequately summarized by a single score.

\paragraph{Training protocol.}
Cross-task rank analysis requires a method to support multiple tasks. We
therefore select models with benchmark interfaces for at least two of G, I,
and P; Table~\ref{tab:app-method-capabilities} lists the methods and their
supported tasks. Each model follows its default training or adaptation
protocol, including its prescribed pretrained or in-context route where
applicable. Dataset-specific fitting and adaptation use the training split
of each \ours{} dataset, and evaluation uses its held-out test split.
The same outer split is shared across tasks, so task comparisons do not
change the underlying training and test populations.
Appendix~\ref{app:baseline-implementation} gives the model configurations.

For each task pair \(a\) and \(b\), we identify the common method pool
\(\mathcal M_{ab}\), align metric directions, average scores over seeds or
replicates, and analyze classification and regression separately.
Imputation uses the target-observed, missing-completely-at-random (MCAR)
setting~\citep{rubin1976inference} with a missing rate of \(0.3\).

Capability-axis scores follow the common-support rank protocol in
Section~\ref{sec:common-support-rank-protocol}. For each axis pair \(c,c'\)
and dataset \(d\), we retain methods in \(\mathcal M_{ab}\) for which all
declared members of both axes are valid. We orient the metrics and compute percentile
ranks on this aligned method set, then aggregate the member ranks
according to Equation~\ref{eq:metric-composite}. The ranking
reference set is specific to each task pair and axis pair.

We compute one pooled Spearman correlation~\citep{spearman1904proof} over the
aligned dataset--method units.  
To assess sensitivity to method-pool composition, we perform leave-one-method-out (LOMO) analysis by removing each method in turn and recomputing the aligned support, within-dataset ranks, axis scores, and pooled correlation.
An axis pair is
\emph{LOMO-direction-preserving} only when every effective deletion preserves
the sign of the full-pool correlation; deletions that change no valid unit are
excluded from this diagnostic.  
Appendix~\ref{app:task-rank-analysis} gives the full construction.

\subsection{Results and Discussion}
\label{sec:tra_results_and_discussion}

\begin{figure}[htbp]
    \centering
    \includegraphics[width=\textwidth]{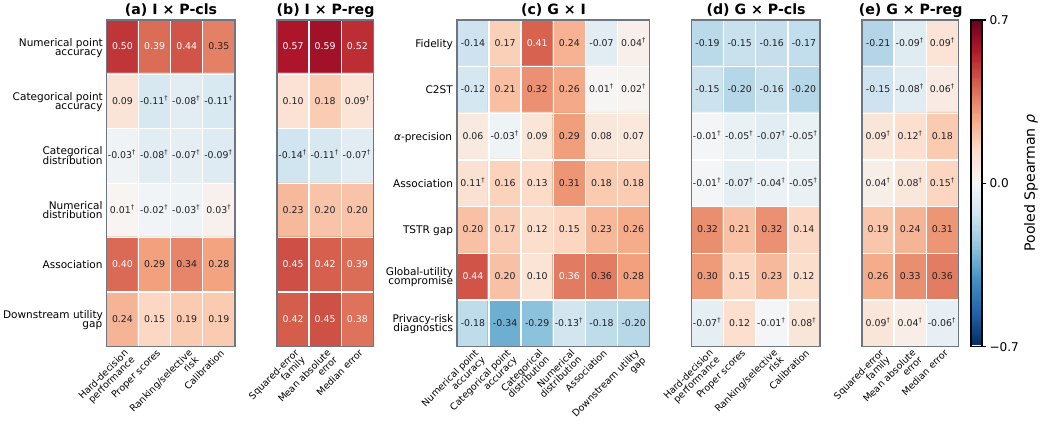}
    \caption{\textbf{Cross-task rank consistency across retained capability axes.}
    Each cell reports the pooled Spearman correlation between member-complete
    capability-axis scores computed over the same aligned dataset--method
    support. Red and blue indicate positive and negative correlations.
    \(^{\dagger}\) marks axis pairs for which at least one effective
    single-method deletion changes the sign of the full-pool correlation or
    produces zero correlation. G, I, P-cls, and P-reg denote generation,
    imputation, classification prediction, and regression prediction.}
    \label{fig:tra_pooled_spearman}
\end{figure}

Cross-task rank agreement depends on the capability axes being compared
(Figure~\ref{fig:tra_pooled_spearman}). No uniform correspondence holds across
all tasks and evaluation dimensions.

\paragraph{Imputation--prediction.}
Imputation and prediction show selective rank agreement, with 13 of the 24 classification axis pairs and 14 of the 18 regression pairs preserving direction under LOMO. 
Across both prediction settings, numerical point accuracy, association, and downstream utility gap have positive, direction-preserving correlations with every prediction axis. 
The distributional axes show a different pattern. 
Numerical distribution maintains positive agreement with all three regression axes, but neither distribution axis has a direction-preserving relationship with classification performance.

Recovering values for individual records and preserving variable relationships can retain information useful for target prediction. 
This is consistent with our intuition that learned statistical dependencies can support both imputation and prediction. 
Matching marginal value distributions emphasizes a different property of the completed data, which helps explain its less uniform correspondence with prediction. 
The connection is more direct for downstream utility gap because it evaluates completion quality through predictive performance. 
Together, these results locate the shared performance of imputation and prediction in numerical recovery, dependence preservation, and predictive utility. 
For unified models, this suggests a focus on learning relationships that support both missing-feature recovery and target prediction, including when the target itself is unavailable as conditioning information.

\paragraph{Generation--imputation.}
Generation and imputation have the highest proportion of LOMO-direction-preserving relationships among the task pairs, with 36 of the 42 axis pairs preserving direction. 
TSTR gap and global-utility compromise show positive, direction-preserving correlations with all six imputation axes. 
Privacy-risk diagnostics show negative, direction-preserving correlations with every imputation axis except numerical distribution.

TSTR gap measures the preservation of target-specific predictive utility relative to predictors trained on real data. 
Its positive rank correlations indicate that generation methods whose synthetic data better preserve this utility also tend to rank higher on imputation. 
This is consistent with our intuition that learned variable dependencies can support both useful data generation and missing-value recovery. 
Fidelity and C2ST assess marginal and low-order statistical similarity and real–synthetic distinguishability, respectively, and their correlations with imputation axes vary in sign. 
Thus, both tasks can use dependencies among table variables, but a method that ranks highly on generation metrics does not necessarily lead on imputation. 
Whether its advantage carries over depends on how effectively the learned relationships support recovery of missing values from observed information.

\paragraph{Generation--prediction.}
For generation and prediction, 17 of the 28 classification axis pairs and 9 of the 21 regression axis pairs preserve direction under LOMO. 
Fidelity and C2ST show negative, direction-preserving correlations with all four classification axes, whereas TSTR gap and global-utility compromise show positive, direction-preserving correlations with all classification and regression axes.

This contrast in correlation signs distinguishes the statistical similarity of synthetic data from their predictive utility. 
Prediction relies on conditional information relevant to a specified target, while generation evaluation also examines other distributional properties whose importance to that target can vary. 
The shared focus of TSTR gap and the prediction axes on target-related information helps explain their positive rank agreement. 
These results connect generation and prediction through the preservation of target-specific predictive utility. 
Future work could explicitly model target-relevant dependencies in generation and assess their effects on predictive utility and distributional fidelity.


%% file: sections/05_causal_transfer.tex
\raggedbottom
\section{From Rank Alignment to Causal Transfer}
\label{sec:transfer}

Section~\ref{sec:task_rank_analysis} characterizes relationships among tasks
through the agreement of method rankings on matched capability axes. This
reveals which capabilities tend to co-occur across methods, but leaves open
how the tasks contribute to one another within a model: architectures,
training protocols, and objective combinations vary together in those
comparisons. To examine these relationships under controlled changes, we ask
whether adding an objective improves another recipient task, and whether
objectives combine additively within a shared model. We call this experiment
\emph{causal transfer}: matched interventions on the enabled objectives
complement the observational evidence from task ranks.

\paragraph{Experimental setup and evaluation.}
We train Gaussian Copula~\citep{sdv}, Forest-VP Joint~\citep{forestdiffusion},
TabDiff~\citep{tabdiff}, TabSyn~\citep{tabsyn}, and TabDDPM~\citep{tabddpm}
under all seven non-empty objective arms
$\mathcal C=\{G,I,P,GI,GP,IP,GIP\}$.
The study uses five datasets---JapaneseVowels, cleveland,
credit\_card\_clients\_default, website\_phishing, and MiamiHousing2016---and
training seeds $\{0,1,2\}$, yielding 525 backbone--dataset--seed--arm systems.
The first four datasets are classification problems and the last is
regression. We use the fixed \ours{} training and test splits, hold the
configured training budget fixed across arms within each backbone, and
sample enabled objectives uniformly during fitting. G scores all table
variables; I reconstructs randomly selected features with the target hidden;
P predicts the target from the features. The neural pipelines reserve an
internal validation split from the training data, and TabSyn shares an
objective-independent VAE across the seven arms of each dataset--seed block.

Each trained system is evaluated on all three tasks. Generation produces as
many synthetic rows as the outer training split; TSTR fits a predictor on
these rows and evaluates it on held-out real data, with a real-trained
predictor as reference. Imputation evaluates held-out feature recovery under
MCAR, MAR, MNAR, and mixed missingness at rates $\{0.1,0.3,0.5\}$, using three
completed-table draws per fixed, method-independent mask and keeping the
target hidden. Prediction evaluates held-out targets, aggregating three
conditional draws for the four diffusion-based backbones; Gaussian Copula
uses its native Monte Carlo query. Comparisons match backbone, dataset,
training seed, and atomic evaluation setting. Within each matched block, we
orient each metric so that higher is better, normalize its seven-arm ranks
to $[0,1]$, and average member ranks within each capability axis. Every member
must be finite for all seven arms. Imputation contrasts are formed separately
for each mechanism--rate setting before equally weighted aggregation;
classification and regression axes remain separate. Intervals use 2,000
crossed backbone--dataset bootstrap draws with seed resampling. Appendix~\ref{app:transfer-implementation}
gives the splits, training configurations, sampling budgets, and unavailable
outcomes; Appendix~\ref{app:causal-transfer} specifies matching and uncertainty.

\subsection{Single-addition effect (SAE)}
\label{sec:causal-design}

Let $S_{A,a}$ denote the normalized score on recipient axis $a$ for objective
arm $A$, suppressing the matched block indices. For source objective $s$,
recipient task $t$, and remaining task $u$, the \emph{single-addition effect}
is
\begin{equation}
\operatorname{SAE}_{s\rightarrow t,a}
=\frac{1}{2}\left[
(S_{st,a}-S_{t,a})+(S_{stu,a}-S_{tu,a})
\right],\qquad \{s,t,u\}=\{G,I,P\}.
\label{eq:transfer}
\end{equation}
It averages the gain from adding $s$ when $u$ is absent and present, with the
recipient objective included on both sides. A positive value indicates an
improvement on the recipient axis. For \emph{self-inclusion}, the added
objective and recipient coincide: we instead average its addition to the
three non-empty arms formed from the other two objectives. No empty training
arm is required. For the diffusion-based backbones, I/P inference uses clean
observed features ($x_0$) when that recipient objective is enabled and the
$x_t$ conditioning path otherwise; Gaussian Copula always uses native
conditioning. Off-diagonal SAE therefore fixes the recipient's inference
rule, whereas I/P self-inclusion compares complete training--inference
systems. Generation inference is fixed throughout.

\begin{figure}[htbp]
    \centering
    \includegraphics[width=\textwidth]{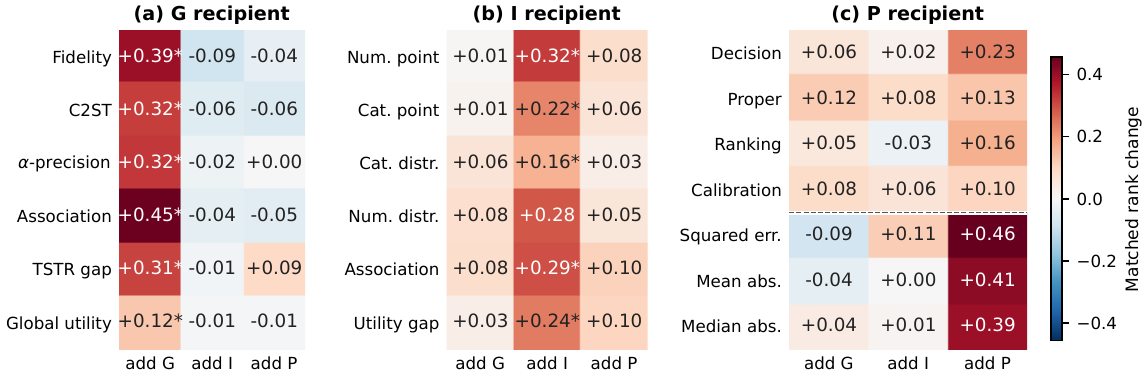}
    \caption{\textbf{Single-addition effects across capability axes.}
    Panels (a)--(c) show generation, imputation, and prediction recipients;
    columns identify the added objective. Values are normalized seven-arm
    rank changes. Self-inclusion cells compare adding the recipient's own
    objective; off-diagonal cells hold its inference rule fixed.
    Stars mark the complete robustness criterion in
    Appendix~\ref{app:causal-transfer}. The privacy-risk diagnostic is omitted
    from this quality-focused display but retained in the full tables.}
    \label{fig:causal-transfer}
\end{figure}

\paragraph{Direct supervision gives the largest gains.}
Self-inclusion is positive on all 19 displayed axes and exceeds both
auxiliary-objective effects on each axis (Figure~\ref{fig:causal-transfer}).
Its effects range from $.122$ to $.445$ for generation, $.159$ to $.317$ for
imputation, and $.098$ to $.456$ for prediction. Thus, explicitly training for
the recipient is the strongest direct intervention in this study. This
supports including the desired task's own objective as a first priority;
it does not imply that a singleton arm always outperforms every joint arm.

\paragraph{Generation provides broadly positive auxiliary effects.}
Beyond self-inclusion, G has the broadest positive coverage: its pooled SAE
is positive on all six imputation axes and five of seven prediction axes.
The exceptions are regression squared error and mean absolute error.
Generation requires fitting values across the joint table distribution,
which may encourage representations of dependencies useful to several
conditional queries. Its broad modeling scope offers a possible explanation
for this pattern, rather than evidence that generation is intrinsically the
hardest task or uniformly beneficial.

Both G and P improve the pooled means on every imputation axis. P has the
larger effect on four of six axes, including numerical point accuracy
($.080$ versus $.007$) and association ($.105$ versus $.082$); G is stronger
on the two distributional axes. One possible explanation concerns the
heterogeneity of reconstruction targets. Some features are weakly
predictable from the observed variables, so their reconstruction losses can
provide limited conditional signal. I trains on smaller queried subsets,
where such features may account for a larger share of the update than in G,
which scores the whole row. P concentrates supervision on a designated
prediction target and may supply a more consistent conditional learning
signal that also helps feature recovery. These are hypotheses about the
observed pattern; we do not measure feature predictability or gradient
alignment, and a supervised target need not be fully predictable.

\begin{figure}[htbp]
    \centering
    \includegraphics[width=\textwidth]{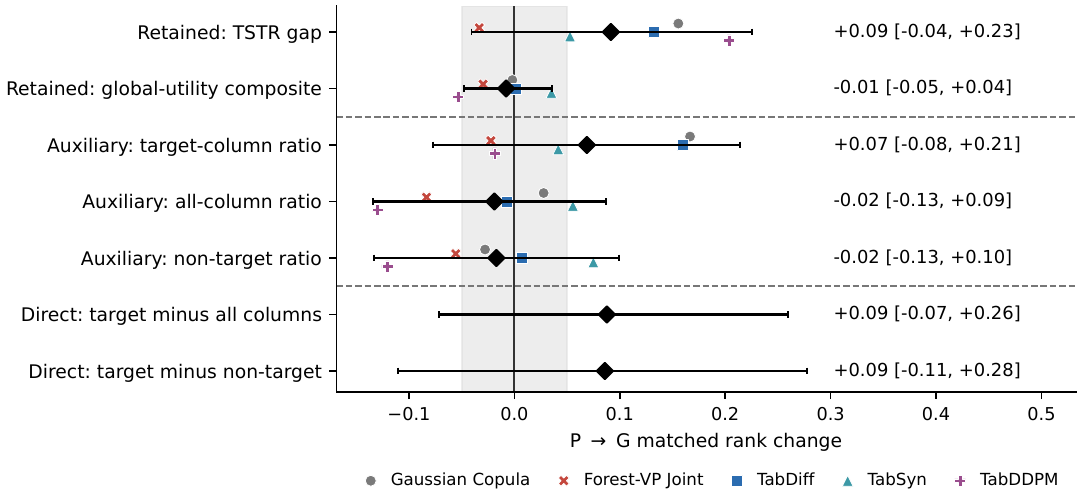}
    \caption{\textbf{Prediction-to-generation utility effects.}
    Diamonds and intervals show pooled SAE and 95\% bootstrap intervals;
    colored markers show backbone means. Retained generation axes and
    auxiliary scope-matched per-column diagnostics use separate
    multiplicity families. Target localization is assessed by direct paired
    contrasts, not by comparing whether separate intervals cross zero.}
    \label{fig:causal-utility-scope}
\end{figure}

\paragraph{Prediction can improve the downstream utility of generated data.}
Adding P to G produces a larger positive response on the normalized
TSTR-gap score ($.092$) than on the global-utility composite ($-.008$), motivating the closer view in
Figure~\ref{fig:causal-utility-scope}. TSTR evaluates a central use of
synthetic data: training predictors that generalize to real records.
A higher TSTR-gap score means a gap closer to the real-trained reference,
not a change of the same magnitude in raw predictive accuracy.
Four of five backbones have positive P$\rightarrow$G TSTR-gap rank effects, with
Forest-VP Joint the exception ($-.033$). The pooled interval remains wide
($[-.041,.225]$). Auxiliary per-column diagnostics also suggest a stronger
response at the designated target than across all columns, although the
paired localization contrasts are inconclusive. Prediction-oriented
supervision is therefore a promising strategy for improving synthetic-data
utility, with benefits that depend on the backbone and evaluation axis.
The off-diagonal patterns here are descriptive: none meets the complete
cross-backbone robustness criterion, which is reported alongside the
uncertainty estimates rather than used to discard their differing responses.

\begin{samepage}
\subsection{Cross-addition effect (CAE)}

To measure whether objectives combine additively, we define the
\emph{cross-addition effect} for added objectives $s,t$ and anchor~$u$:
\begin{equation}
\operatorname{CAE}_{s\times t\mid u,a}
=(S_{stu,a}-S_{su,a})-(S_{tu,a}-S_{u,a}),
\qquad \{s,t,u\}=\{G,I,P\}.
\label{eq:gi-interaction}
\end{equation}
\end{samepage}
Here the recipient is determined by axis $a$ and may differ from the anchor.
A positive CAE means that the score increment from adding $t$ is larger
when $s$ is already present (complementarity); a negative CAE means that its increment is smaller
(sub-additivity). A zero value is consistent with additive effects on this
rank scale. Sub-additivity does not require either added objective to be
harmful on its own.

\begin{figure}[htbp]
    \centering
    \includegraphics[width=\textwidth]{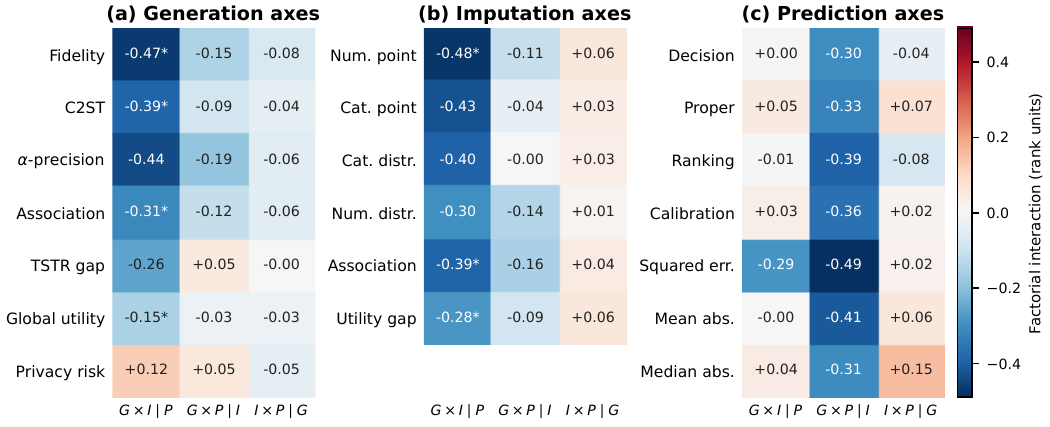}
    \caption{\textbf{Cross-addition effects conditional on an anchor objective.}
    Panels (a)--(c) report generation, imputation, and prediction recipient
    axes. Columns are $G\times I\mid P$, $G\times P\mid I$, and
    $I\times P\mid G$. Positive and negative values indicate complementarity
    and sub-additivity on the normalized seven-arm rank scale. Stars mark
    the complete robustness criterion; blanks denote unavailable estimates.
    I/P interactions that toggle the recipient objective also include its
    conditioning-policy switch.}
    \label{fig:causal-interactions}
\end{figure}

\paragraph{Objective interactions depend on context.}
Figure~\ref{fig:causal-interactions} shows sub-additivity for
$G\times I\mid P$ across all six non-privacy generation axes and all six
imputation axes, and for $G\times P\mid I$ across all seven prediction axes.
Seven G--I effects meet the full robustness criterion. With G as the anchor,
$I\times P\mid G$ instead has positive means on all six imputation and five
of seven prediction axes, consistent with the P$\rightarrow$I SAE pattern;
however, these intervals cross zero. Joint gains therefore depend on the
recipient and objective context. Shared information may yield overlap or
complementarity, while fixed budgets may create update or capacity
competition. These mechanisms remain hypotheses, and I/P interactions that
toggle the recipient also include its conditioning-policy switch.

\subsection{Objective-arm Pareto profiles}
\label{sec:objective-pareto}

\begin{figure}[!htbp]
    \centering
    \includegraphics[width=0.94\textwidth]{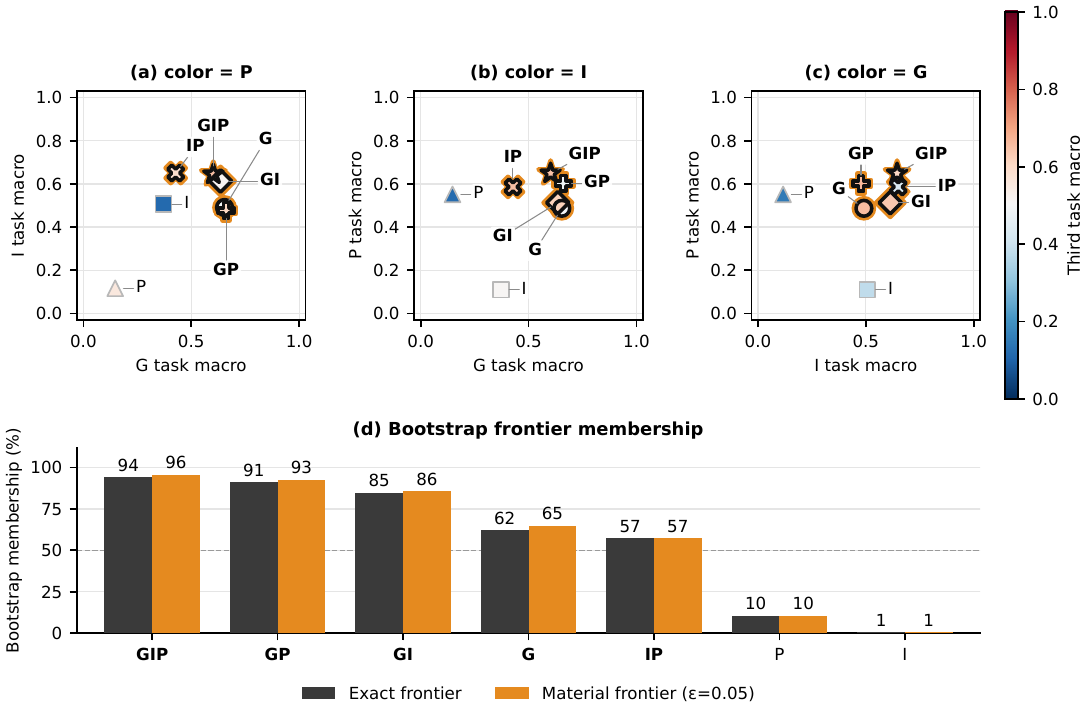}
    \caption{\textbf{Three-task trade-offs among objective arms.}
    Upper panels project one joint $(G,I,P)$ frontier; color shows the third
    coordinate. Black outlines and orange halos mark exact and material
    frontiers. Lower bars show bootstrap membership frequencies. This
    available-coordinate analysis has 75 candidate blocks, with G/I/P
    coordinates on 70/69/75 blocks; it does not impose identical retained
    axes across backbones. Frequencies describe resampling stability.}
    \label{fig:causal-pareto-combined}
\end{figure}

CAE describes interactions on individual capability axes. We next examine
the resulting three-task trade-offs among the seven training arms.
Figure~\ref{fig:causal-pareto-combined} uses available coordinates across
all five backbones: within each block, retained axes and imputation settings
must be scored for all seven arms. Settings are averaged within axes, then
retained non-privacy axes are equally weighted; seeds, datasets within each
backbone, and backbones are averaged successively. Different blocks may
retain different axes.

All three task coordinates are oriented so that higher is better. An arm
\emph{exactly dominates} another if it is no worse on any coordinate and
strictly better on at least one. \emph{Material dominance} requires no loss
on any coordinate and a gain of at least $\epsilon=.05$ on at least one.
The \emph{exact frontier} and \emph{material frontier} consist of arms not
dominated by any other candidate under the respective rule. Both are
computed jointly in $(G,I,P)$, rather than separately in pairwise
projections.

The exact and material frontiers both contain $\{G,GI,GP,IP,GIP\}$.
In the pooled estimates, GP has the strongest generation coordinate, IP the
strongest imputation coordinate, and GIP the strongest prediction coordinate.
GIP combines high I/P scores with a lower G coordinate than GP, and
has $94\%$ exact-frontier bootstrap membership. This is a competitive
compromise, rather than universal superiority. Support also matters: the
complete-block analysis (47 blocks) retains only $\{GI,GP,GIP\}$
(Appendix~\ref{app:causal-transfer}). These objective-arm trade-offs lead to
the next section's comparison of fixed, deployable method identities with
specialist portfolios.

\flushbottom

%% file: sections/06_method_landscape.tex
\section{Fixed-Method Generality}
\label{sec:methods}
\raggedbottom

We now compare fixed model identities across tasks and ask when one
generalist is competitive with a specialist portfolio. Each method retains
the same identity across datasets, capability axes, and missingness
settings. Comparisons preserve the task contracts and percentile composites
in Equation~\ref{eq:metric-composite}; no method switches identities to
assemble an oracle.

\begin{figure}[htbp]
    \centering
    \includegraphics[width=\textwidth]
        {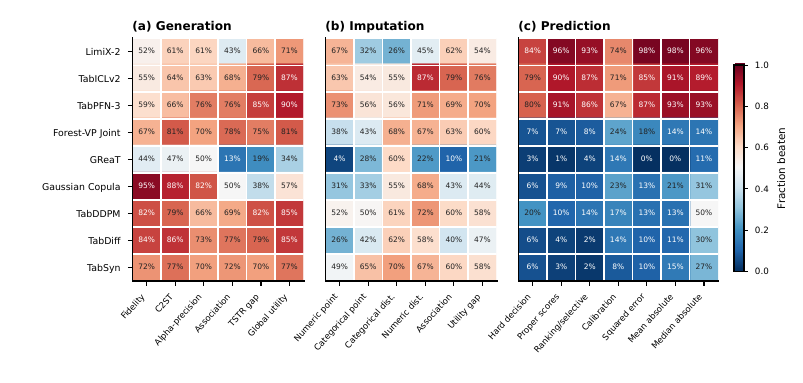}
    \caption{\textbf{A generalist's advantage is capability-specific.}
    Cells are mean fractions of fixed specialists beaten on each axis.
    Panels show (a) generation, (b) imputation averaged over twelve separately
    evaluated settings, and (c) prediction. Columns are not pooled into a
    leaderboard.}
    \label{fig:position}
\end{figure}

\paragraph{Performance rankings depend on the capability axis.}
Figure~\ref{fig:position} reports the fraction of contract-compatible fixed
specialists beaten by each generalist, averaged over paired datasets with
half credit for ties. Imputation comparisons are computed separately in each
of the twelve target-observed mechanism--rate settings before averaging.
These fractions describe relative position, not raw accuracy.

Generation shows a clear separation between distributional fidelity and
predictive utility: Gaussian Copula leads fidelity, C2ST, and
$\alpha$-precision, Forest-VP Joint leads association, and TabPFN-3 leads the
TSTR-gap axis and global-utility composite. Imputation is similarly divided:
TabPFN-3 leads numerical point accuracy, TabICLv2 leads numerical
distribution, association, and utility preservation, and TabSyn leads both
categorical axes. LimiX-2 leads every displayed prediction axis.
Thus, competence on one criterion does not imply a matching advantage on
other criteria, even within the same task.

\begin{figure}[htbp]
    \centering
    \includegraphics[width=\textwidth]
        {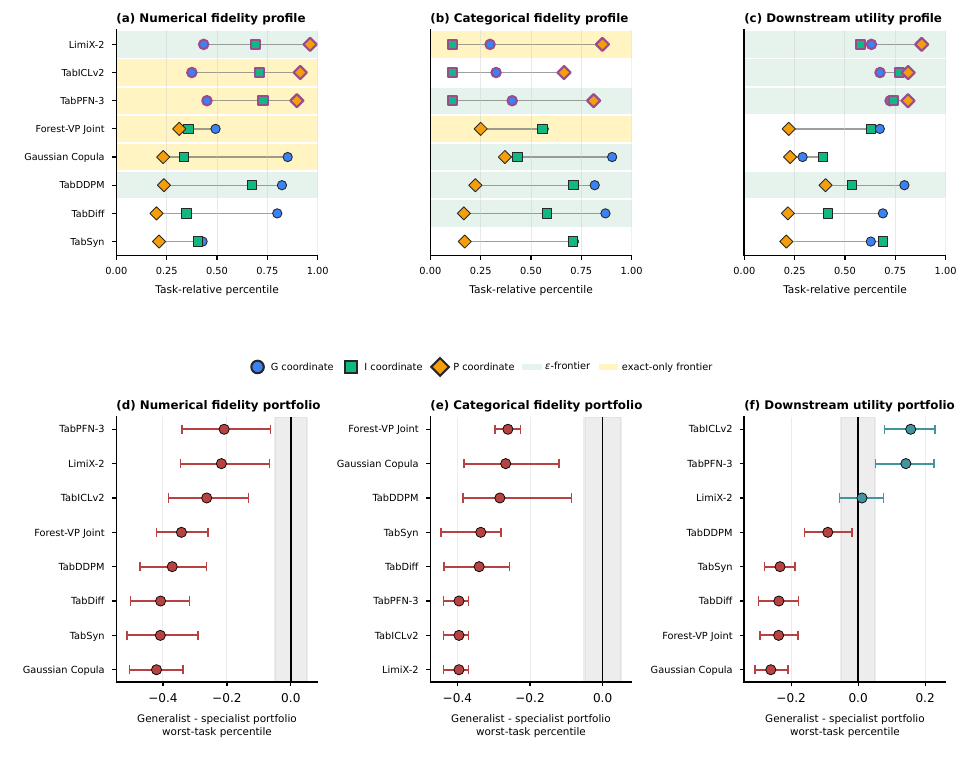}
    \caption{\textbf{Fixed identities show profile-dependent deployment
    trade-offs.} Panels (a)--(c) connect G, I, and P percentiles for numerical
    fidelity, categorical fidelity, and utility on 12, 3, and 26 common
    datasets. Green and yellow rows indicate practical and exact-only
    frontiers. Panels (d)--(f) show generalist-minus-fixed-portfolio gaps in
    worst-task percentile on 11, 3, and 26 common datasets, with bootstrap
    intervals and a gray $\pm.05$ band. Positive gaps favor the generalist.}
    \label{fig:fixed-identity-deployment}
\end{figure}

\paragraph{Joint profiles expose different bottlenecks and trade-offs.}
Figure~\ref{fig:fixed-identity-deployment}(a)--(c) connects each method's
G, I, and P coordinates under three prespecified profiles. Numerical fidelity
combines generation fidelity, numerical imputation point accuracy, and
regression squared error; categorical fidelity substitutes categorical
imputation point accuracy and classification hard decisions; downstream
utility combines generation TSTR, imputation utility preservation, and
classification hard decisions. Each profile uses its own common dataset
cohort. Its percentile-composite coordinates differ from the
specialist-beating fractions in Figure~\ref{fig:position}.

High prediction and numerical imputation can coexist with weaker generation
fidelity, while several generative methods show the opposite imbalance.
Categorical reconstruction introduces another bottleneck, and the utility
profile produces a different, often more balanced ordering. These contrasts
show why a single cross-task average would obscure useful distinctions.
The practical Pareto frontier ($\epsilon=.05$) contains LimiX-2 and TabDDPM
for numerical fidelity; Gaussian Copula, TabDDPM, TabDiff, and TabPFN-3 for
categorical fidelity; and LimiX-2, TabDDPM, TabICLv2, and TabPFN-3 for utility.
No fixed identity is uniformly best across the three profiles.

\paragraph{Specialist portfolios retain a fidelity advantage.}
Figure~\ref{fig:fixed-identity-deployment}(d)--(f) compares each generalist
with one declared specialist triple per profile
(Appendix~\ref{app:fixed-portfolios}), using paired differences in worst-task
percentile. The numerical-fidelity portfolio outperforms every generalist.
The categorical portfolio also has higher mean performance throughout:
generalist-minus-portfolio gaps range from $-.396$ to $-.261$, with all
bootstrap intervals below zero, although this cohort contains only three
datasets. Utility yields a different conclusion. TabICLv2 exceeds its
portfolio by $.157$ ($95\%$ CI $[.078,.229]$) and TabPFN-3 by $.142$
($[.051,.227]$). LimiX-2's interval spans zero, while the remaining methods
have negative mean gaps. Generalist deployment is therefore most competitive
for the evaluated utility profile; fidelity-oriented workloads continue to
benefit from specialization. Appendix~\ref{app:method-landscape} gives the
setting-specific comparisons and statistical diagnostics.

\paragraph{Implications for foundation models.}
LimiX-2, TabICLv2, and TabPFN-3 achieve strong numerical point imputation
and prediction without single-dataset fine-tuning, while their generation is
more competitive on utility than fidelity. LimiX-2 includes masked-feature
reconstruction, whereas TabICLv2 and TabPFN-3 reuse pretrained predictors
through conditional imputation
extensions~\citep{limix2,tabiclv2,tabpfn3,tabpfn_extensions}. Categorical
reconstruction on the common profile cohort and faithful joint generation
remain limitations. Amid rapid progress and growing interest in tabular
foundation models, these findings highlight their potential across all three
tasks. We hope our results encourage foundation-model researchers to develop
a unified model that excels at generation, imputation, and prediction.

%% file: sections/07_conclusion.tex
\section{Conclusion}
\label{sec:conclusion}

\ours{} provides a common setting for studying how generation, imputation,
and prediction relate through their capabilities, training objectives, and
fixed-model performance. Task-rank comparisons connect imputation and
prediction through numerical recovery and dependence preservation, and link
generation to both through predictive utility. Distributional fidelity
follows different relationships, so task competence cannot be inferred from
one criterion alone. Matched objective interventions show the largest pooled gains from recipient
self-inclusion and suggest backbone-dependent auxiliary benefits from
generation and prediction. Their
benefits need not add: joint outcomes depend on the recipient and the
objectives already present. Fixed-model profiles further show why utility
and fidelity can favor different generalist or specialist choices.

Foundation models already demonstrate strong prediction and numerical point
imputation without dataset-specific fine-tuning, together with promising
synthetic-data utility. Amid rapid progress in this field, these findings
highlight their potential across all three tasks. We hope \ours{} encourages
researchers to develop foundation models that excel at generation,
imputation, and prediction, including reliable categorical reconstruction
and faithful joint sampling.

%% file: sections/08_limitations_and_future_work.tex
\section{Limitations and Future Work}
\label{sec:limitations}

\paragraph{Scope and comparability.}
Our conclusions depend on the capability axes, information contracts, and
available method--dataset support. Rank alignment describes capability
co-occurrence, rather than identifying objective transfer. In particular,
the rank study's target-observed imputation and the controlled study's
target-hidden imputation provide complementary evidence under different
contracts. Profile and portfolio comparisons also use different common
cohorts; the categorical profile contains only three datasets. Cross-task
imputation coordinates average twelve settings and do not replace
setting-specific evaluation.

\paragraph{Objective effects and their mechanisms.}
The factorial study covers five backbones and five datasets, with only one
regression dataset and some unavailable evaluations. Auxiliary single-addition effects remain
exploratory and heterogeneous across backbones. SAE and CAE quantify changes
on a normalized rank scale; they do not identify the mechanisms behind those
changes. I/P self-inclusion and some interaction comparisons can also change
the conditioning policy, while fixed training budgets redistribute updates
among objectives. Objective overlap, update allocation, capacity competition,
and gradient interference therefore require separate interventions.

\paragraph{Directions for unified models.}
The frozen foundation-model results establish the capabilities of the
evaluated inference systems, without isolating the contribution of
pretraining, representations, or decoders. Matched pretraining and inference
ablations are needed before attributing categorical or generation gaps to
specific backbone properties. Future work should broaden the dataset and
model populations, separate training from inference effects, and test how
categorical reconstruction and joint sampling can improve alongside
prediction.

%% file: sections/statements.tex
\section*{AI Use Statement}

We used AI tools to assist with manuscript drafting and editing, figure
preparation, and LaTeX formatting. The authors are responsible for the final
manuscript and its scientific claims.

\section*{Reproducibility Statement}

To facilitate reproducibility, we describe the task definitions in
Section~\ref{sec:prelimilary_queries} and the evaluation and aggregation
protocols in Section~\ref{sec:metrics}.
Dataset sources, selection criteria, data splits, and preprocessing
procedures are detailed in Appendix~\ref{app:dataset-construction}.
Baseline configurations, hyperparameter settings and selection rules,
random seeds, and computational resources are provided in
Appendix~\ref{app:implementation}.
Metric definitions and result aggregation procedures are specified in
Appendices~\ref{app:methods-metrics} and~\ref{app:metric-audit}, with
analysis-specific protocols in
Appendices~\ref{app:task-rank-analysis}--\ref{app:method-landscape}.
Benchmark source code, method configurations, and evaluation routines are
publicly available at
\url{https://github.com/limix-ldm-ai/OTBench}.

\section*{Ethics Statement}

This study evaluates tabular learning methods using publicly available
datasets under their respective licenses and terms of use.
We do not collect additional personal data or attempt to identify
individuals in the datasets.
We recognize that synthetic data may retain sensitive information
from the training data, and that empirical privacy evaluations do not
constitute formal privacy guarantees.
Furthermore, aggregate benchmark performance does not establish
fairness across demographic groups or suitability for deployment in
high-stakes settings.
Any downstream use should therefore be accompanied by
application-specific assessments of privacy, fairness, and reliability.

%% file: sections/appendix.tex
\raggedbottom
\input{sections/a_dataset_construction}

\clearpage
\input{sections/b_methods_metrics}

\clearpage
\input{sections/c_metric_space}
\clearpage
\input{sections/d_task_rank_analysis}
\clearpage
\input{sections/e_causal_transfer}
\clearpage
\input{sections/f_method_landscape}
\clearpage
\input{sections/g_implementation}

\input{sections/h_related_work}

%% file: sections/a_dataset_construction.tex
\providecommand{\datasetfigdir}{assets}
\providecommand{\code}[1]{\texttt{#1}}
\providecommand{\clip}{\operatorname{clip}}

\section{Dataset Construction Details}
\label{app:dataset-construction}

\subsection{Construction Objective and Overall Pipeline}
\begin{figure}[htbp]
    \centering
    \begin{adjustbox}{max width=.98\linewidth,center}
    \begin{tikzpicture}[
        node distance=3.5mm and 2.0mm,
        every node/.style={
            font=\footnotesize,
            align=center,
            execute at begin node={
                \hyphenpenalty=10000
                \exhyphenpenalty=10000
                \relax
            }
        },
        box/.style={
            rectangle,
            draw=black,
            fill=none,
            line width=0.45pt,
            minimum width=.235\linewidth,
            minimum height=1.05cm,
            text width=.215\linewidth,
            inner sep=3pt
        },
        source/.style={box},
        process/.style={box},
        scm/.style={box,dashed},
        selected/.style={box,line width=0.70pt},
        final/.style={box,line width=0.85pt},
        flow/.style={
            -{Stealth[length=1.6mm,width=1.2mm]},
            line width=0.50pt,
            draw=black
        },
        external/.style={flow,dashed}
    ]

    \node[source] (lit) {
        \textbf{24 literature source groups}\\
        Candidate dataset discovery
    };

    \node[source,right=of lit] (pool) {
        \textbf{492 OpenML datasets}\\
        Version and task verification
    };

    \node[process,right=of pool] (filter) {
        \textbf{File and size constraints}\\
        $100 \le N \le 10^5$, $P \le 160$, $NP \le 10^6$
    };

    \node[selected,right=of filter] (cand) {
        \textbf{377 eligible datasets}\\
        From 469 complete files
    };

    \node[process,below=of cand] (rawmf) {
        \textbf{113 raw meta-features}\\
        Groups A/B/C/D: $9/68/16/20$
    };

    \node[process,left=of rawmf] (cleanmf) {
        \textbf{Cleaning and normalization}\\
        113 $\rightarrow$ 72 dimensions
    };

    \node[process,left=of cleanmf] (pca) {
        \textbf{PCA representation}\\
        41 components (99.1\%)
    };

    \node[process,left=of pca] (km) {
        \textbf{$k$-means clustering}\\
        $k=50$, \texttt{n\_init=10}
    };

    \node[selected,below=of pca] (openml) {
        \textbf{50 OpenML datasets}\\
        Nearest eligible dataset\\
        to each centroid
    };

    \node[process,left=of openml] (dedup) {
        \textbf{Same-source suppression}\\
        31 candidates blocked
    };

    \node[final,right=of openml] (benchmark) {
        \textbf{Complete benchmark}\\
        56 datasets
    };

    \node[scm,right=of benchmark] (scmnode) {
        \textbf{6 TabStruct SCM datasets}\\
        Independent structural track
    };

    \draw[flow] (lit) -- (pool);
    \draw[flow] (pool) -- (filter);
    \draw[flow] (filter) -- (cand);
    \draw[flow] (cand) -- (rawmf);
    \draw[flow] (rawmf) -- (cleanmf);
    \draw[flow] (cleanmf) -- (pca);
    \draw[flow] (pca) -- (km);
    \draw[flow] (km) -- (dedup);
    \draw[flow] (dedup) -- (openml);
    \draw[flow] (openml) -- (benchmark);
    \draw[external] (scmnode) -- (benchmark);

    \end{tikzpicture}
    \end{adjustbox}

    \caption{
    Dataset construction pipeline.
    The OpenML track uses prespecified size constraints, a meta-feature
    representation, and clustering to obtain 50 representative datasets.
    The 6 SCM datasets form a separate structural evaluation track.
    }
    \label{fig:dataset-pipeline}
\end{figure}
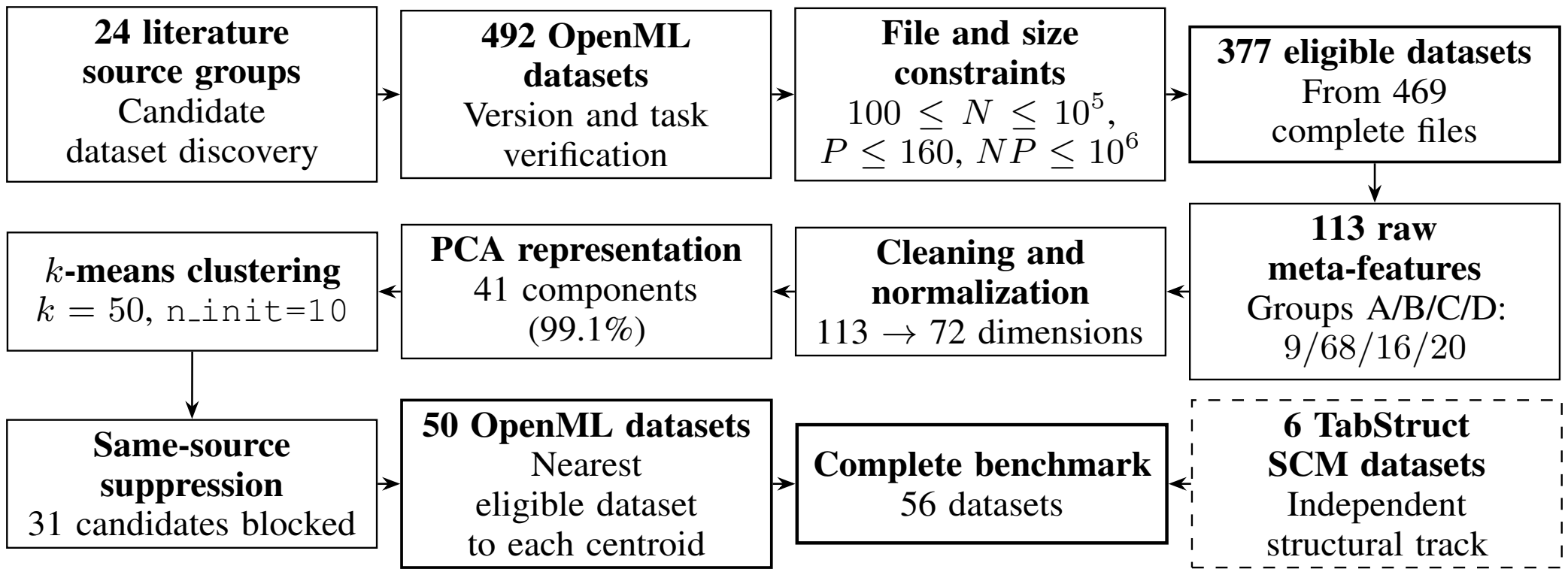

The benchmark supports prediction, missing-value imputation, and data generation under a controlled computational budget. Evaluating every candidate dataset would be costly and would overrepresent sources with multiple versions and groups of structurally similar datasets.
We therefore select representative datasets using only method-independent dataset properties and fixed landmarking statistics~\citep{pfahringer2000landmarking}.
All selection rules and configurations are specified before method evaluation begins, so no evaluated method's performance influences dataset selection.
Figure~\ref{fig:dataset-pipeline} summarizes the construction process, from the literature-derived candidate pool to the two final evaluation sets.

\subsection{Literature-Driven Candidate Dataset Pool}
We identify candidate datasets from 24 deduplicated literature source groups.
These sources cover general tabular prediction benchmarks as well as specialized studies of missing-value imputation, data generation, structural fidelity, and robustness.
Table~\ref{tab:dataset-sources-single} lists these sources.
ForestDiffusion covers both generation and imputation and is counted once; the two relevant TALENT studies form one source group.
Literature sources define only the candidate pool and do not determine the weights used for representative selection.

After normalizing dataset names and common aliases, we resolve dataset identities using Dataset IDs reported in papers or official configurations, Task IDs that can be mapped to specific versions, frozen identifiers in official benchmark suites, and metadata that agree in dataset name, size, task type, and original source.
We merge references to the same Dataset ID across studies while retaining all source attributions.
Records with the same name but different Dataset IDs, versions, or sizes are not merged automatically.
This process yields 492 datasets with unambiguous OpenML Dataset IDs~\citep{openml}.

\begin{table}[htbp]
\centering
\caption{The 24 deduplicated literature source groups used to construct the candidate dataset pool.}
\label{tab:dataset-sources-single}
\small
\begin{tabularx}{0.98\linewidth}{>{\bfseries\raggedright\arraybackslash}p{0.22\linewidth}>{\raggedright\arraybackslash}X}
\toprule
Category & Literature sources \\
\midrule
Imputation and cross-task studies
& Shades-of-Null~\citep{shadesnull}; TabImpute~\citep{tabimpute}; ForestDiffusion~\citep{forestdiffusion}; DiffPuter~\citep{diffputer}; TabCSDI~\citep{tabcsdi}; MissDiff~\citep{missdiff}; ReMasker~\citep{remasker}; HyperImpute~\citep{hyperimpute}; TabGenDDPM~\citep{tabgenddpm}.\\
Generation and structural evaluation
& TabStruct~\citep{tabstruct}; Systematic Assessment of Tabular Data Synthesis~\citep{systematic_synthesis}; Synthcity~\citep{synthcity}; Structured Evaluation of Synthetic Tabular Data~\citep{structuredeval}; CauTabBench~\citep{cautabbench}; SynthEval~\citep{syntheval}; TabDDPM~\citep{tabddpm}; TabDiff~\citep{tabdiff}.\\
Prediction and robustness
& TabArena~\citep{tabarena}; TALENT~\citep{talent,talent_closer}; TabZilla~\citep{tabzilla}; Why Trees~\citep{whytrees}; TabReD~\citep{tabred}; CARD~\citep{card}; TabularBench~\citep{tabularbench}.\\
\bottomrule
\end{tabularx}
\end{table}

\subsection{Data Parsing and Hard Filtering}

\paragraph{Column types and target columns.}
Variable types follow the ARFF header rather than the appearance of the values. Nominal attributes are treated as categorical variables, whereas \code{numeric}, \code{real}, and \code{integer} attributes are treated as numerical variables. A \code{string} attribute is treated as categorical only when its cardinality does not exceed $\min(50,0.2N_i)$. Other free-text fields and all \code{date}/\code{relational} attributes are removed. Thus, $P_i$ counts usable features after removing target columns and unsupported fields, rather than raw CSV columns.

We use the OpenML \code{default\_target\_attribute} field to identify the target whenever available. If this field is missing, the target is inferred in order from common names, including \code{class}, \code{label}, \code{target}, \code{y}, and \code{outcome}, or from a low-cardinality final column. Such cases are marked by \code{target\_inferred}. A numerical final column is inferred as a classification target only when it has between 2 and 30 unique values and the sample size is at least 20 times the number of unique values. For multi-target datasets, only the first target is used to compute prediction-related meta-features, while the original number of targets is retained.

\begin{figure}[htbp]
    \centering
    \begin{subfigure}[t]{0.49\linewidth}
      \centering
      \includegraphics[width=\linewidth]{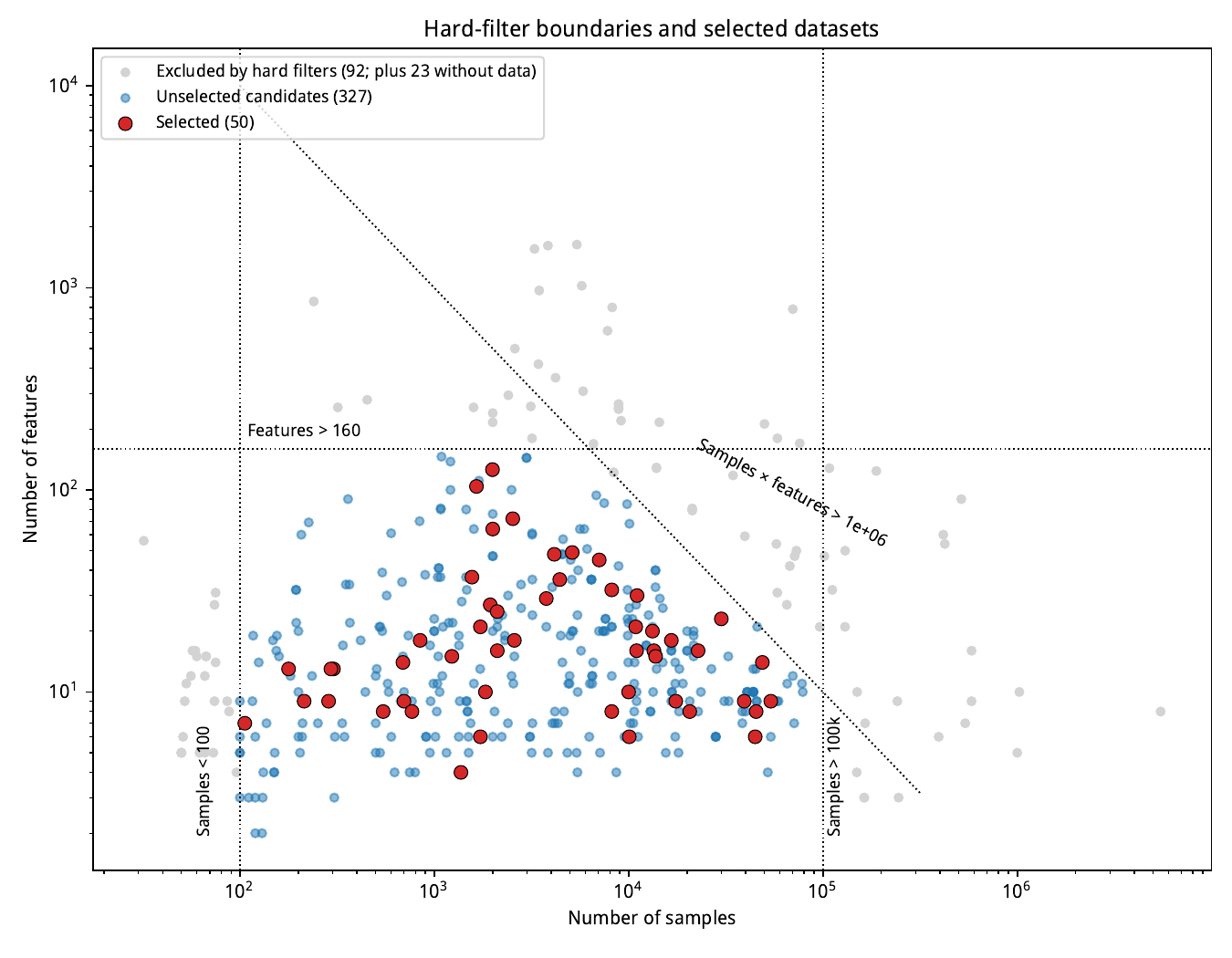}
      \caption{Hard-filtering boundaries and selected datasets.}
    \end{subfigure}\hfill
    \begin{subfigure}[t]{0.49\linewidth}
      \centering
      \includegraphics[width=\linewidth]{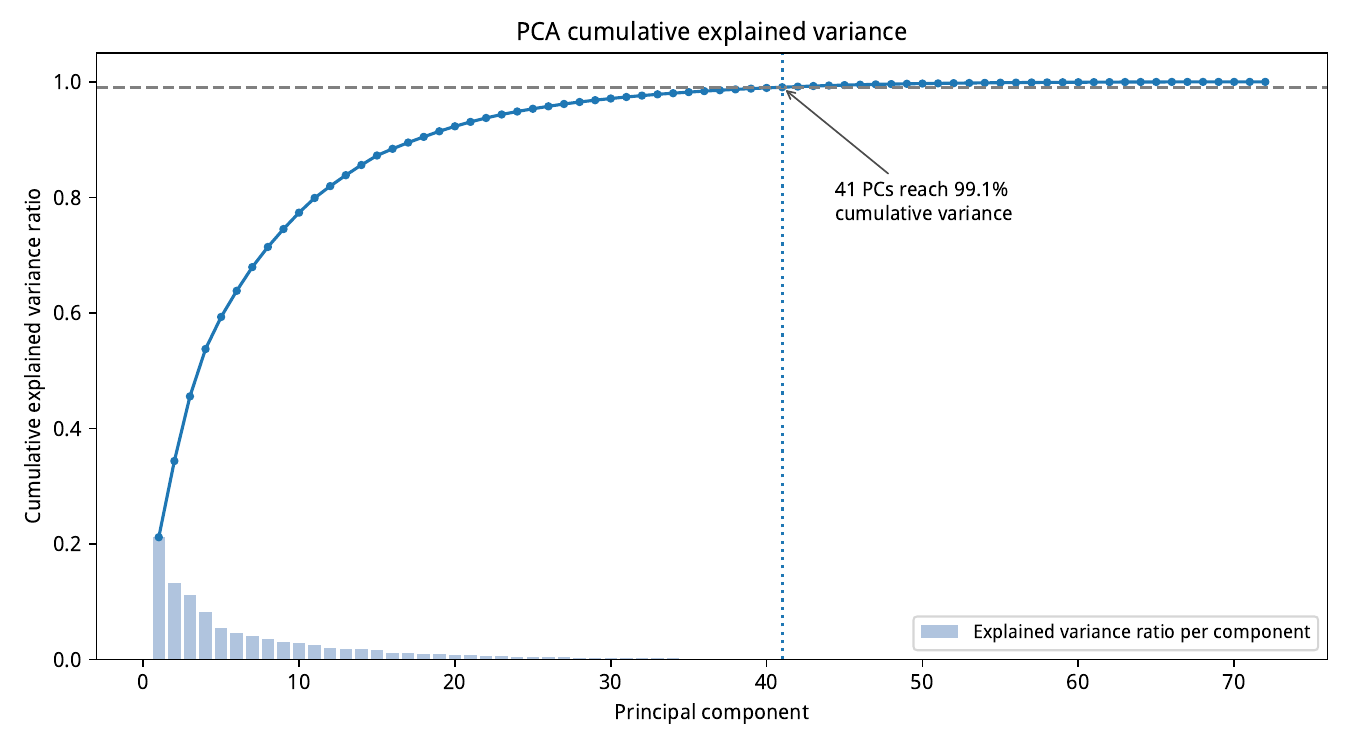}
      \caption{PCA cumulative explained variance. Forty-one components reach the 99\% threshold.}
    \end{subfigure}
    \caption{Dataset size constraints and PCA dimensionality reduction. In the left panel, gray points denote datasets removed by the size constraints, blue points denote eligible but unselected candidates, and red points denote the selected datasets.}
    \label{fig:dataset-filter-pca}
\end{figure}

\paragraph{Size constraints.}
To keep evaluation within a common computational budget, we impose the following constraints on each dataset $\mathcal D_i$:
\begin{equation}
100\leq N_i\leq100{,}000,\qquad
P_i\leq160,\qquad
N_iP_i\leq10^6.
\label{eq:dataset-hard-filter-single}
\end{equation}
The 492 source entries yield 469 datasets for size screening.
The size constraints exclude 92 datasets, leaving 377 candidates.
Figure~\ref{fig:dataset-filter-pca}(a) shows the filtering outcome in sample-size--feature-count space.

\subsection{Dataset Meta-Feature Representation}

We recompute 113 meta-features for each candidate dataset using a unified local extractor.
The extractor includes reproducible PyMFE statistics~\citep{alcobaca2020mfe} and standard clustering diagnostics~\citep{hopkins1954new,rousseeuw1987silhouettes}.
The four meta-feature groups serve only for interpretation and post hoc diagnostics; we assign no manual group weights.
Table~\ref{tab:metafeatures-single} reports the original counts and lists all dimensions retained after cleaning.

\begingroup
\scriptsize
\setlength{\tabcolsep}{3pt}
\renewcommand{\arraystretch}{1.13}
\renewcommand{\UrlFont}{\ttfamily\scriptsize}
\begin{longtable}{>{\centering\arraybackslash}p{0.045\linewidth}
                  >{\raggedleft\arraybackslash}p{0.055\linewidth}
                  >{\raggedleft\arraybackslash}p{0.055\linewidth}
                  >{\raggedright\arraybackslash}p{0.775\linewidth}}
\caption{Composition of the candidate-dataset meta-features and all 72 dimensions retained after cleaning.}
\label{tab:metafeatures-single}\\
\toprule
Group & Original & Retained & Retained meta-features \\
\midrule
\endfirsthead
\multicolumn{4}{l}{\footnotesize Table~\ref{tab:metafeatures-single} (continued)}\\
\toprule
Group & Original & Retained & Retained meta-features \\
\midrule
\endhead
\midrule
\multicolumn{4}{r}{\footnotesize Continued on next page}\\
\endfoot
\bottomrule
\endlastfoot
A & 9 & 9
& \nolinkurl{cat_feature_ratio}; \nolinkurl{frac_binary_features}; \nolinkurl{frac_numeric_lowcard}; \nolinkurl{log_attr_to_inst}; \nolinkurl{log_n_features}; \nolinkurl{log_n_samples}; \nolinkurl{max_log_cat_cardinality}; \nolinkurl{mean_log_cat_cardinality}; \nolinkurl{mean_numeric_unique_ratio}.\\
\addlinespace[2pt]
B & 68 & 29
& \nolinkurl{duplicate_row_ratio}; \nolinkurl{frac_bounded_01}; \nolinkurl{frac_continuous_cols}; \nolinkurl{frac_heavy_tail}; \nolinkurl{frac_integer_valued}; \nolinkurl{frac_multimodal}; \nolinkurl{frac_nonneg}; \nolinkurl{frac_outlier_rows}; \nolinkurl{frac_skewed}; \nolinkurl{hopkins}; \nolinkurl{kmeans_silhouette_max}; \nolinkurl{mean_n_modes}; \nolinkurl{pymfe_attr_conc.mean}; \nolinkurl{pymfe_attr_conc.sd}; \nolinkurl{pymfe_attr_ent.mean}; \nolinkurl{pymfe_attr_ent.sd}; \nolinkurl{pymfe_cor.mean}; \nolinkurl{pymfe_cor.sd}; \nolinkurl{pymfe_frac_bin}; \nolinkurl{pymfe_frac_norm}; \nolinkurl{pymfe_frac_outliers}; \nolinkurl{pymfe_kurtosis.mean}; \nolinkurl{pymfe_kurtosis.sd}; \nolinkurl{pymfe_nr_cor_attr}; \nolinkurl{pymfe_skewness.mean}; \nolinkurl{pymfe_skewness.sd}; \nolinkurl{pymfe_sparsity.mean}; \nolinkurl{pymfe_sparsity.sd}; \nolinkurl{zero_ratio}.\\
\addlinespace[2pt]
C & 16 & 16
& \nolinkurl{cat_num_association}; \nolinkurl{frac_cols_with_missing}; \nolinkurl{frac_rows_with_missing}; \nolinkurl{frac_strong_cor_pairs}; \nolinkurl{max_abs_cor}; \nolinkurl{max_col_missing_ratio}; \nolinkurl{max_norm_mutual_info}; \nolinkurl{mean_abs_cor}; \nolinkurl{mean_missing_per_missing_row}; \nolinkurl{mean_norm_mutual_info}; \nolinkurl{missing_cell_ratio}; \nolinkurl{missing_mar_score}; \nolinkurl{missing_pattern_count_log}; \nolinkurl{missing_pattern_entropy}; \nolinkurl{pca_dims_95_ratio}; \nolinkurl{top_eigen_ratio}.\\
\addlinespace[2pt]
D & 20 & 18
& \nolinkurl{class_entropy_norm}; \nolinkurl{class_imbalance_log}; \nolinkurl{is_binary}; \nolinkurl{is_multiclass}; \nolinkurl{is_regression}; \nolinkurl{lm_best}; \nolinkurl{lm_knn}; \nolinkurl{lm_linear}; \nolinkurl{lm_rf}; \nolinkurl{lm_tree}; \nolinkurl{log_n_classes}; \nolinkurl{minority_class_ratio}; \nolinkurl{n_targets_log}; \nolinkurl{noise_level}; \nolinkurl{nonlinearity_gain}; \nolinkurl{target_skew_abs}; \nolinkurl{target_unique_ratio}; \nolinkurl{tree_vs_knn}.\\
\midrule
Total & 113 & 72 & -- \\
\end{longtable}
\endgroup

\paragraph{Sampling and failure handling.}
Dataset size, missingness rates, missingness patterns, and the duplicate-row ratio are computed on the full dataset.
Other computationally expensive statistics use a random sample of at most 10,000 records.
For meta-feature computation, we impute missing numerical values with the median and fill entirely missing numerical columns with 0.
Missing categorical values are encoded as \code{\_\_missing\_\_}.
These operations apply only to meta-feature computation and leave the original data unchanged.
Failures are handled separately in the PyMFE, clustering-structure, dependency-structure, and landmarking modules.
If a module fails, only its dimensions are marked as missing.

\paragraph{Landmarking.}
Following the landmarking principle~\citep{pfahringer2000landmarking}, we characterize target predictability with four fixed lightweight models: a logistic/ridge linear model, a depth-3 decision tree, 5-nearest neighbors, and a random forest with 100 trees~\citep{hoerl1970ridge,breiman1984cart,cover1967nearest,randomforest}. Within each fold, numerical features are median-imputed and standardized. Categorical features are constant-imputed and one-hot encoded, with at most 20 categories retained per column. We remove records with missing targets and, for classification, classes with fewer than 6 samples. If more than 5,000 records remain, 5,000 are sampled without replacement using random seed 0, followed by three-fold cross-validation. This limit applies only to auxiliary meta-feature computation; the final dataset is not truncated. For a $K$-class classification task and a regression task, respectively, the scores are
\begin{equation}
s_{\mathrm{cls}}=\clip\!\left(\frac{\operatorname{BalAcc}-1/K}{1-1/K},0,1\right),
\qquad
s_{\mathrm{reg}}=\clip(R^2,0,1).
\label{eq:landmarking-single}
\end{equation}
These scores describe the dataset's target structure and exclude the performance of all methods evaluated in this paper.

\subsection{Meta-Feature Cleaning and Normalization}

Of the raw meta-features, 28 PyMFE statistics depend directly on the original measurement units, including means, standard deviations, extrema, covariances, and raw eigenvalues.
Another 11 PyMFE counts or ratios duplicate the logarithmic or proportional representations in group A.
The features \code{has\_target} and \code{target\_missing\_ratio} are nearly constant in the current candidate pool.
Removing these 41 dimensions leaves 72 meta-features.
We apply no further within-group correlation pruning.
PCA summarizes the remaining correlated variation in a lower-dimensional representation.

For datasets without an identified target, missing group-D meta-features are imputed using the corresponding median over datasets with valid targets.
All other missing values are imputed using the median of the corresponding dimension over the full candidate pool.
Any meta-feature with more than 20\% missing values would also be removed during preprocessing.

Meta-features with at most two observed unique values are standardized using a $z$-score. All other dimensions are mapped to a standard normal distribution by a \texttt{QuantileTransformer}~\citep{pedregosa2011scikit} fitted with at most 200 quantiles, then clipped to $[-3.5,3.5]$:
\begin{equation}
\widetilde h_{ij}
=
\clip\!\left[
\Phi^{-1}\!\left(\widehat F^{(200)}_j(h_{ij})\right),
-3.5,3.5
\right],
\label{eq:quantile-single}
\end{equation}
where $\widehat F^{(200)}_j$ denotes the quantile map fitted with at most 200 empirical quantiles. Of the 72 retained dimensions, 69 use quantile normalization and 3 binary task-type indicators use $z$-score standardization.

\subsection{PCA Representation, Clustering, and Representative Selection}

\paragraph{PCA representation.}
Let $H\in\mathbb R^{377\times72}$ denote the standardized candidate-dataset matrix. We fit PCA~\citep{jolliffe2002pca} without whitening and choose the smallest dimension $q$ satisfying
\begin{equation}
q=\min\left\{r:
\frac{\sum_{j=1}^{r}\lambda_j}
     {\sum_{j=1}^{72}\lambda_j}
\geq0.99\right\}.
\label{eq:pca-single}
\end{equation}
This gives $q=41$ and 99.0726\% cumulative explained variance. The 41 principal components (PCs) are linear combinations of the 72 meta-features, not a subset of 41 original features. Figure~\ref{fig:dataset-filter-pca}(b) plots the cumulative explained variance against the number of PCs and marks the 41-component cutoff required to reach the 99\% threshold.

\paragraph{Clustering and representative datasets.}
Let $\mathbf e_i\in\mathbb R^{41}$ denote the PCA coordinates of dataset $\mathcal D_i$. We apply $k$-means~\citep{lloyd1982least} in this space with $k=50$, \code{n\_init}$=10$, and random seed 0. The clustering objective is
\begin{equation}
\min_{\{a_i\},\{\boldsymbol\mu_c\}}
\sum_{i=1}^{377}
\left\|\mathbf e_i-\boldsymbol\mu_{a_i}\right\|_2^2.
\label{eq:kmeans-single}
\end{equation}
Clusters are processed in decreasing order of size. Within each cluster, we select the candidate nearest to the centroid among those allowed by the same-source rule. Let $\mathcal E_c$ denote the set of eligible nonduplicate candidates when cluster $c$ is processed. The selected representative is
\begin{equation}
\mathcal D_c^\star
=
\arg\min_{\mathcal D_i\in C_c\cap\mathcal E_c}
\left\|\mathbf e_i-\boldsymbol\mu_c\right\|_2.
\label{eq:representative-single}
\end{equation}
The representative is an observed candidate dataset; the synthetic centroid cannot be evaluated directly.

\begin{figure}[htbp]
    \centering
    \includegraphics[width=0.78\linewidth]{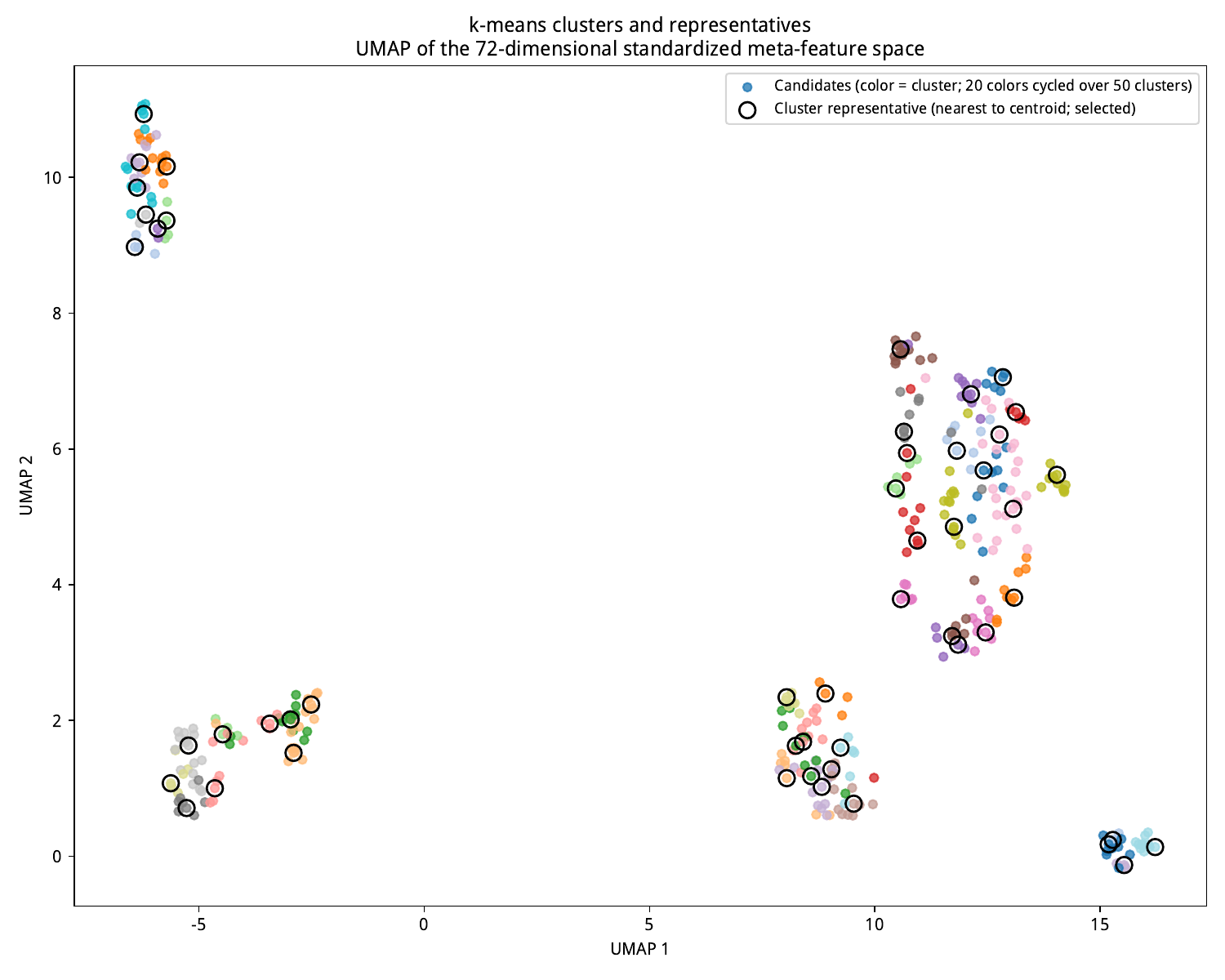}
    \caption{Two-dimensional UMAP projection of the candidate datasets and selected representatives. Colors denote $k$-means clusters, with the palette reused across clusters, and black circles mark the selected representative of each cluster. UMAP is used only for visualization; PCA, clustering, and representative selection are all performed in the high-dimensional space.}
    \label{fig:dataset-umap-single}
\end{figure}

\paragraph{Same-source version suppression.}
Dataset names are converted to lowercase and stripped of nonalphanumeric characters. Two datasets are marked as potential versions of the same source if their normalized names are identical, or if the shorter normalized name has at least 5 characters and is a prefix of the other, provided that their sample counts differ by no more than a factor of 1.5. After one version is selected, other versions from the same source cannot represent later clusters. This procedure excludes 31 potential duplicates and selects one representative from each of the 50 clusters. Figure~\ref{fig:dataset-umap-single} shows a two-dimensional UMAP projection~\citep{mcinnes2018umap} of the candidate datasets and selected representatives.

\subsection{Final Benchmark, Data Splits, and Structural Evaluation Set}

\paragraph{OpenML evaluation set.}
The 50 selected OpenML-hosted datasets comprise 26 binary-classification, 9 multiclass-classification, and 15 regression datasets. By source type, 40 datasets are real-world or derived from real-world data, 9 are synthetic, simulated, or manually constructed, and 1 combines real and synthetic data. Task type and source type are separate classification schemes. The datasets contain 106--53,940 samples and 4--126 features. The largest table has 690,000 cells, and 8 datasets contain native missing values.

\paragraph{Fixed train--test split.}
Dataset selection is completed before the final train--test split. Let $\mathcal D_i=\{z_{ij}\}_{j=1}^{N_i}$ have sample-index set $I_i=\{1,\ldots,N_i\}$. We define:
\begin{equation}
    \begin{aligned}
    & \mathcal D_{i,\mathrm{train}}=\{z_{ij}:j\in I_{i,\mathrm{train}}\}, \qquad & & \mathcal D_{i,\mathrm{test}}=\{z_{ij}:j\in I_{i,\mathrm{test}}\},\\
    & I_{i,\mathrm{train}}\cap I_{i,\mathrm{test}}=\varnothing, \qquad & & I_{i,\mathrm{train}}\cup I_{i,\mathrm{test}}=I_i.
    \end{aligned}
    \label{eq:dataset-split-single}
\end{equation}
By default, the data are split once in an 80\%/20\% ratio. Classification datasets are stratified by the target whenever class counts permit. Generation, imputation, and prediction share the same fixed split. All model parameters, value domains, category sets, and normalization statistics are determined only from the training set. Methods that require validation data create an internal split from the training set, while the test set is reserved for final held-out evaluation.

\paragraph{Structural causal model evaluation set.}
In addition to the OpenML evaluation set, we include the six structural causal model datasets used by TabStruct~\citep{tabstruct}: Hailfinder, Healthcare, Insurance, Magic-Irri, Mehra, and Sangiovese. Their data-generating mechanisms or variable-dependence structures are specified in advance, allowing controlled evaluation of structural recovery. These datasets are excluded from the clustering of the OpenML candidate pool and are reported separately as the structural evaluation set.

\clearpage
\subsection{Complete Dataset Inventory}
Table~\ref{tab:dataset-inventory} summarizes the complete 50-dataset OpenML
evaluation set and the six-dataset structural evaluation set, including their
task types, sample and feature counts, variable composition, and class
statistics where applicable.

\input{tables/dataset_inventory}

%% file: tables/dataset_inventory.tex
\begingroup
\scriptsize
\setlength{\tabcolsep}{1.5pt}
\renewcommand{\arraystretch}{1.00}
\begin{longtable}{@{}>{\raggedright\arraybackslash}p{0.18\linewidth}
                        >{\raggedright\arraybackslash}p{0.07\linewidth}
                        >{\raggedleft\arraybackslash}p{0.055\linewidth}
                        >{\centering\arraybackslash}p{0.04\linewidth}
                        >{\raggedleft\arraybackslash}p{0.075\linewidth}
                        >{\raggedleft\arraybackslash}p{0.04\linewidth}
                        >{\raggedleft\arraybackslash}p{0.055\linewidth}
                        >{\raggedleft\arraybackslash}p{0.045\linewidth}
                        >{\raggedleft\arraybackslash}p{0.045\linewidth}
                        >{\raggedleft\arraybackslash}p{0.035\linewidth}
                        >{\raggedleft\arraybackslash}p{0.045\linewidth}
                        >{\raggedleft\arraybackslash}p{0.052\linewidth}
                        >{\raggedleft\arraybackslash}p{0.052\linewidth}@{}}
\caption{Complete inventory of the 56 benchmark datasets.  \(N\) combines
the fixed training and test splits; \(D\) excludes the target.  Num., Cat.,
and ID give the numbers of numerical, categorical, and identifier columns.
Class counts and class-size ranges apply only to classification datasets.}
\label{tab:dataset-inventory}\\
\toprule
Dataset & Source & \makecell[r]{OML\\ID} & Task & \(N\) & \(D\) & \(N/D\) &
\makecell[r]{Num.} & \makecell[r]{Cat.} & \makecell[r]{ID} &
\makecell[r]{Cls.} & \makecell[r]{Class\\min} & \makecell[r]{Class\\max} \\
\midrule
\endfirsthead
\multicolumn{13}{l}{\footnotesize Table~\ref{tab:dataset-inventory} (continued)}\\
\toprule
Dataset & Source & \makecell[r]{OML\\ID} & Task & \(N\) & \(D\) & \(N/D\) &
\makecell[r]{Num.} & \makecell[r]{Cat.} & \makecell[r]{ID} &
\makecell[r]{Cls.} & \makecell[r]{Class\\min} & \makecell[r]{Class\\max} \\
\midrule
\endhead
\midrule
\multicolumn{13}{r}{\footnotesize Continued on next page}\\
\endfoot
\bottomrule
\endlastfoot
adult-\allowbreak{}census & OpenML & 1119 & Cls. & 32,561 & 15 & 2170.73 & 6 & 8 & 1 & 2 & 7,841 & 24,720 \\
allbp & OpenML & 40707 & Cls. & 3,772 & 29 & 130.07 & 6 & 23 & 0 & 3 & 14 & 3,609 \\
Amazon\_\allowbreak{}employee\_\allowbreak{}access & OpenML & 46905 & Cls. & 32,769 & 9 & 3641.00 & 0 & 9 & 0 & 2 & 1,897 & 30,872 \\
appendicitis\_\allowbreak{}test & OpenML & 45589 & Cls. & 106 & 7 & 15.14 & 7 & 0 & 0 & 2 & 21 & 85 \\
Australian & OpenML & 40981 & Cls. & 690 & 14 & 49.29 & 6 & 8 & 0 & 2 & 307 & 383 \\
auto\_\allowbreak{}price & OpenML & 745 & Cls. & 159 & 15 & 10.60 & 14 & 1 & 0 & 2 & 54 & 105 \\
avocado\_\allowbreak{}sales & OpenML & 43927 & Reg. & 18,249 & 13 & 1403.77 & 11 & 2 & 0 & -- & -- & -- \\
bank32nh & OpenML & 833 & Cls. & 8,192 & 32 & 256.00 & 32 & 0 & 0 & 2 & 2,543 & 5,649 \\
BNG-\allowbreak{}echoMonths & OpenML & 1199 & Reg. & 17,496 & 9 & 1944.00 & 6 & 3 & 0 & -- & -- & -- \\
Brazilian\_\allowbreak{}houses & OpenML & 44062 & Reg. & 10,692 & 11 & 972.00 & 8 & 3 & 0 & -- & -- & -- \\
car\_\allowbreak{}evaluation & OpenML & 43921 & Cls. & 1,728 & 6 & 288.00 & 0 & 6 & 0 & 4 & 65 & 1,210 \\
cleveland & OpenML & 786 & Cls. & 303 & 13 & 23.31 & 6 & 7 & 0 & 2 & 139 & 164 \\
colleges & OpenML & 42727 & Reg. & 7,063 & 45 & 156.96 & 32 & 12 & 1 & -- & -- & -- \\
CPMP-\allowbreak{}2015-\allowbreak{}regression & OpenML & 41700 & Reg. & 2,108 & 25 & 84.32 & 23 & 2 & 0 & -- & -- & -- \\
cpu\_\allowbreak{}act & OpenML & 45029 & Reg. & 8,192 & 21 & 390.10 & 21 & 0 & 0 & -- & -- & -- \\
credit\_\allowbreak{}card\_\allowbreak{}clients\_\allowbreak{}default & OpenML & 46919 & Cls. & 30,000 & 23 & 1304.35 & 20 & 3 & 0 & 2 & 6,636 & 23,364 \\
delta\_\allowbreak{}ailerons & OpenML & 803 & Cls. & 7,129 & 5 & 1425.80 & 5 & 0 & 0 & 2 & 3,346 & 3,783 \\
diamonds & OpenML & 46923 & Reg. & 53,940 & 9 & 5993.33 & 6 & 3 & 0 & -- & -- & -- \\
E-\allowbreak{}CommereShippingData & OpenML & 46924 & Cls. & 10,999 & 10 & 1099.90 & 6 & 4 & 0 & 2 & 4,436 & 6,563 \\
FICO-\allowbreak{}HELOC-\allowbreak{}cleaned & OpenML & 45554 & Cls. & 9,871 & 23 & 429.17 & 21 & 2 & 0 & 2 & 4,735 & 5,136 \\
FOREX\_\allowbreak{}audjpy-\allowbreak{}hour-\allowbreak{}High & OpenML & 41718 & Cls. & 43,825 & 10 & 4382.50 & 10 & 0 & 0 & 2 & 21,254 & 22,571 \\
healthcare\_\allowbreak{}insurance\_\allowbreak{}expenses & OpenML & 46931 & Reg. & 1,338 & 6 & 223.00 & 3 & 3 & 0 & -- & -- & -- \\
heloc & OpenML & 45026 & Cls. & 10,000 & 22 & 454.55 & 22 & 0 & 0 & 2 & 5,000 & 5,000 \\
houses & OpenML & 46934 & Reg. & 20,640 & 8 & 2580.00 & 8 & 0 & 0 & -- & -- & -- \\
JapaneseVowels & OpenML & 976 & Cls. & 9,961 & 14 & 711.50 & 14 & 0 & 0 & 2 & 1,614 & 8,347 \\
jungle\_\allowbreak{}chess\_\allowbreak{}2pcs\_\allowbreak{}raw\_\allowbreak{}endgame\_\allowbreak{}complete & OpenML & 41027 & Cls. & 44,819 & 6 & 7469.83 & 6 & 0 & 0 & 3 & 4,335 & 23,062 \\
kc1 & OpenML & 1067 & Cls. & 2,109 & 21 & 100.43 & 21 & 0 & 0 & 2 & 326 & 1,783 \\
KDDCup09\_\allowbreak{}upselling & OpenML & 44186 & Cls. & 5,128 & 49 & 104.65 & 34 & 15 & 0 & 2 & 2,564 & 2,564 \\
kr-\allowbreak{}vs-\allowbreak{}kp & OpenML & 3 & Cls. & 3,196 & 36 & 88.78 & 0 & 36 & 0 & 2 & 1,527 & 1,669 \\
MiamiHousing2016 & OpenML & 43093 & Reg. & 13,932 & 16 & 870.75 & 16 & 0 & 0 & -- & -- & -- \\
MiceProtein & OpenML & 40966 & Cls. & 1,080 & 81 & 13.33 & 77 & 3 & 1 & 8 & 105 & 150 \\
mushroom & OpenML & 43923 & Cls. & 12,960 & 8 & 1620.00 & 0 & 8 & 0 & 5 & 2 & 4,320 \\
ozone-\allowbreak{}level-\allowbreak{}8hr & OpenML & 1487 & Cls. & 2,534 & 72 & 35.19 & 72 & 0 & 0 & 2 & 160 & 2,374 \\
pendigits & OpenML & 42857 & Cls. & 10,992 & 16 & 687.00 & 16 & 0 & 0 & 9 & 6 & 3,566 \\
Performance-\allowbreak{}Prediction & OpenML & 43812 & Cls. & 1,340 & 19 & 70.53 & 19 & 0 & 0 & 2 & 509 & 831 \\
planning-\allowbreak{}relax & OpenML & 1490 & Cls. & 182 & 12 & 15.17 & 12 & 0 & 0 & 2 & 52 & 130 \\
primary-\allowbreak{}tumor & OpenML & 1003 & Cls. & 339 & 17 & 19.94 & 0 & 17 & 0 & 2 & 84 & 255 \\
puma8NH & OpenML & 816 & Cls. & 8,192 & 8 & 1024.00 & 8 & 0 & 0 & 2 & 4,078 & 4,114 \\
pwLinear & OpenML & 721 & Cls. & 200 & 10 & 20.00 & 10 & 0 & 0 & 2 & 97 & 103 \\
qsar-\allowbreak{}biodeg & OpenML & 1494 & Cls. & 1,055 & 41 & 25.73 & 41 & 0 & 0 & 2 & 356 & 699 \\
QSAR\_\allowbreak{}fish\_\allowbreak{}toxicity & OpenML & 46954 & Reg. & 907 & 6 & 151.17 & 6 & 0 & 0 & -- & -- & -- \\
SolarPower & OpenML & 42328 & Reg. & 204 & 4 & 51.00 & 4 & 0 & 0 & -- & -- & -- \\
steel-\allowbreak{}plates-\allowbreak{}fault & OpenML & 40982 & Cls. & 1,941 & 27 & 71.89 & 27 & 0 & 0 & 7 & 55 & 673 \\
stock\_\allowbreak{}fardamento02 & OpenML & 42545 & Reg. & 6,277 & 6 & 1046.17 & 5 & 1 & 0 & -- & -- & -- \\
us\_\allowbreak{}crime & OpenML & 42730 & Reg. & 1,994 & 126 & 15.83 & 126 & 0 & 0 & -- & -- & -- \\
vehicle & OpenML & 994 & Cls. & 846 & 18 & 47.00 & 18 & 0 & 0 & 2 & 218 & 628 \\
website\_\allowbreak{}phishing & OpenML & 46963 & Cls. & 1,353 & 9 & 150.33 & 0 & 9 & 0 & 3 & 103 & 702 \\
wine & OpenML & 44091 & Cls. & 2,554 & 11 & 232.18 & 11 & 0 & 0 & 2 & 1,277 & 1,277 \\
wine\_\allowbreak{}quality & OpenML & 44136 & Reg. & 6,497 & 11 & 590.64 & 11 & 0 & 0 & -- & -- & -- \\
WineDataset & OpenML & 43571 & Cls. & 178 & 13 & 13.69 & 13 & 0 & 0 & 3 & 48 & 71 \\
hailfinder & TabStruct & -- & Cls. & 20,000 & 55 & 363.64 & 0 & 55 & 0 & 3 & 4,988 & 8,993 \\
healthcare & TabStruct & -- & Reg. & 20,000 & 6 & 3333.33 & 3 & 3 & 0 & -- & -- & -- \\
insurance & TabStruct & -- & Cls. & 20,000 & 26 & 769.23 & 0 & 26 & 0 & 4 & 367 & 11,176 \\
magic-\allowbreak{}irri & TabStruct & -- & Reg. & 20,000 & 63 & 317.46 & 63 & 0 & 0 & -- & -- & -- \\
mehra & TabStruct & -- & Reg. & 20,000 & 23 & 869.57 & 17 & 6 & 0 & -- & -- & -- \\
sangiovese & TabStruct & -- & Cls. & 20,000 & 14 & 1428.57 & 14 & 0 & 0 & 16 & 1,164 & 1,359 \\
\end{longtable}
\endgroup

%% file: sections/b_methods_metrics.tex
\section{Methods and Quality Metric Collection}
\label{app:methods-metrics}

This appendix lists the selected method identities and the quality metrics
displayed in Figures~\ref{fig:metric-correlations-generation}
and~\ref{fig:metric-correlations-imputation-prediction}.  The collection
contains 20 generation, 10 imputation, and 15 prediction metrics.
The 20 retained semantic axes and their 42 metric members are specified
separately in Table~\ref{tab:metric-groups}.

\subsection{Method capabilities}
\label{app:method-capabilities}

Table~\ref{tab:app-method-capabilities} lists 45 selected method identities,
including benchmark controls.  G, I, and P denote generation, imputation, and
prediction.  Support refers to the implemented benchmark interface, including
conditional or adaptation routes.  Methods with the same configuration share
one cell.

\begingroup
\newcommand{\appcapyes}{\tikz[baseline=-0.6ex]{\node[circle,fill=green!9!white,inner sep=1.5pt]{\textcolor{green!40!black}{\scriptsize$\checkmark$}};}}
\newcommand{\appcapno}{\tikz[baseline=-0.6ex]{\node[circle,fill=red!6!white,inner sep=1.5pt]{\textcolor{red!50!black}{\scriptsize$\times$}};}}
\small
\setlength{\tabcolsep}{4pt}
\renewcommand{\arraystretch}{1.2}
\begin{longtable}{@{}p{0.77\linewidth}ccc@{}}
\caption{Selected methods grouped by benchmark capability. Each method is cited separately; symbols indicate supported and unsupported interfaces.}
\label{tab:app-method-capabilities}\\
\toprule
\textbf{Methods} & \textbf{G} & \textbf{I} & \textbf{P}\\
\midrule
\endfirsthead
\multicolumn{4}{l}{\small Table~\thetable{} (continued)}\\
\toprule
\textbf{Methods} & \textbf{G} & \textbf{I} & \textbf{P}\\
\midrule
\endhead
\midrule
\multicolumn{4}{r}{\footnotesize Continued on next page}\\
\endfoot
\bottomrule
\endlastfoot
ForestDiffusion-VP (joint)~\citep{forestdiffusion}; Gaussian copula~\citep{sdv}; GReaT~\citep{great}; LimiX-2~\citep{limix2}; TabDDPM~\citep{tabddpm}; TabDiff~\citep{tabdiff}; TabICLv2~\citep{tabiclv2}; TabPFN-3 + unsupervised~\citep{tabpfn3,tabpfn_extensions}; TabSyn~\citep{tabsyn}. & \appcapyes & \appcapyes & \appcapyes \\[3pt]
ForestDiffusion-VP (label-conditioned)~\citep{forestdiffusion}. & \appcapyes & \appcapyes & \appcapno \\[3pt]
ARF~\citep{arf}; Bayesian network~\citep{synthcity}; CTGAN~\citep{ctgan}; Independent marginals$^{\ddagger}$; NFlow~\citep{neuralsplineflows,synthcity}; NRGBoost~\citep{nrgboost}; SCM (DirectLiNGAM)~\citep{directlingam}; SMOTE~\citep{smote}; SMOTE-NC~\citep{smote}; SMOTER-style interpolation~\citep{smoter}; TabEBM~\citep{tabebm}; Training-row bootstrap~\citep{efron1979bootstrap}; TVAE~\citep{ctgan}. & \appcapyes & \appcapno & \appcapno \\[3pt]
DiffPuter~\citep{diffputer}; HyperImpute~\citep{hyperimpute}; kNN imputation~\citep{knnimpute,pedregosa2011scikit}; Mean/mode imputation~\citep{pedregosa2011scikit}; Median/mode imputation~\citep{pedregosa2011scikit}; MICE$^{\dagger}$~\citep{mice}; MissForest$^{\dagger}$~\citep{missforest}; ReMasker~\citep{remasker}; TabCSDI~\citep{tabcsdi}; TabImpute~\citep{tabimpute}. & \appcapno & \appcapyes & \appcapno \\[3pt]
CARD~\citep{card}; CatBoost~\citep{catboost}; FT-Transformer~\citep{fttransformer}; LightGBM~\citep{lightgbm}; LightGBM-Quantile~\citep{lightgbm}; Logistic regression~\citep{pedregosa2011scikit}; MLP~\citep{pedregosa2011scikit}; NGBoost~\citep{ngboost}; Random forest~\citep{randomforest}; Ridge regression~\citep{hoerl1970ridge,pedregosa2011scikit}; TabM~\citep{tabm}; XGBoost~\citep{xgboost}. & \appcapno & \appcapno & \appcapyes \\[3pt]
\end{longtable}
\endgroup
\noindent\footnotesize
$^{\dagger}$ MICE and MissForest denote the benchmark implementations.
$^{\ddagger}$ Independent marginals is a benchmark-defined control
that samples each column independently from its empirical training marginal
(\texttt{methods/independent\_marginal/adapter.py}).
The analysis alias map assigns historical TabICL outputs to the displayed
TabICLv2 identity.
\normalsize

\subsection{Quality metric collection}
\label{app:quality-metrics}

The tables summarize each metric.  Each stored key consists of a family
prefix and a leaf name.  Arrows indicate the ranking preference:
$\uparrow$ larger, $\downarrow$ smaller, and $\rightarrow a$ closer to $a$.
Superscript D denotes a paired replicate diagnostic.  Privacy diagnostics do
not constitute privacy guarantees.

\par\ifdim\dimexpr\pagegoal-\pagetotal\relax<125pt\relax\newpage\fi
\subsubsection{Generation}
\label{app:quality-generation}
Metrics compare real and synthetic tables.

\begingroup
\small
\setlength{\tabcolsep}{0pt}
\renewcommand{\arraystretch}{1.13}
\setlength{\LTpre}{6pt}
\setlength{\LTpost}{8pt}
\begin{longtable}{@{}>{\raggedright\arraybackslash}p{0.285\linewidth}@{\hspace{1em}}>{\raggedright\arraybackslash}p{0.685\linewidth}@{}}
\caption{Generation quality metric collection (20 metrics). }\label{tab:app-quality-generation}\\
\toprule
\textbf{Metric (exact leaf)} & \textbf{Core meaning}\\
\midrule
\endfirsthead
\multicolumn{2}{l}{\small Table~\thetable{} (continued)}\\
\toprule
\textbf{Metric (exact leaf)} & \textbf{Core meaning}\\
\midrule
\endhead
\midrule
\multicolumn{2}{r}{\footnotesize Continued on next page}\\
\endfoot
\bottomrule
\endlastfoot
\multicolumn{2}{@{}p{\linewidth}@{}}{\footnotesize\textbf{\texttt{generation.\allowbreak{}fidelity.\allowbreak{}*}}~\citep{sdmetrics2026documentation}}\\*[4pt]
{\ttfamily\scriptsize categorical\_\allowbreak{}tvd\_\allowbreak{}error} & $\downarrow$ Mismatch in categorical marginal distributions.\\[2pt]
{\ttfamily\scriptsize mixed\_\allowbreak{}association\_\allowbreak{}error} & $\downarrow$ Mismatch in pairwise association strengths across mixed column types.\\[2pt]
{\ttfamily\scriptsize numerical\_\allowbreak{}ks\_\allowbreak{}error} & $\downarrow$ Mismatch in numerical marginal distributions.\\[2pt]
{\ttfamily\scriptsize pairwise\_\allowbreak{}trend\_\allowbreak{}error} & $\downarrow$ Mismatch in signed numerical trends and categorical joint distributions.\\[2pt]
{\ttfamily\scriptsize shape\_\allowbreak{}error} & $\downarrow$ Overall numerical and categorical marginal mismatch.\\[2pt]
\addlinespace[3pt]
\multicolumn{2}{@{}p{\linewidth}@{}}{\footnotesize\textbf{\texttt{generation.\allowbreak{}distinguishability.\allowbreak{}*}}~\citep{lopezpaz2017revisiting}}\\*[4pt]
{\ttfamily\scriptsize gbdt\_\allowbreak{}c2st\_\allowbreak{}auroc} & $\rightarrow 0.5$ Real-versus-synthetic discrimination by gradient-boosted trees.\\[2pt]
{\ttfamily\scriptsize logistic\_\allowbreak{}c2st\_\allowbreak{}auroc} & $\rightarrow 0.5$ Real-versus-synthetic discrimination by logistic regression.\\[2pt]
\addlinespace[3pt]
\multicolumn{2}{@{}p{\linewidth}@{}}{\footnotesize\textbf{\texttt{generation.\allowbreak{}query.\allowbreak{}*}}~\citep{mckenna2019graphical}}\\*[4pt]
{\ttfamily\scriptsize one\_\allowbreak{}way\_\allowbreak{}error} & $\downarrow$ Real-versus-synthetic discrepancy in one-column query answers.\\[2pt]
{\ttfamily\scriptsize two\_\allowbreak{}way\_\allowbreak{}error} & $\downarrow$ Real-versus-synthetic discrepancy in two-column query answers.\\[2pt]
{\ttfamily\scriptsize three\_\allowbreak{}way\_\allowbreak{}error} & $\downarrow$ Real-versus-synthetic discrepancy in three-column query answers.\\[2pt]
\addlinespace[3pt]
\multicolumn{2}{@{}p{\linewidth}@{}}{\footnotesize\textbf{\texttt{generation.\allowbreak{}support.\allowbreak{}*}}~\citep{alaa2022faithful}}\\*[4pt]
{\ttfamily\scriptsize alpha\_\allowbreak{}precision} & $\uparrow$ Fidelity of synthetic samples to real-data support.\\[2pt]
{\ttfamily\scriptsize authenticity} & $\uparrow$ Synthetic novelty under the benchmark's real-row-based proxy.\\[2pt]
{\ttfamily\scriptsize beta\_\allowbreak{}recall\_\allowbreak{}coverage} & $\uparrow$ Coverage of real-data support by synthetic samples.\\[2pt]
\addlinespace[3pt]
\multicolumn{2}{@{}p{\linewidth}@{}}{\footnotesize\textbf{\texttt{generation.\allowbreak{}utility.\allowbreak{}*}}~\citep{esteban2017realvalued}}\\*[4pt]
{\ttfamily\scriptsize global\_\allowbreak{}utility\_\allowbreak{}ratio} & $\uparrow$ Synthetic-to-real predictive utility across column targets.\\[2pt]
{\ttfamily\scriptsize tstr\_\allowbreak{}gap} & $\rightarrow 0$ Target-prediction utility gap between synthetic-trained and real-trained models.\\[2pt]
\addlinespace[3pt]
\multicolumn{2}{@{}p{\linewidth}@{}}{\footnotesize\textbf{\texttt{generation.\allowbreak{}memorization.\allowbreak{}*}}}\\*[4pt]
{\ttfamily\scriptsize exact\_\allowbreak{}copy\_\allowbreak{}rate} & $\downarrow$ Frequency of exact training-record copies in synthetic data.\\[2pt]
\addlinespace[3pt]
\multicolumn{2}{@{}p{\linewidth}@{}}{\footnotesize\textbf{\texttt{generation.\allowbreak{}privacy.\allowbreak{}*}}~\citep{carlini2022membership}}\\*[4pt]
{\ttfamily\scriptsize delta\_\allowbreak{}presence} & $\downarrow$ Cluster-level training-versus-synthetic presence discrepancy; a noncanonical benchmark diagnostic.\\[2pt]
{\ttfamily\scriptsize distance\_\allowbreak{}mia\_\allowbreak{}auroc} & $\rightarrow 0.5$ Distance-based discrimination of training members from held-out nonmembers.\\[2pt]
{\ttfamily\scriptsize distance\_\allowbreak{}mia\_\allowbreak{}tpr\_\allowbreak{}at\_\allowbreak{}1pct\_\allowbreak{}fpr} & $\downarrow$ Membership-attack sensitivity at a 1\% false-positive rate.\\[2pt]
{\ttfamily\scriptsize mean\_\allowbreak{}min\_\allowbreak{}euclidean\_\allowbreak{}dcr} & $\uparrow$ Euclidean separation between synthetic records and their nearest training records.\\
\end{longtable}
\endgroup

\par\ifdim\dimexpr\pagegoal-\pagetotal\relax<125pt\relax\newpage\fi
\subsubsection{Imputation}
\label{app:quality-imputation}
Metrics describe masked-feature recovery in the target-observed setting.

\begingroup
\small
\setlength{\tabcolsep}{0pt}
\renewcommand{\arraystretch}{1.13}
\setlength{\LTpre}{6pt}
\setlength{\LTpost}{8pt}
\begin{longtable}{@{}>{\raggedright\arraybackslash}p{0.285\linewidth}@{\hspace{1em}}>{\raggedright\arraybackslash}p{0.685\linewidth}@{}}
\caption{Imputation quality metric collection (10 metrics). }\label{tab:app-quality-imputation}\\
\toprule
\textbf{Metric (exact leaf)} & \textbf{Core meaning}\\
\midrule
\endfirsthead
\multicolumn{2}{l}{\small Table~\thetable{} (continued)}\\
\toprule
\textbf{Metric (exact leaf)} & \textbf{Core meaning}\\
\midrule
\endhead
\midrule
\multicolumn{2}{r}{\footnotesize Continued on next page}\\
\endfoot
\bottomrule
\endlastfoot
\multicolumn{2}{@{}p{\linewidth}@{}}{\footnotesize\textbf{\texttt{imputation.\allowbreak{}point.\allowbreak{}*}}}\\*[4pt]
{\ttfamily\scriptsize categorical\_\allowbreak{}balanced\_\allowbreak{}accuracy} & $\uparrow$ Class-balanced accuracy of recovered categorical values.\\[2pt]
{\ttfamily\scriptsize categorical\_\allowbreak{}macro\_\allowbreak{}f1} & $\uparrow$ Class-balanced precision and recall of recovered categorical values.\\[2pt]
{\ttfamily\scriptsize macro\_\allowbreak{}nmae} & $\downarrow$ Normalized absolute error of numerical reconstruction.\\[2pt]
{\ttfamily\scriptsize macro\_\allowbreak{}nrmse} & $\downarrow$ Normalized numerical reconstruction error, emphasizing larger errors.\\[2pt]
\addlinespace[3pt]
\multicolumn{2}{@{}p{\linewidth}@{}}{\footnotesize\textbf{\texttt{imputation.\allowbreak{}structure.\allowbreak{}*}}}\\*[4pt]
{\ttfamily\scriptsize association\_\allowbreak{}matrix\_\allowbreak{}error} & $\downarrow$ Relative distortion of pairwise dependence.\\[2pt]
{\ttfamily\scriptsize categorical\_\allowbreak{}jsd} & $\downarrow$ Symmetric discrepancy in categorical value distributions.\\[2pt]
{\ttfamily\scriptsize forward\_\allowbreak{}kl} & $\downarrow$ Directional mismatch from true to imputed value distributions.\\[2pt]
{\ttfamily\scriptsize numeric\_\allowbreak{}wasserstein1} & $\downarrow$ Normalized discrepancy in numerical value distributions.\\[2pt]
\addlinespace[3pt]
\multicolumn{2}{@{}p{\linewidth}@{}}{\footnotesize\textbf{\texttt{imputation.\allowbreak{}downstream.\allowbreak{}*}}}\\*[4pt]
{\ttfamily\scriptsize utility\_\allowbreak{}gap} & $\rightarrow 0$ Change in downstream predictive utility after imputation.\\[2pt]
\addlinespace[3pt]
\multicolumn{2}{@{}p{\linewidth}@{}}{\footnotesize\textbf{\texttt{imputation.\allowbreak{}robustness.\allowbreak{}*}}}\\*[4pt]
{\ttfamily\scriptsize missingness\_\allowbreak{}shift\_\allowbreak{}degradation}$^{\mathrm D}$ & $\downarrow$ Worst reconstruction degradation under changing missingness conditions.\\
\end{longtable}
\endgroup

\par\ifdim\dimexpr\pagegoal-\pagetotal\relax<125pt\relax\newpage\fi
\subsubsection{Prediction}
\label{app:quality-prediction}
The collection distinguishes classification (10 metrics) from point regression (5 metrics).

\begingroup
\small
\setlength{\tabcolsep}{0pt}
\renewcommand{\arraystretch}{1.13}
\setlength{\LTpre}{6pt}
\setlength{\LTpost}{8pt}
\begin{longtable}{@{}>{\raggedright\arraybackslash}p{0.285\linewidth}@{\hspace{1em}}>{\raggedright\arraybackslash}p{0.685\linewidth}@{}}
\caption{Prediction quality metric collection (15 metrics). }\label{tab:app-quality-prediction}\\
\toprule
\textbf{Metric (exact leaf)} & \textbf{Core meaning}\\
\midrule
\endfirsthead
\multicolumn{2}{l}{\small Table~\thetable{} (continued)}\\
\toprule
\textbf{Metric (exact leaf)} & \textbf{Core meaning}\\
\midrule
\endhead
\midrule
\multicolumn{2}{r}{\footnotesize Continued on next page}\\
\endfoot
\bottomrule
\endlastfoot
\multicolumn{2}{@{}p{\linewidth}@{}}{\footnotesize\textbf{\texttt{prediction.\allowbreak{}classification.\allowbreak{}*}}}\\*[4pt]
{\ttfamily\scriptsize accuracy} & $\uparrow$ Fraction of correctly classified examples.\\[2pt]
{\ttfamily\scriptsize average\_\allowbreak{}precision} & $\uparrow$ Precision-recall quality of probability-based class rankings.\\[2pt]
{\ttfamily\scriptsize balanced\_\allowbreak{}accuracy} & $\uparrow$ Classification accuracy weighted equally across classes.\\[2pt]
{\ttfamily\scriptsize macro\_\allowbreak{}f1} & $\uparrow$ Balance of precision and recall, weighted equally across classes.\\[2pt]
{\ttfamily\scriptsize mcc} & $\uparrow$ Correlation between predicted and true class labels.\\[2pt]
{\ttfamily\scriptsize roc\_\allowbreak{}auc} & $\uparrow$ Class discrimination across decision thresholds.\\[2pt]
\addlinespace[3pt]
\multicolumn{2}{@{}p{\linewidth}@{}}{\footnotesize\textbf{\texttt{prediction.\allowbreak{}calibration.\allowbreak{}*}}~\citep{geifman2019bias,gneiting2007strictly,guo2017calibration}}\\*[4pt]
{\ttfamily\scriptsize aurc} & $\downarrow$ Prediction error as low-confidence cases are progressively withheld.\\[2pt]
{\ttfamily\scriptsize brier\_\allowbreak{}score} & $\downarrow$ Squared discrepancy between class probabilities and observed outcomes.\\[2pt]
{\ttfamily\scriptsize ece} & $\downarrow$ Mismatch between predictive confidence and observed accuracy.\\[2pt]
{\ttfamily\scriptsize log\_\allowbreak{}loss} & $\downarrow$ Probabilistic prediction error that penalizes confidently incorrect predictions.\\[2pt]
\addlinespace[3pt]
\multicolumn{2}{@{}p{\linewidth}@{}}{\footnotesize\textbf{\texttt{prediction.\allowbreak{}regression.\allowbreak{}*}}}\\*[4pt]
{\ttfamily\scriptsize mae} & $\downarrow$ Average absolute prediction error in target units.\\[2pt]
{\ttfamily\scriptsize median\_\allowbreak{}absolute\_\allowbreak{}error} & $\downarrow$ Typical absolute prediction error, less sensitive to outliers.\\[2pt]
{\ttfamily\scriptsize nrmse} & $\downarrow$ Prediction error relative to training-target variability.\\[2pt]
{\ttfamily\scriptsize r2} & $\uparrow$ Prediction quality relative to a test-mean baseline.\\[2pt]
{\ttfamily\scriptsize rmse} & $\downarrow$ Prediction error in target units, emphasizing large residuals.\\
\end{longtable}
\endgroup

%% file: sections/c_metric_space.tex
\section{Metric-Space Audit: Protocol and Results}
\label{app:metric-audit}

Section~\ref{sec:metrics} defines the common-support rank protocol and the
criteria for retaining capability axes.  Table~\ref{tab:metric-groups}
in the main text specifies their exact metric membership.  This appendix
describes the audit protocol and task-specific evidence for these choices.
Appendix~\ref{app:quality-metrics} defines the metrics displayed in the two rank-correlation figures.

For both generation and imputation, the mixed-type association matrix uses
$|r|$ for numerical--numerical pairs, Cram\'er's $V$ for
categorical--categorical pairs~\citep{cramer1946mathematical}, and the
correlation ratio $\eta$ for mixed pairs~\citep{pearson1905skew}.  The two
tasks use different scalar summaries of this shared representation:
generation reports the mean absolute difference over comparable off-diagonal
pairs and its complement, whereas imputation reports the relative Frobenius
distance.  The metrics compare the same pairwise dependence representation but are not
numerically interchangeable.

\subsection{Audit protocol: common-support ranks before correlations}

After constructing the ranks in Equation~\ref{eq:metric-composite}, the
pairwise audit retains only cells in which both quantities are observed,
discards strata in which either metric is constant, requires at least three
methods per stratum and at least 30 paired cells over six datasets, and
computes the Pearson correlation of the pooled rank pairs (the
Spearman-equivalent construction used by the audit).  The catalog sets
$a_k=0.5$ for C2ST and distance-MIA AUROC and $a_k=0$ for utility gaps.  The
catalog marks normalized median Gower DCR as neutral; analyses that require a
direction use the explicit provisional target-one transform $-|x-1|$ and do
not promote it to confirmatory privacy evidence.  For the spectral audit, $c$
is the fixed metric cohort.
The strata are dataset for generation, dataset--missingness setting--rate--
contract for imputation, and dataset--task type for prediction.  Confidence
intervals use 500 bootstrap resamples~\citep{efron1979bootstrap} with whole
datasets as the resampling units.

\begin{figure}[htbp]
    \centering
    \includegraphics[width=\textwidth]{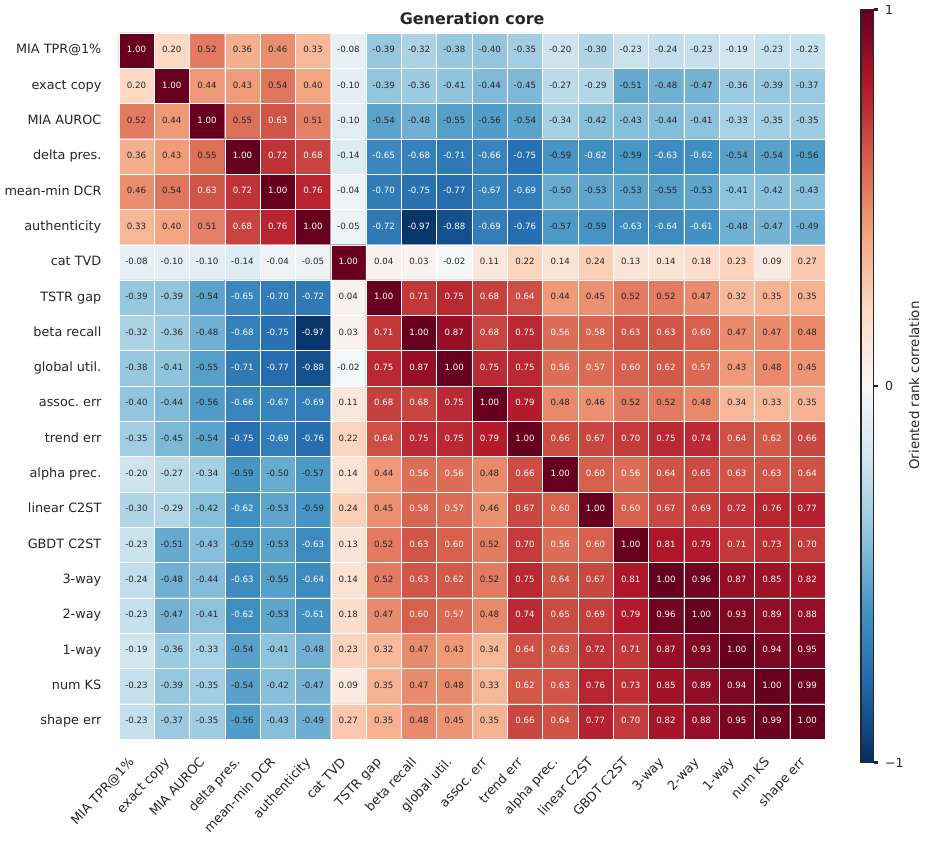}
    \caption{\textbf{Generation rank-correlation geometry (main metric-audit
    view).}  Each cell reports the Pearson correlation $\rho$ of common-support,
    within-dataset percentile ranks after orienting values so that larger is
    better, rounded to two decimals.  Red denotes aligned method orderings and
    blue denotes trade-offs; average-linkage ordering uses distance $1-\rho$.
    The map shows a dense marginal/query-fidelity block, while
    utility and $\beta$-recall align with one another but oppose authenticity
    and distance-to-training diagnostics.  White denotes correlation near
    zero.}
    \label{fig:metric-correlations-generation}
\end{figure}

\begin{figure}[htbp]
    \centering
    \includegraphics[width=\textwidth]{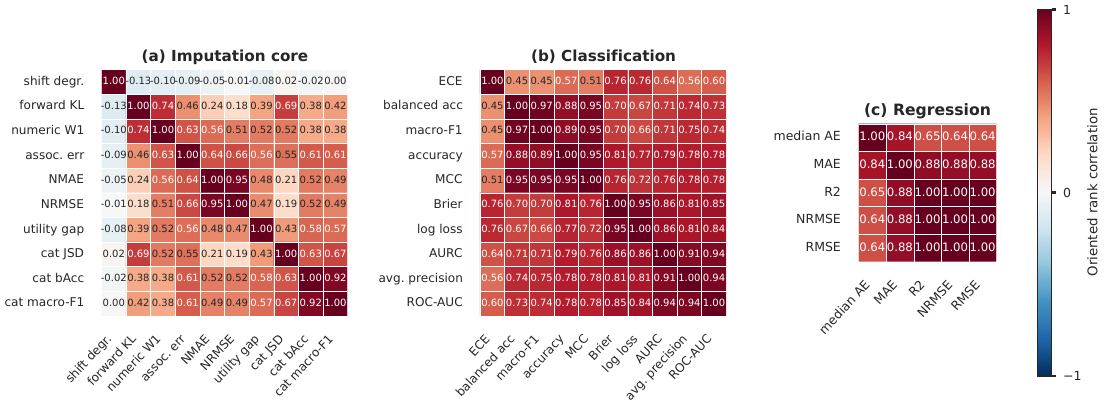}
    \caption{\textbf{Rank-correlation geometry for imputation and prediction.}
    Cell values report oriented rank correlation $\rho$, rounded to two
    decimals, under the same common-support construction and color scale as
    Figure~\ref{fig:metric-correlations-generation}.  Imputation contains tight
    numerical- and categorical-point-accuracy pairs but separates
    distributional, association, and utility quantities.  Classification
    retains distinct hard-decision, proper-score, ranking/selective-risk, and
    calibration views despite broad shared signal.  Regression exhibits an
    almost rank-equivalent squared-error block while retaining distinct
    absolute-error summaries.}
    \label{fig:metric-correlations-imputation-prediction}
\end{figure}

For relationships with finite confidence intervals, we call a pair a
\emph{redundancy candidate} only when its oriented correlation satisfies
$\rho\geq0.85$ and the 95\% dataset-bootstrap lower bound is at least $0.70$.
Conversely, $\rho\leq-0.50$ with an upper bound at most $-0.20$ marks a stable
trade-off.  These thresholds inform, but do not replace, semantic judgment.
For the cluster counts in Table~\ref{tab:metric-redundancy-summary}, we apply
average linkage to $1-|\rho|$ and cut at $0.15$.  This absolute-correlation
clustering detects both duplicate orderings and two-sided trade-offs; only the
former can be pruned.  The heatmap ordering instead uses signed distance
$1-\rho$ so that opposing directions remain visually separate.
The orderings in
Figures~\ref{fig:metric-correlations-generation} and
\ref{fig:metric-correlations-imputation-prediction} display the empirical
blocks; adjacency alone does not justify merging metrics.

Figures~\ref{fig:metric-correlations-generation} and
\ref{fig:metric-correlations-imputation-prediction} show the larger generation metric space separately from imputation and
prediction to keep individual rank correlations legible.

\subsection{From correlation blocks to redundancy clusters}

Figures~\ref{fig:metric-correlations-generation} and
\ref{fig:metric-correlations-imputation-prediction} inform metric selection.
Within each task, a positive block supports pooling
only when its metrics share an estimand; weakly aligned metrics are not pruned
and can share a named construct only as reportable complementary probes;
mixed-sign blocks are retained as explicit trade-offs.  The following
decisions account for each pooled family and retained direction in
Table~\ref{tab:metric-groups}.

\paragraph{Generation.}
The full 29-metric quality audit yields 17 local correlation clusters, the
largest number among the three tasks.  Figure~\ref{fig:metric-correlations-generation}
shows the fixed 20-metric core; the full audit additionally evaluates alternate
parameterizations and validity guards.  The five marginal/query-fidelity metrics form
the clearest positive block: pairwise correlations range from $0.823$ to
$0.990$, and eight of ten pairs pass the conservative redundancy rule.  An
equivalent shape metric brings the cluster to six.
By~contrast,~the~two~C2ST~variants are only moderately
aligned ($\rho=0.599$), so linear and nonlinear distinguishability remain
separate, reportable probes within the named C2ST axis rather than statistical
substitutes.  We pool the dense marginal/query block into \emph{fidelity}, but
keep $\alpha$-precision, association recovery, and TSTR separate because
support, dependence, and transfer are not interchangeable with marginal
fidelity.  Four removable generation metrics are exact alternative
parameterizations; semantic review identifies only five additional metrics
as substantive reductions.

A \emph{signed} cluster requires separate treatment.  Global utility
and $\beta$-recall align ($\rho=0.872$), whereas authenticity opposes both
global utility ($\rho=-0.881$) and $\beta$-recall ($\rho=-0.973$).  Absolute
correlation therefore places them nearby, but they measure opposite ends of a
support--novelty trade-off and each must remain reportable.  The privacy metrics
also resist reduction: mean-min Euclidean DCR has correlations of $0.464$,
$0.630$, and $0.720$ with MIA TPR@1\%, MIA AUROC, and delta presence.  We group
these as \emph{privacy-risk diagnostics} because they provide complementary
views of record proximity, membership inference, and group-presence
disclosure, rather than estimates of one interchangeable quantity.  None is
a privacy guarantee, and exact-copy rate remains a separate memorization
guardrail.  These decisions yield the seven generation axes in
Table~\ref{tab:metric-groups}.

\paragraph{Imputation.}
The imputation panel of
Figure~\ref{fig:metric-correlations-imputation-prediction} contains ten quality
metrics in eight clusters, six of them singletons.  The only stable
substitutions are numerical NMAE--NRMSE ($\rho=0.947$, 95\% CI
$[0.931,0.959]$) and categorical balanced accuracy--macro-F1 ($\rho=0.925$,
$[0.899,0.949]$).  Point accuracy therefore does not subsume distribution
matching, association recovery, or downstream utility.  Missingness-shift
degradation is nearly orthogonal to all level metrics, but constant imputers
can appear artificially robust because their overall performance is already at
a floor; we retain it as a paired diagnostic rather than a standalone ranking
axis.  Draw-level quantities likewise remain diagnostic rather than entering
the reduced headline registry.  Accordingly, Table~\ref{tab:metric-groups}
averages only the two point-recovery pairs and retains distribution,
association, and utility as separate headline axes.

\paragraph{Prediction.}
The prediction panels in
Figure~\ref{fig:metric-correlations-imputation-prediction} show four observed
clusters among the ten classification metrics.  Hard-decision correlations
range from $0.882$ to $0.971$; Brier score and log loss correlate at $0.951$;
and correlations among ROC--AUC, average precision, and AURC range from $0.909$ to $0.943$.
Accordingly, Table~\ref{tab:metric-groups} pools hard decisions, proper scores,
and ranking/selective risk within their respective axes, while ECE remains a
singleton because calibration is not equivalent to discrimination, proper
scoring, or selective prediction.  In regression, $R^2$, NRMSE, and RMSE
induce almost identical method orderings ($\rho>0.999$), MAE joins this block at
the $0.85$ cut, and median absolute error remains separate.  We nevertheless
report MAE independently because mean absolute and squared error encode
different loss functionals.  Cluster membership indicates similar rankings
but does not make the estimands interchangeable.  This yields four classification and
three regression axes in Table~\ref{tab:metric-groups}.

\subsection{Construct-specific stress tests}

Controlled synthetic examples illustrate how selected metrics respond to
specific perturbations.
Independently shuffling columns preserves their marginal distributions while
disrupting associations; replaying training rows can improve apparent fidelity
while worsening memorization diagnostics; marginal-only imputation can
approximate univariate distributions without accurately recovering individual
entries or joint structure; and monotone probability rescaling can alter
calibration without changing ranking.  In these examples, marginal fidelity,
dependence, memorization, point recovery, and calibration metrics respond
differently to the imposed perturbations.

\subsection{Reliability qualifies every redundancy claim}

\begin{figure}[htbp]
    \centering
    \includegraphics[width=\textwidth]{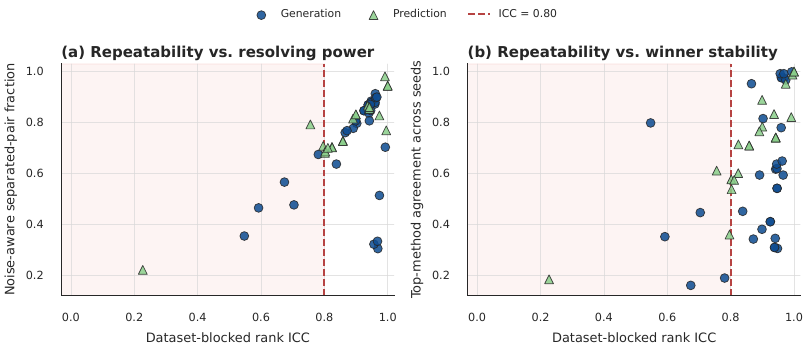}
    \caption{\textbf{Reliability and operational discrimination (auxiliary
    metric-audit view).}  Each point is a direction-defined, informative catalog
    metric from generation or prediction with at least 20 repeated cells across
    at least eight datasets.  The dashed line is a visual reference at rank
    ICC $=0.80$; the shaded region lies below it.  Left: the fraction of
    repeated method pairs separated by more than $1.96$ times the seed-level
    standard error of their difference.  Right: agreement of top-method sets
    across replicate pairs.  Repeatability, pairwise
    resolving power, and winner stability are related but non-equivalent
    properties.}
    \label{fig:metric-reliability}
\end{figure}

Measurement noise can obscure correlation structure or create apparent
structure.  Figure~\ref{fig:metric-reliability} therefore reports dataset-
blocked rank ICC~\citep{shrout1979intraclass} alongside two operational quantities.  The separated-pair
fraction counts method pairs whose replicate-mean difference exceeds $1.96$
times the seed-level standard error of the difference; this threshold describes resolving power
and is not a significance test.  Top-method agreement is the mean Jaccard overlap of
the best-method sets across replicate pairs.  These summaries measure different properties: even a highly repeatable
metric may identify different winners when several methods perform similarly.  Among selected direction-defined generation
components, rank ICCs range from $0.592$ to
$0.963$.  MIA TPR@1\% and MIA AUROC form the low end at $0.592$ and $0.674$,
and winner agreement can be as low as $0.161$.  These values support
retaining component-level privacy diagnostics alongside their summary.
For classification, correcting correlations by the three-seed test--retest
reliabilities merges hard-decision, proper-score, and ranking metrics at the
same $0.85$ cut, while ECE remains separate.  We treat this correction as a sensitivity
analysis rather than a mechanical selection rule: it supports one broad
predictive-quality factor plus a distinct calibration direction.

\subsection{Compressed, but not one-dimensional}

The PCA audit~\citep{jolliffe2002pca} uses each task's full quality cohort.  The Kaiser rule~\citep{kaiser1960application} retains
five generation components
(80\% cumulative variance; PC1 53\%), three imputation components (74\%; 41\%),
two classification components (88\%; 75\%), and one regression component
(86\%; 86\%).  These counts are smaller than the corresponding
$17/8/4/2$ local clusters because PCA captures shared variation rather than
pairwise substitutability.  A dominant quality factor can
coexist with a scientifically important low-variance direction such as
calibration or typical absolute error.

\begin{figure}[htbp]
    \centering
    \includegraphics[width=\textwidth]{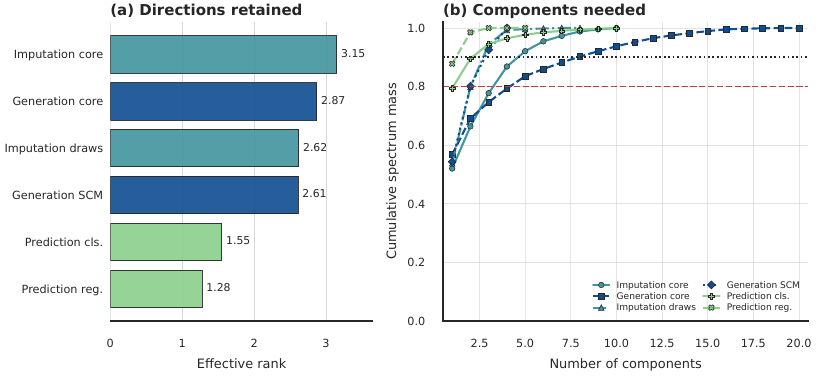}
    \caption{\textbf{Residual effective dimensionality (auxiliary
    metric-audit view).}  Left: participation-ratio effective rank,
    $(\sum_j\lambda_j)^2/\sum_j\lambda_j^2$, using the nonnegative-clipped
    correlation spectrum.  Right: cumulative spectrum mass, with 80\% and
    90\% guides.  A cohort is shown only when every metric pair satisfies the
    common-support criteria stated above.  ``Imputation core'' and
    ``Imputation draws'' use the target-observed contract; feature-only
    imputation is analyzed separately.  The eight-metric draws spectrum
    summarizes additional distributional metrics.  Generation SCM reports
    four structural/association quantities for the six SCM datasets,
    separately from the twenty-metric generation core.  The curves summarize
    within-cohort compression.}
    \label{fig:metric-dimensions}
\end{figure}

Figure~\ref{fig:metric-dimensions} shows a complementary spectrum for the fixed
core cohorts used in the heatmaps.  The
participation-ratio effective rank is $3.15$ for the ten imputation-core
metrics and $2.87$ for the twenty generation-core metrics; reaching 90\%
spectrum mass requires five and eight components, respectively.
Classification and point regression are more compressed ($1.55$ and $1.28$),
consistent with the tight blocks in
Figure~\ref{fig:metric-correlations-imputation-prediction}.  These values
describe a correlation spectrum, not performance.  Because pairwise
correlations can use different common-support cells, the assembled matrix need
not be positive semidefinite; negative eigenvalues are clipped only for this
descriptive spectrum.  Compression does not imply one-dimensionality: the retained semantic axes preserve stable trade-offs and
distinct estimands that a purely spectral summary would blend.

Using the retained axis counts $(7,6,4,3)$ only as dimensionality budgets, the
corresponding leading eigenspaces retain $88.22\%$, $95.46\%$, $96.49\%$, and
$99.99\%$ of the nonnegative spectrum for generation, imputation,
classification, and regression.  These are optimal same-dimensional spectral
upper bounds, not the mass captured by the particular semantic composites.
For the latter, let $R_+$ denote the nonnegative-eigenvalue reconstruction of
the correlation matrix and let $W$ encode the equal-weight composite members.
We report the reconstruction fraction
\begin{equation}
 \mathcal{C}(W)=
 \frac{\operatorname{tr}\!\left[R_+W(W^\top R_+W)^+W^\top R_+\right]}
      {\operatorname{tr}(R_+)},
 \label{eq:composite-spectrum-retention}
\end{equation}
where $(\cdot)^+$ is the Moore--Penrose inverse.  The final composite spans
retain $79.65\%$, $86.21\%$, $96.40\%$, and $99.99\%$, respectively.  The
generation value is necessarily partial: normalized median Gower DCR is
neutral and therefore absent from $R_+$, so only the direction-defined members
of the privacy-risk diagnostics axis enter this projection.

%% file: sections/d_task_rank_analysis.tex
\section{Task Rank Analysis Details}
\label{app:task-rank-analysis}

\newcommand{\trfigdir}{figures/appendix_task_rank_analysis}
\newcommand{\trstabilitycaption}[1]{Axis-pair LOMO diagnostics for #1. From upper left to lower right, the panels show the pooled Spearman correlation for the full method pool, the median correlation across valid, effective single-method deletions, the LOMO span, and the fraction of effective deletions preserving the full-pool sign. Red and blue denote positive and negative correlations; sequential blue encodes the latter two diagnostics. A dagger (\(^{\dagger}\)) marks axis pairs for which at least one effective deletion changes sign or produces zero correlation. Each deletion requires realigning support, reranking within datasets, and recomputing axis scores; no-op deletions are excluded from the effective denominator. LOMO ranges are sensitivity diagnostics, not confidence intervals.}
\newcommand{\trdetailcaptionfirst}[1]{Complete leave-one-method-out results for #1 (part 1 of 2). The ``Full method pool'' panel is the baseline; every other panel removes the named method. Each cell reports the pooled Spearman correlation. Red and blue denote positive and negative correlations. Support alignment, within-dataset ranks, and capability-axis scores are recomputed for every panel.}
\newcommand{\trdetailcaptionsecond}[2]{Complete leave-one-method-out results for #1 (part 2 of 2), continuing Figure~\ref{#2}. The encoding and recomputation procedure are unchanged. A panel can match the baseline when the named deletion is a no-op for that axis pair; these deletions are excluded from the sign-agreement denominator.}

This appendix details the procedure and aligned support for the cross-task rank analysis in Section~\ref{sec:task_rank_analysis}.
We compare method rankings across generation (G), imputation (I), classification prediction (P-cls), and regression prediction (P-reg).
For each task pair, we restrict the analysis to the common method pool and recompute within-dataset ranks on the aligned support.
This controls for differences in method availability, dataset difficulty, and metric scale.

\subsection{Task Settings and Shared Support}

\paragraph{Input units.}
The input table for each task is indexed by $(d,m)$, where $d$ denotes a
dataset and $m$ denotes a method.  Each column contains one evaluation metric
for that task.  Raw records are first grouped using the declared
seed/replicate fields and then averaged after metric orientation.
Prediction datasets are separated into P-cls and P-reg according to task type.  For imputation, we use
the target-observed MCAR setting with a missing rate of $0.3$.

\paragraph{Metric direction.}
We orient metrics with a clear optimization direction so that higher values indicate better performance.
We negate metrics for which lower values are better, use $-|x|$ for metrics whose ideal value is $0$, and use $-|x-0.5|$ for metrics whose ideal value is $0.5$.
Neutral diagnostics retain their stored direction.
For these diagnostics, a correlation sign describes the relation to the stored direction; it does not by itself indicate better or worse performance.

\paragraph{Common method pool.}
For tasks $a$ and $b$, the common method pool is defined as $\mathcal{M}_{ab}=\mathcal{M}_a\cap\mathcal{M}_b$.
Classification and regression use the same candidate intersections for I--P and G--P, but are analyzed independently on their respective dataset subsets.
Table~\ref{tab:tra-shared-support} reports the task-family candidate pools.
We align support separately for each axis pair, retaining only dataset--method
units with valid values for all declared members on both sides.

\begin{table}[htbp]
    \centering
    \caption{Candidate method pools and member-complete aligned support for
    each task pair.  ``Datasets'' and $n$ give the axis-pair ranges of
    effective datasets and valid $(\text{dataset},\text{method})$ units,
    respectively.}
    \label{tab:tra-shared-support}
    \small
    \begin{tabularx}{0.99\linewidth}{l c c c>{\raggedright\arraybackslash}X}
        \toprule
        Task pair & Candidate pool & Datasets & Range of $n$ & Candidate methods \\
        \midrule
        I $\times$ P-cls
        & 9 & 13--38 & 92--287
        & Forest-VP Joint, Gaussian Copula, GReaT, LimiX-2, TabDDPM, TabDiff, TabICLv2, TabPFN-3, TabSyn \\
        I $\times$ P-reg
        & 9 & 6--18 & 44--139
        & Forest-VP Joint, Gaussian Copula, GReaT, LimiX-2, TabDDPM, TabDiff, TabICLv2, TabPFN-3, TabSyn \\
        G $\times$ I
        & 10 & 12--56 & 98--482
        & Forest-VP, Forest-VP Joint, Gaussian Copula, GReaT, LimiX-2, TabDDPM, TabDiff, TabICLv2, TabPFN-3, TabSyn \\
        G $\times$ P-cls
        & 9 & 30--38 & 246--300
        & Forest-VP Joint, Gaussian Copula, GReaT, LimiX-2, TabDDPM, TabDiff, TabICLv2, TabPFN-3, TabSyn \\
        G $\times$ P-reg
        & 9 & 18 & 138--141
        & Forest-VP Joint, Gaussian Copula, GReaT, LimiX-2, TabDDPM, TabDiff, TabICLv2, TabPFN-3, TabSyn \\
        \bottomrule
    \end{tabularx}
\end{table}

\subsection{Computing Axis-Pair Correlations}

\paragraph{Cell-wise alignment.}
For capability axes $(c,c')$ from tasks $(a,b)$, let $\mathcal K_c$ and
$\mathcal K_{c'}$ denote their fixed member-metric sets.  We define the
aligned support for this axis pair as
\begin{equation}
\mathcal S_{cc'}^{ab}
=
\left\{(d,m):m\in\mathcal M_{ab},\;
\bigwedge_{k\in\mathcal K_c}v_{dmk}^{(a)}\ \text{is valid},\;
\bigwedge_{\ell\in\mathcal K_{c'}}v_{dm\ell}^{(b)}\ \text{is valid}
\right\}.
\label{eq:tra-pair-support}
\end{equation}
Thus, a $(d,m)$ unit is retained only when all declared members are valid on
both sides.  We construct $\mathcal S_{cc'}^{ab}$ independently for every
axis pair rather than imposing complete support across all retained axes.

\paragraph{Within-dataset percentile ranks.}
Let
$\mathcal M_{d,cc'}^{ab}=\{m:(d,m)\in\mathcal S_{cc'}^{ab}\}$.
Every member metric on both sides is ranked over this same method set:
\begin{equation}
r_{d,m,k}^{(t)}
=
\frac{\operatorname{rank}_{\mathrm{avg}}
\!\left(v_{d,m,k}^{(t)};
\{v_{d,m',k}^{(t)}:m'\in\mathcal M_{d,cc'}^{ab}\}\right)}
{|\mathcal M_{d,cc'}^{ab}|}
\in(0,1].
\label{eq:tra-within-dataset-rank}
\end{equation}
Methods with identical metric values receive the average of the ranks
they occupy. Percentile ranks are normalized by the size of the aligned method
set and therefore lie in $(0,1]$, with the smallest possible value equal to
$1/|\mathcal M_{d,cc'}^{ab}|$.

\paragraph{Axis aggregation.}
Following Equation~\ref{eq:metric-composite}, the axis score is the
equal-weight mean of the declared member percentile ranks:
\begin{equation}
a_{d,m,c}^{(t)}
=
\frac{1}{|\mathcal K_c|}
\sum_{k\in\mathcal K_c} r_{d,m,k}^{(t)}.
\label{eq:tra-axis-implementation}
\end{equation}
The member set is fixed at $\mathcal K_c$, and no additional standardization
is applied after ranking.  Member-complete alignment ensures that each axis
score includes all its declared members.

\paragraph{Pooled Spearman correlation.}
After constructing the axis scores, we compute
\begin{equation}
\rho_{cc'}^{ab}
=
\operatorname{Spearman}
\left(
\{a_{d,m,c}^{(a)}:(d,m)\in\mathcal S_{cc'}^{ab}\},
\{a_{d,m,c'}^{(b)}:(d,m)\in\mathcal S_{cc'}^{ab}\}
\right)
\label{eq:tra-pooled-spearman}
\end{equation}
over all units in $\mathcal S_{cc'}^{ab}$.  This is a single correlation over
the pooled dataset--method units; we do not first compute a correlation within
each dataset and then average correlations across datasets.  A correlation is
left undefined when fewer than 10 paired units remain.

\subsection{Leave-One-Method-Out Influence Diagnostics}

For every method $u\in\mathcal M_{ab}$, define the reduced method pool
$\mathcal M_{ab}^{(-u)}=\mathcal M_{ab}\setminus\{u\}$.  After each removal,
we recompute the aligned support, within-dataset percentile
ranks, axis scores, and final correlation $\rho_{cc'}^{ab,(-u)}$.  A deletion
is a no-op when it changes no valid unit for that axis pair.  Let
$\mathcal E_{cc'}^{ab}$ contain the effective deletions with finite
recomputations.  The LOMO sensitivity range and span are
\begin{equation}
\mathcal R_{cc'}^{ab}
=
\left[
\min_{u\in\mathcal E_{cc'}^{ab}}\rho_{cc'}^{ab,(-u)},
\max_{u\in\mathcal E_{cc'}^{ab}}\rho_{cc'}^{ab,(-u)}
\right],
\qquad
\Delta_{cc'}^{ab}
=\max_{u\in\mathcal E_{cc'}^{ab}}\rho_{cc'}^{ab,(-u)}
-\min_{u\in\mathcal E_{cc'}^{ab}}\rho_{cc'}^{ab,(-u)}.
\label{eq:tra-loo-range}
\end{equation}
An axis pair is called LOMO-direction-preserving when every effective
recomputation has the same sign as its full-pool correlation.  No-op deletions
are excluded from the sign-agreement denominator and the effective deletion
count.  A dagger (\(^{\dagger}\)) marks axis pairs for which an effective
deletion changes sign or produces zero correlation.  The LOMO range is an influence
diagnostic for the composition of the method pool, not a confidence interval
or a significance test.

\subsection{Cross-Task Axis-Level Correlations and LOMO Stability}

\paragraph{Imputation and classification prediction (I $\times$ P-cls).}
Among the 24 axis pairs, 13 are LOMO-direction-preserving.  Numerical point
accuracy, association, and downstream utility gap have positive,
direction-preserving correlations with all four classification axes.
Categorical point accuracy has this relation only with hard-decision
performance; neither distribution axis has a direction-preserving
classification relation.

\begin{figure}[htbp]
    \centering
    \includegraphics[width=0.99\linewidth]{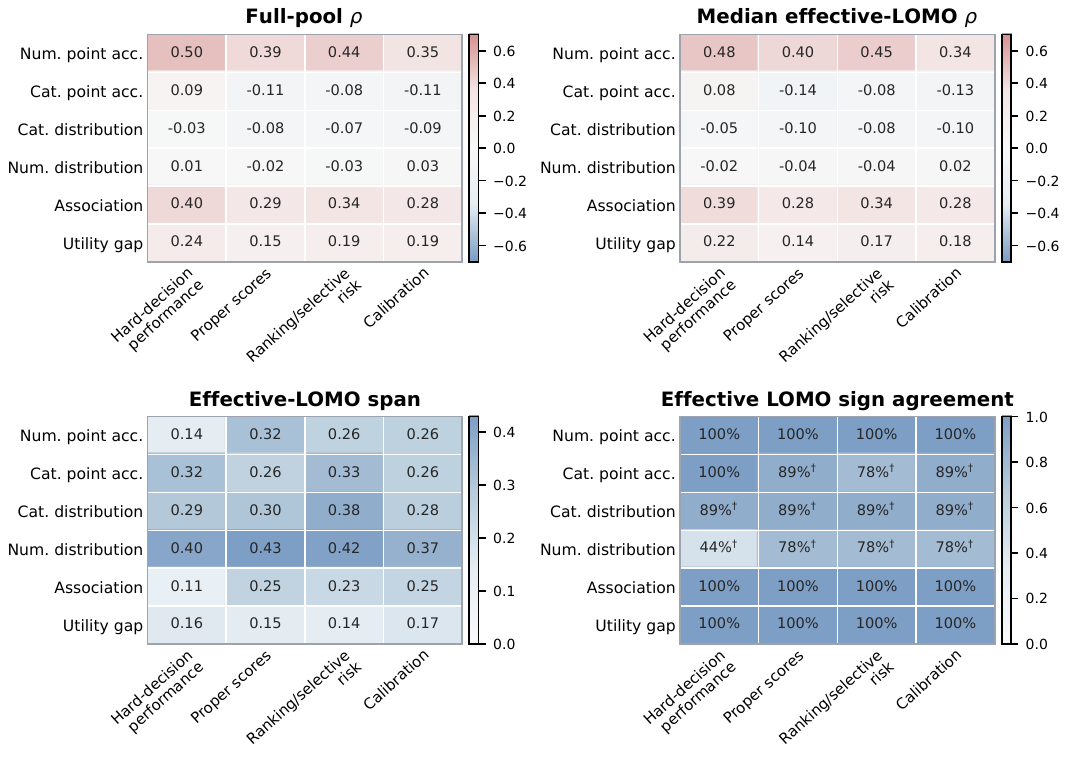}
    \caption{\trstabilitycaption{I $\times$ P-cls}}
    \label{fig:tra-stability-i-pcls}
\end{figure}

\paragraph{Imputation and regression prediction (I $\times$ P-reg).}
Fourteen of the 18 axis pairs are LOMO-direction-preserving.  Numerical point
accuracy, numerical distribution, association, and downstream utility gap have
positive, direction-preserving correlations with all three regression axes.
Categorical point accuracy also preserves a positive direction with
squared error and mean absolute error, but not median error; categorical
distribution is direction-unstable for all three.

\begin{figure}[htbp]
    \centering
    \includegraphics[width=0.99\linewidth]{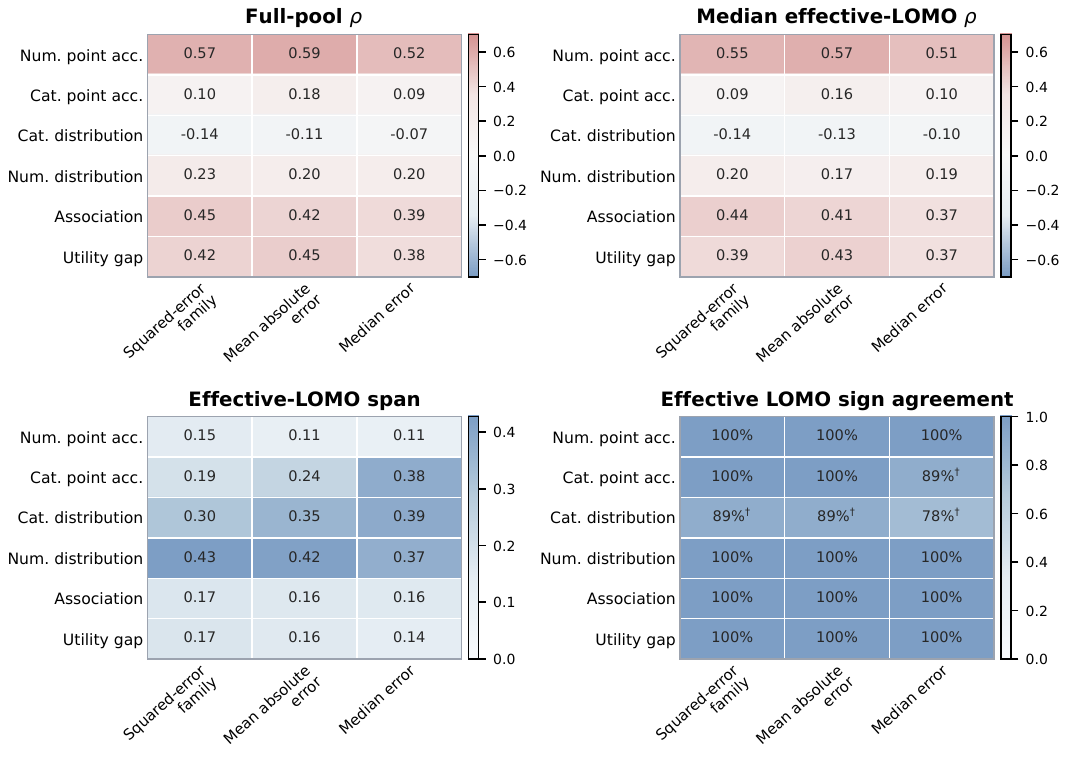}
    \caption{\trstabilitycaption{I $\times$ P-reg}}
    \label{fig:tra-stability-i-preg}
\end{figure}

\paragraph{Generation and imputation (G $\times$ I).}
Of the 42 G $\times$ I pairs, 36 are LOMO-direction-preserving.  TSTR gap and
global-utility compromise are positive for all six imputation axes.
Privacy-risk diagnostics are negative for five axes; numerical distribution is the exception.  Fidelity preserves direction for five pairs, C2ST for four,
and both $\alpha$-precision and association for five; their signs remain
axis-specific.

\begin{figure}[htbp]
    \centering
    \includegraphics[width=0.99\linewidth]{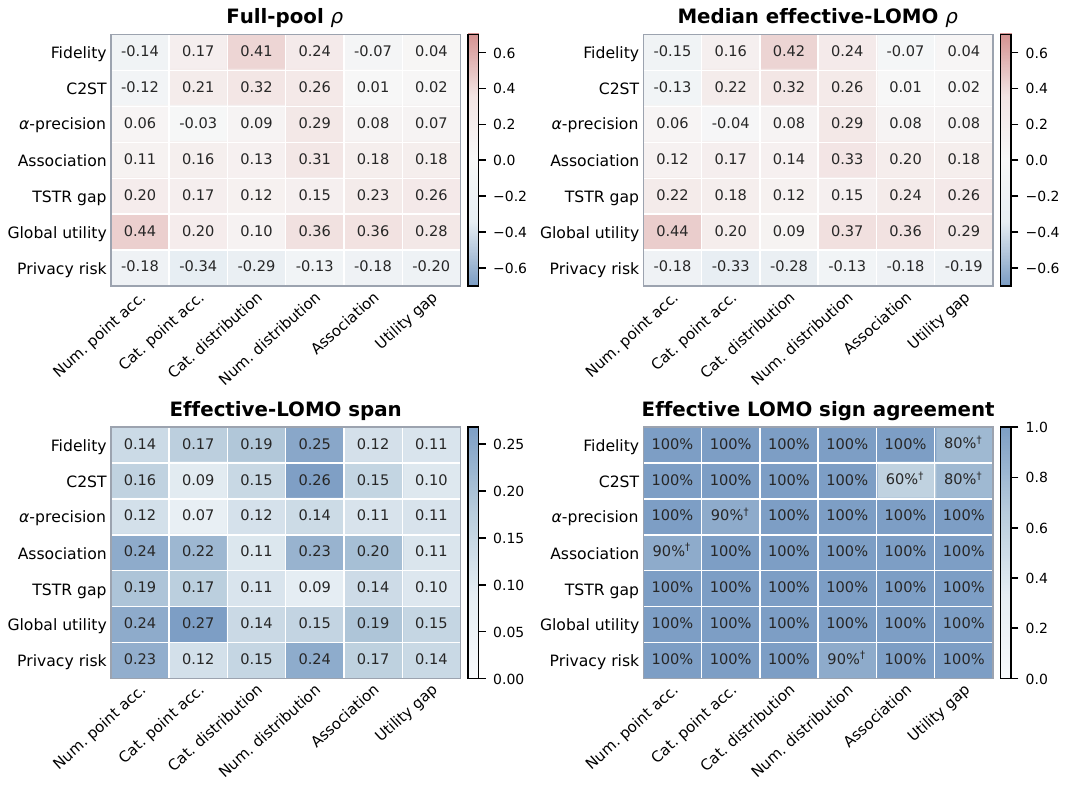}
    \caption{\trstabilitycaption{G $\times$ I}}
    \label{fig:tra-stability-g-i}
\end{figure}

\paragraph{Generation and classification prediction (G $\times$ P-cls).}
Seventeen of the 28 pairs are LOMO-direction-preserving.  Fidelity and C2ST
are negative with all four classification axes, while TSTR gap and
global-utility compromise are positive with all four.  Privacy-risk
diagnostics also have positive, direction-preserving correlations with proper
scores.  All $\alpha$-precision and association pairs are
direction-unstable.

\begin{figure}[htbp]
    \centering
    \includegraphics[width=0.99\linewidth]{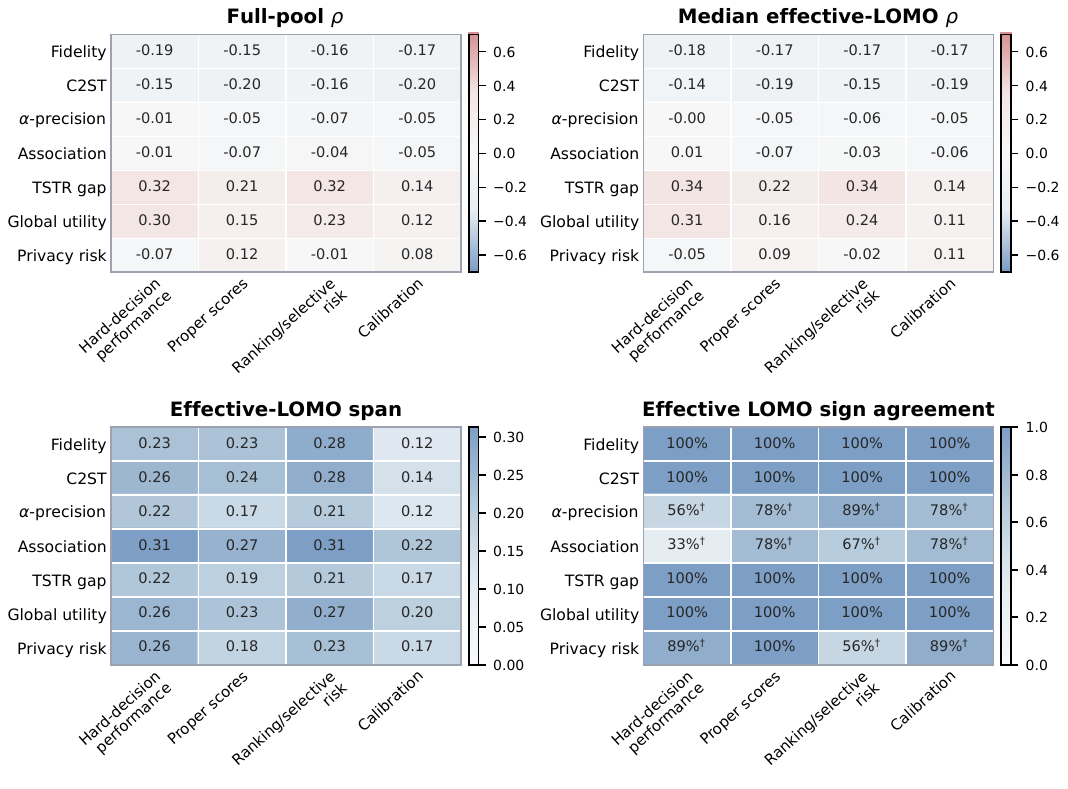}
    \caption{\trstabilitycaption{G $\times$ P-cls}}
    \label{fig:tra-stability-g-pcls}
\end{figure}

\paragraph{Generation and regression prediction (G $\times$ P-reg).}
Nine of the 21 pairs are LOMO-direction-preserving.  TSTR gap and
global-utility compromise are positive with all three regression axes.
Fidelity and C2ST are negative only against squared error, while
$\alpha$-precision is positive only against median error.  Every association
and privacy-risk pair is direction-unstable.

\begin{figure}[htbp]
    \centering
    \includegraphics[width=0.99\linewidth]{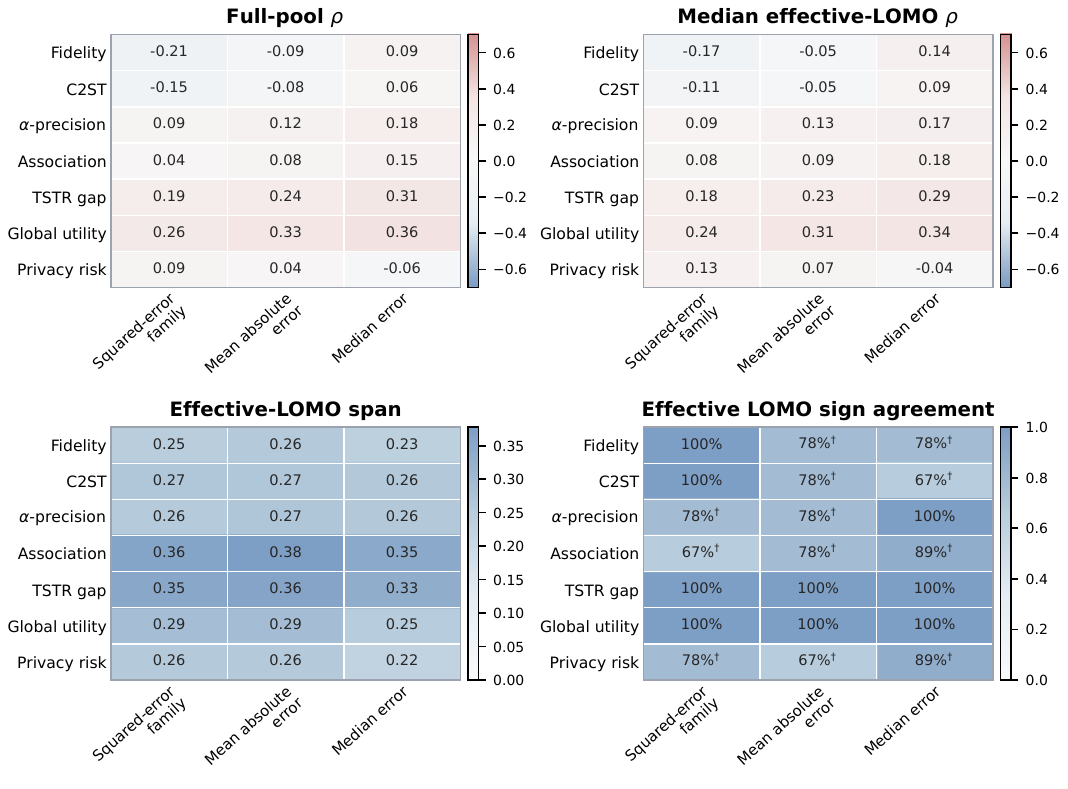}
    \caption{\trstabilitycaption{G $\times$ P-reg}}
    \label{fig:tra-stability-g-preg}
\end{figure}

\subsection{Complete Leave-One-Method-Out Grids}

\begin{figure}[htbp]
    \centering
    \includegraphics[width=\linewidth,height=0.79\textheight,keepaspectratio]{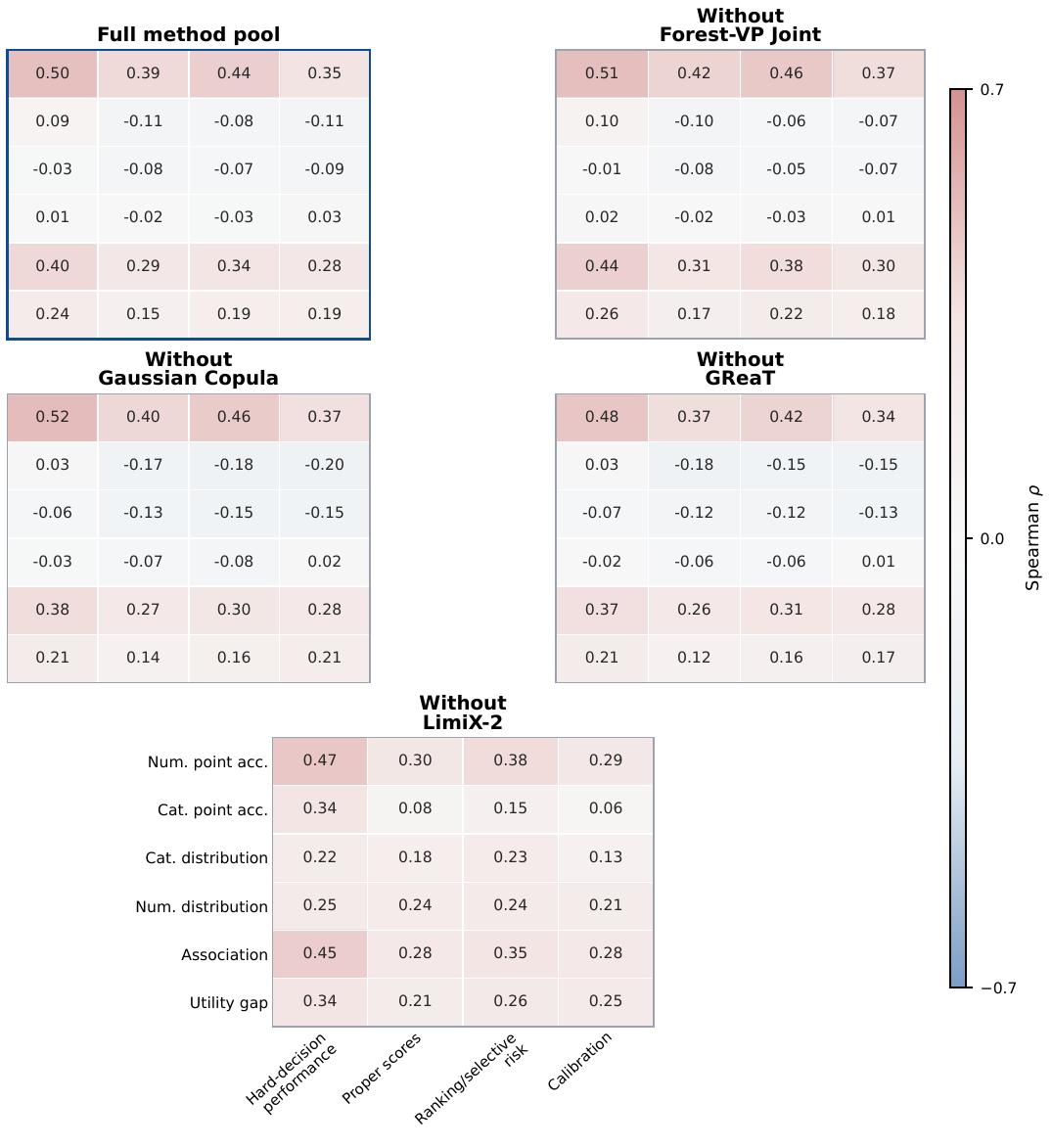}
    \caption{\trdetailcaptionfirst{I $\times$ P-cls}}
    \label{fig:tra-detail-i-pcls-1}
\end{figure}

\begin{figure}[htbp]
    \centering
    \includegraphics[width=\linewidth,height=0.79\textheight,keepaspectratio]{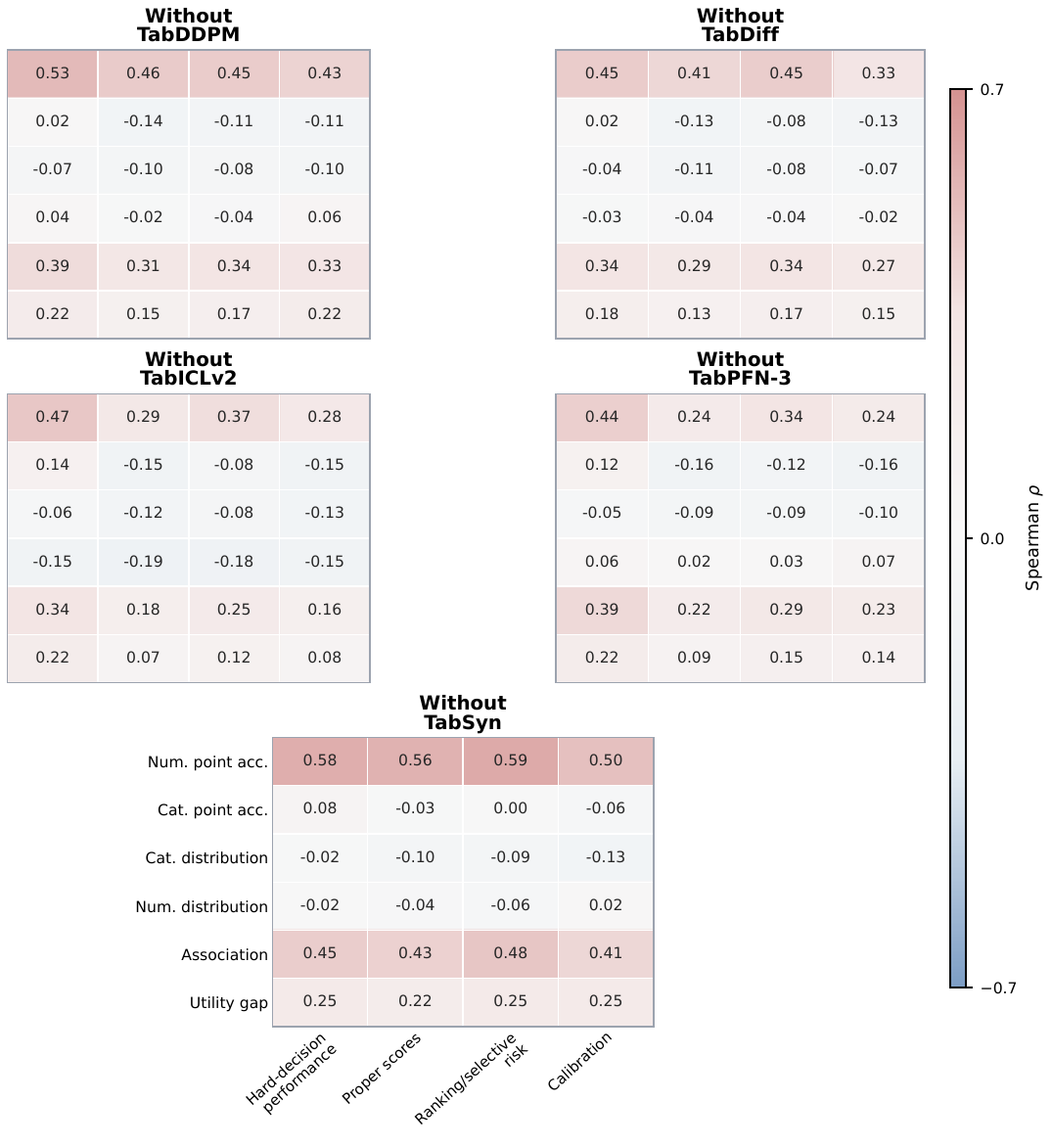}
    \caption{\trdetailcaptionsecond{I $\times$ P-cls}{fig:tra-detail-i-pcls-1}}
    \label{fig:tra-detail-i-pcls-2}
\end{figure}

\begin{figure}[htbp]
    \centering
    \includegraphics[width=\linewidth,height=0.79\textheight,keepaspectratio]{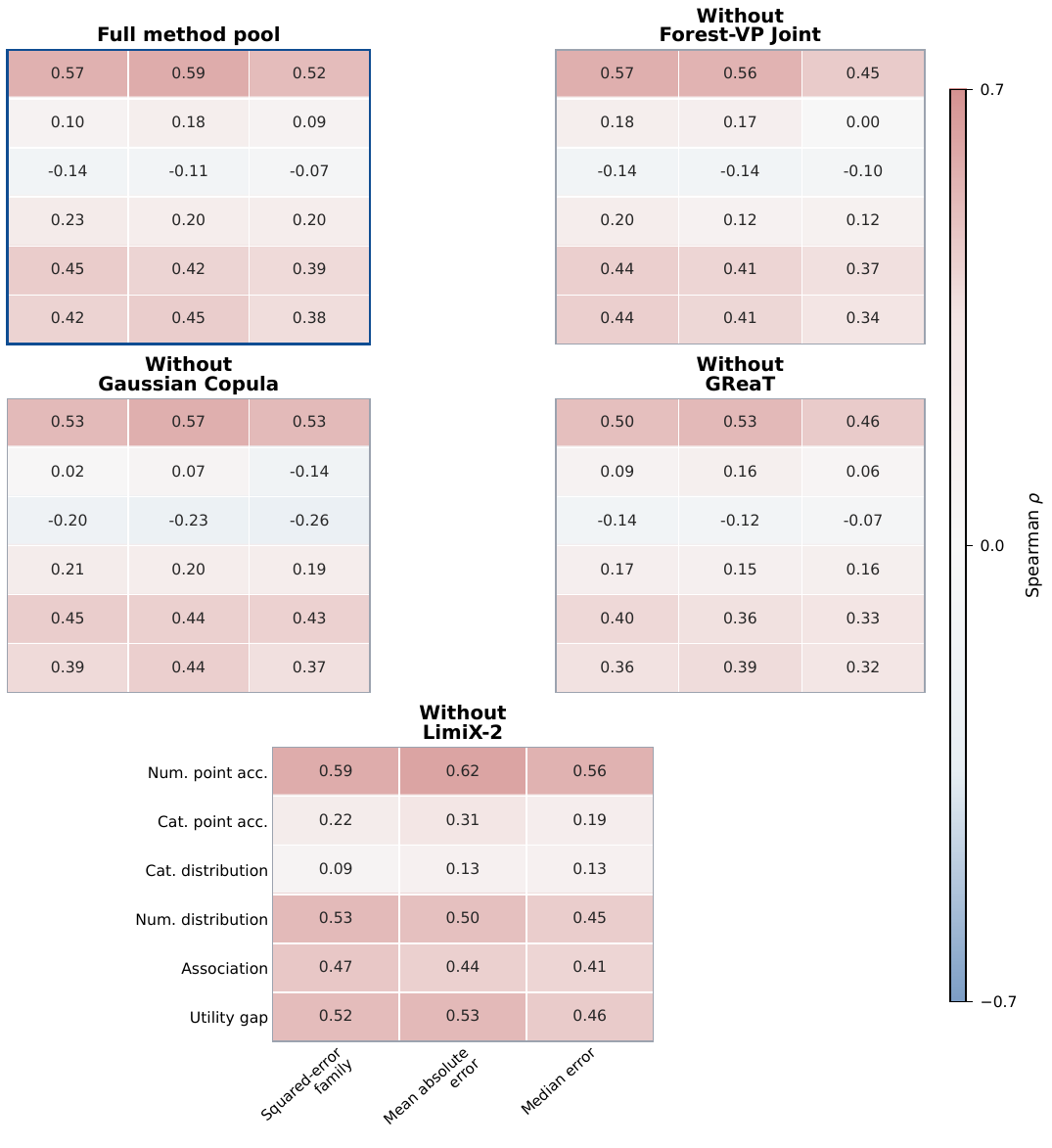}
    \caption{\trdetailcaptionfirst{I $\times$ P-reg}}
    \label{fig:tra-detail-i-preg-1}
\end{figure}

\begin{figure}[htbp]
    \centering
    \includegraphics[width=\linewidth,height=0.79\textheight,keepaspectratio]{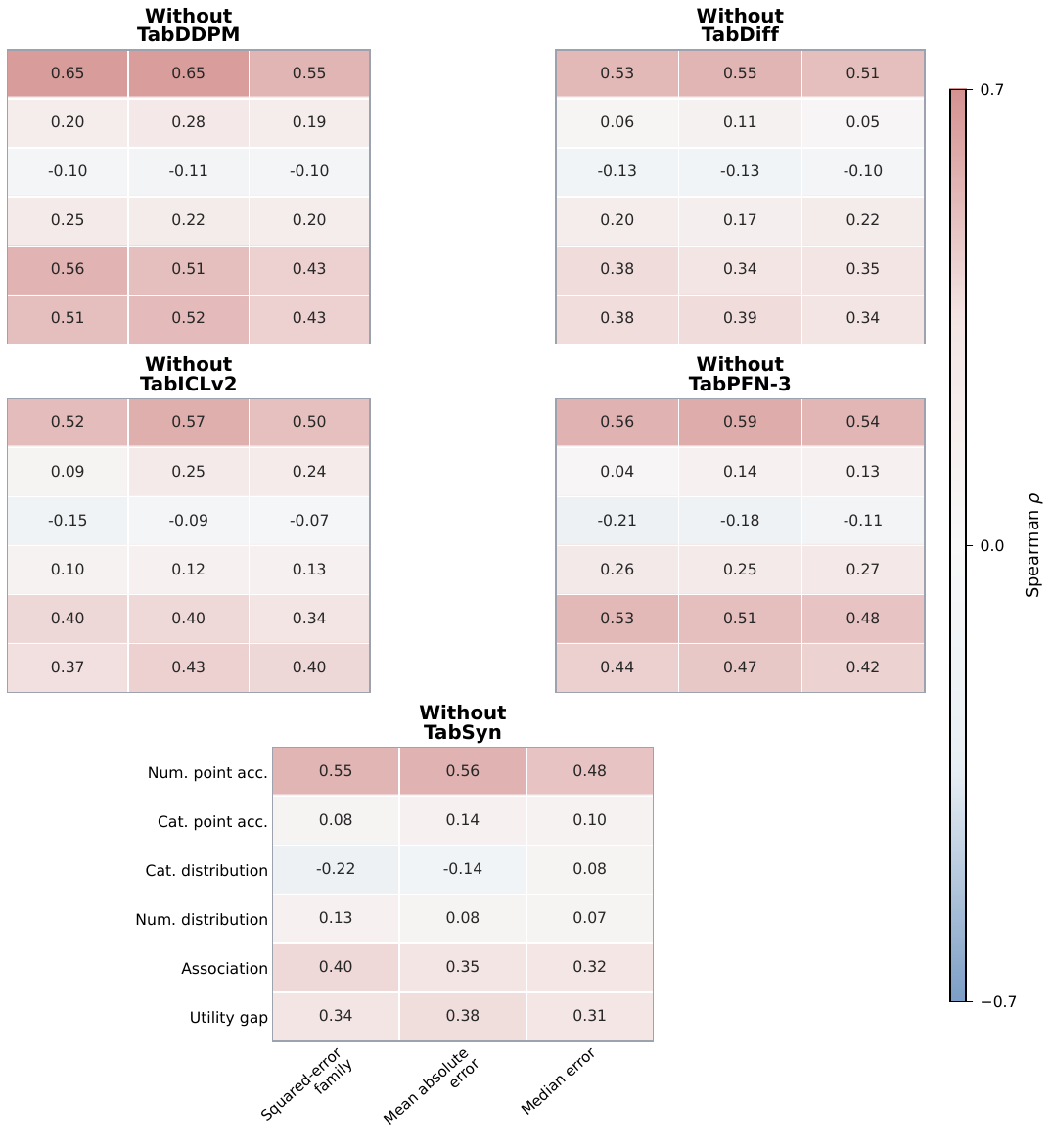}
    \caption{\trdetailcaptionsecond{I $\times$ P-reg}{fig:tra-detail-i-preg-1}}
    \label{fig:tra-detail-i-preg-2}
\end{figure}

\begin{figure}[htbp]
    \centering
    \includegraphics[width=\linewidth,height=0.79\textheight,keepaspectratio]{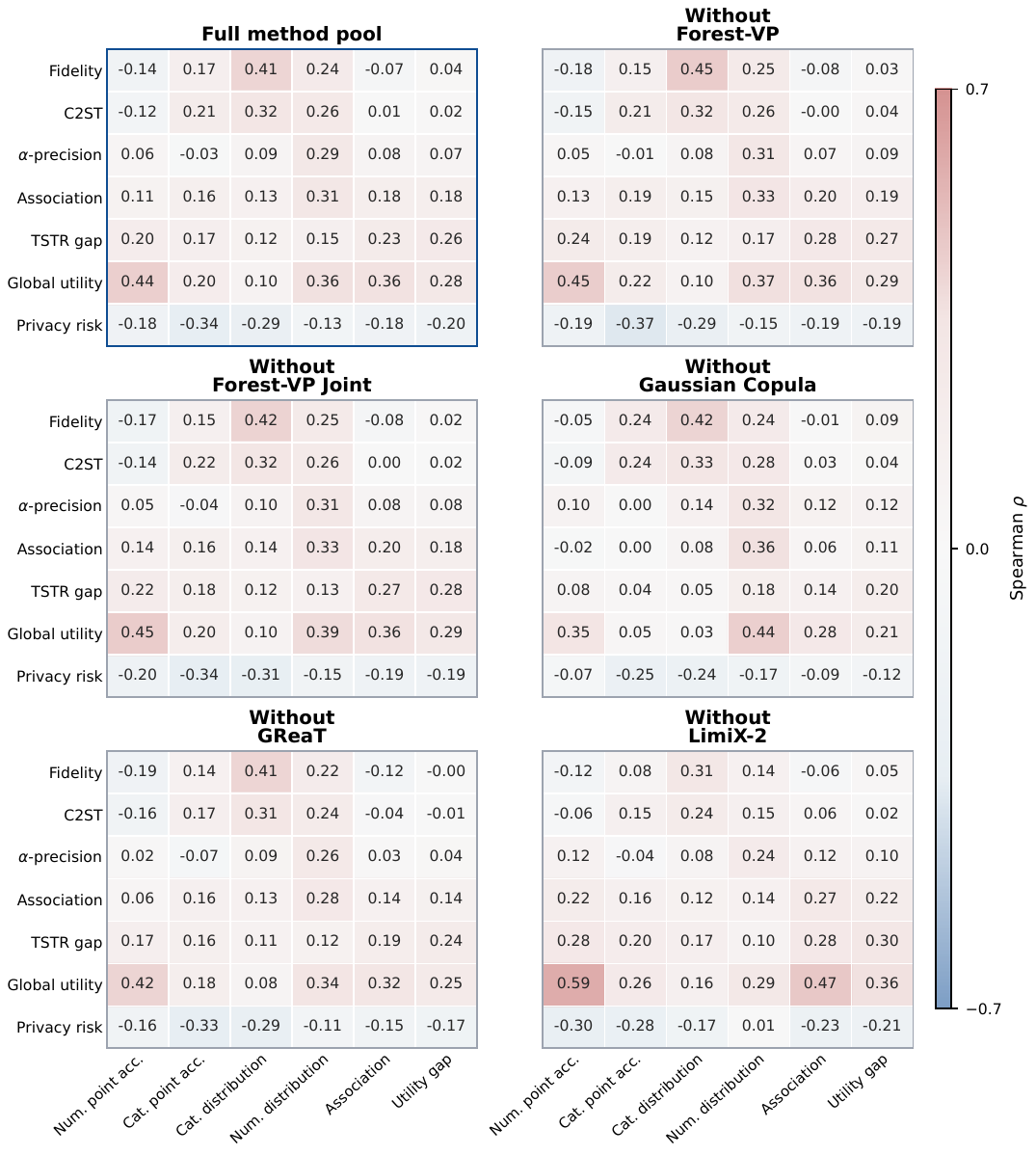}
    \caption{\trdetailcaptionfirst{G $\times$ I}}
    \label{fig:tra-detail-g-i-1}
\end{figure}

\begin{figure}[htbp]
    \centering
    \includegraphics[width=\linewidth,height=0.79\textheight,keepaspectratio]{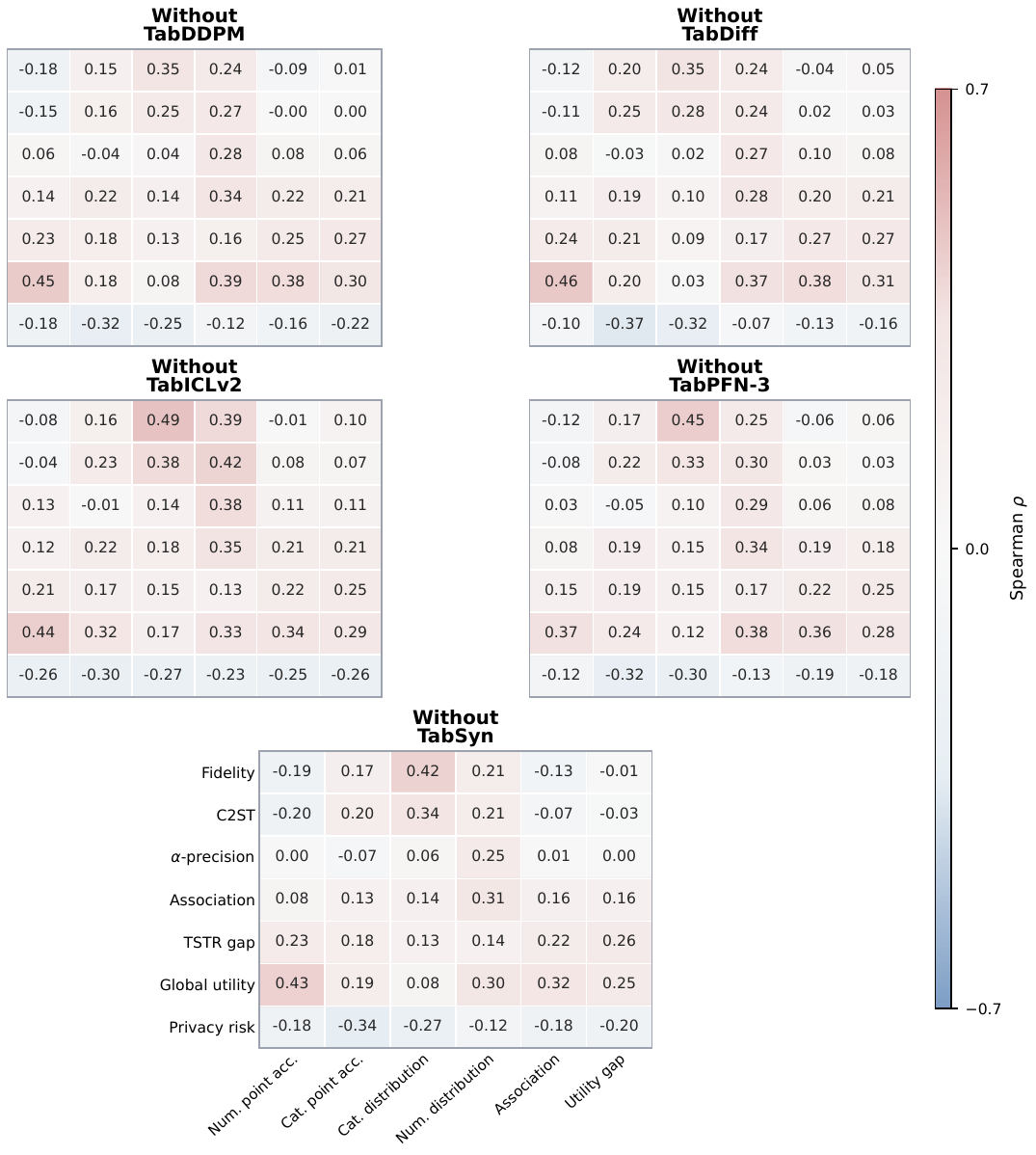}
    \caption{\trdetailcaptionsecond{G $\times$ I}{fig:tra-detail-g-i-1}}
    \label{fig:tra-detail-g-i-2}
\end{figure}

\begin{figure}[htbp]
    \centering
    \includegraphics[width=\linewidth,height=0.79\textheight,keepaspectratio]{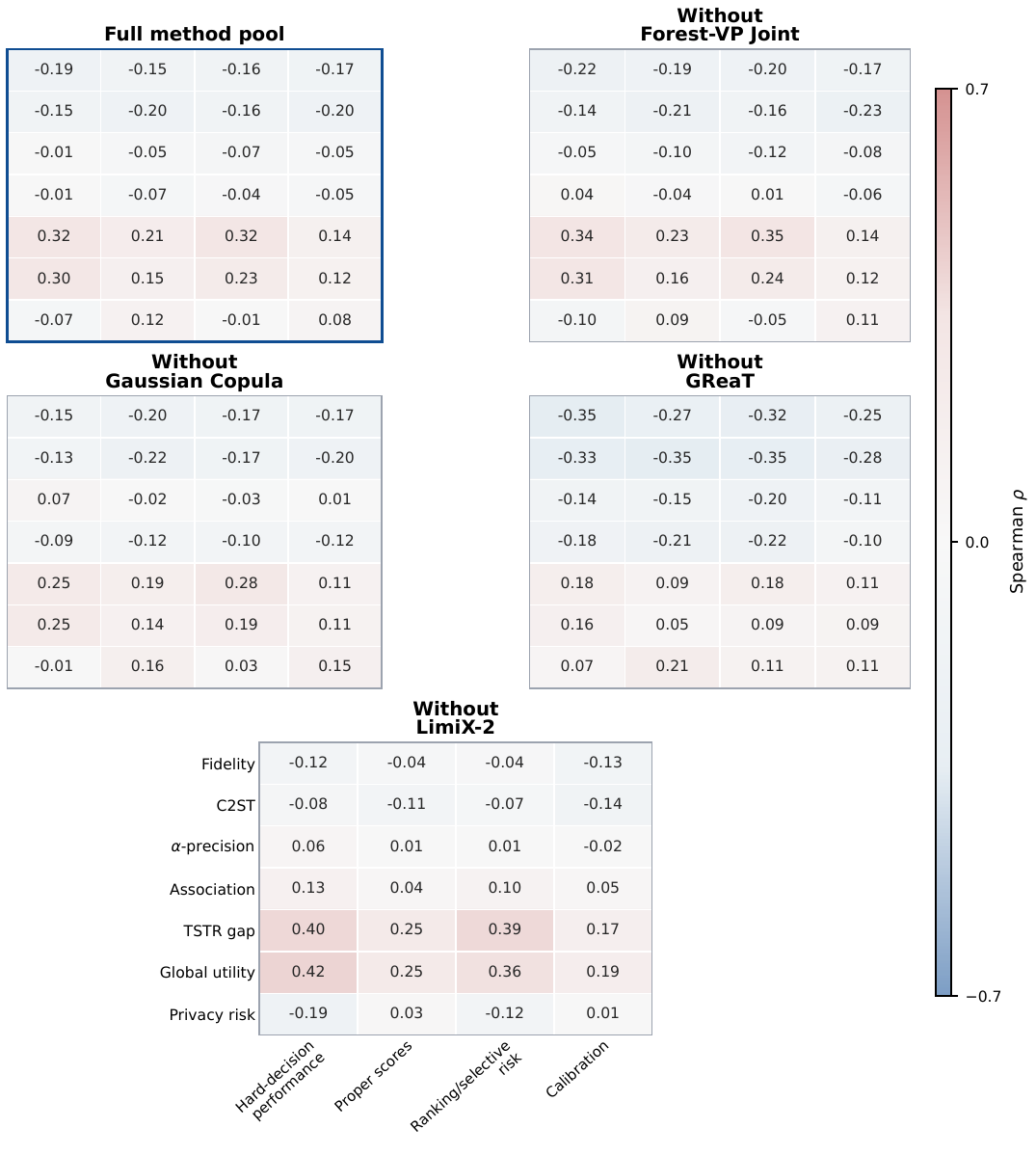}
    \caption{\trdetailcaptionfirst{G $\times$ P-cls}}
    \label{fig:tra-detail-g-pcls-1}
\end{figure}

\begin{figure}[htbp]
    \centering
    \includegraphics[width=\linewidth,height=0.79\textheight,keepaspectratio]{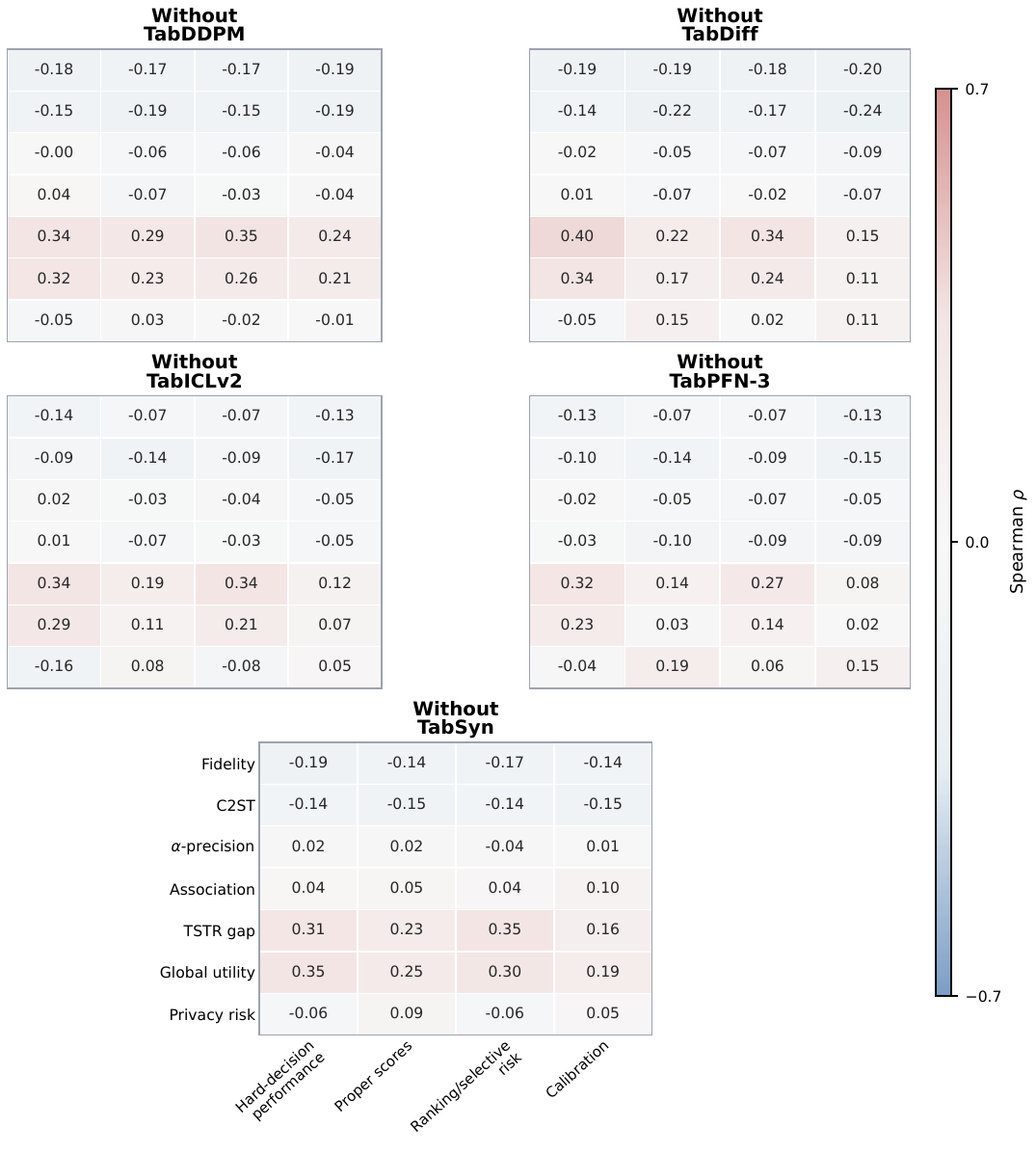}
    \caption{\trdetailcaptionsecond{G $\times$ P-cls}{fig:tra-detail-g-pcls-1}}
    \label{fig:tra-detail-g-pcls-2}
\end{figure}

\begin{figure}[htbp]
    \centering
    \includegraphics[width=\linewidth,height=0.79\textheight,keepaspectratio]{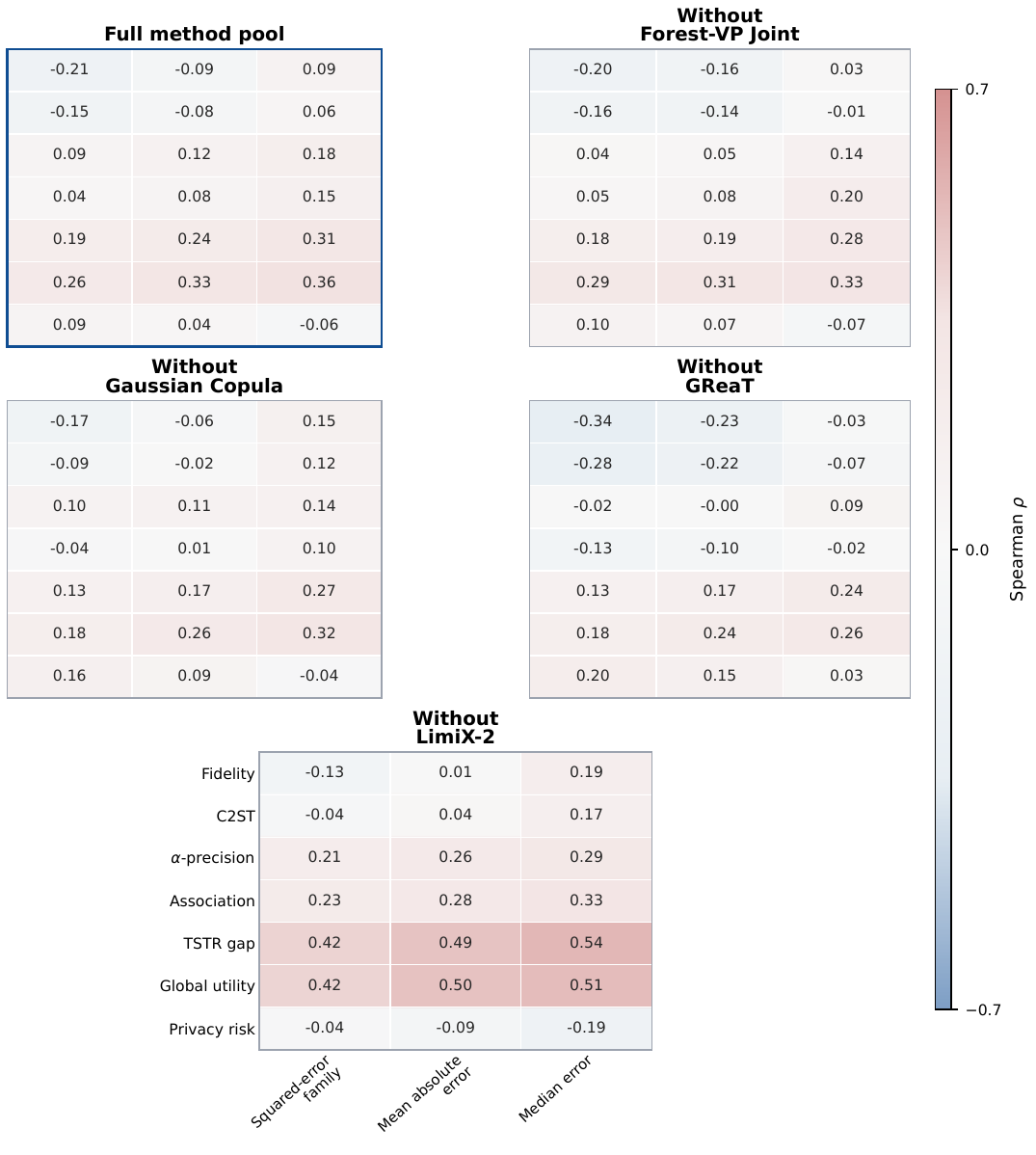}
    \caption{\trdetailcaptionfirst{G $\times$ P-reg}}
    \label{fig:tra-detail-g-preg-1}
\end{figure}

\begin{figure}[htbp]
    \centering
    \includegraphics[width=\linewidth,height=0.79\textheight,keepaspectratio]{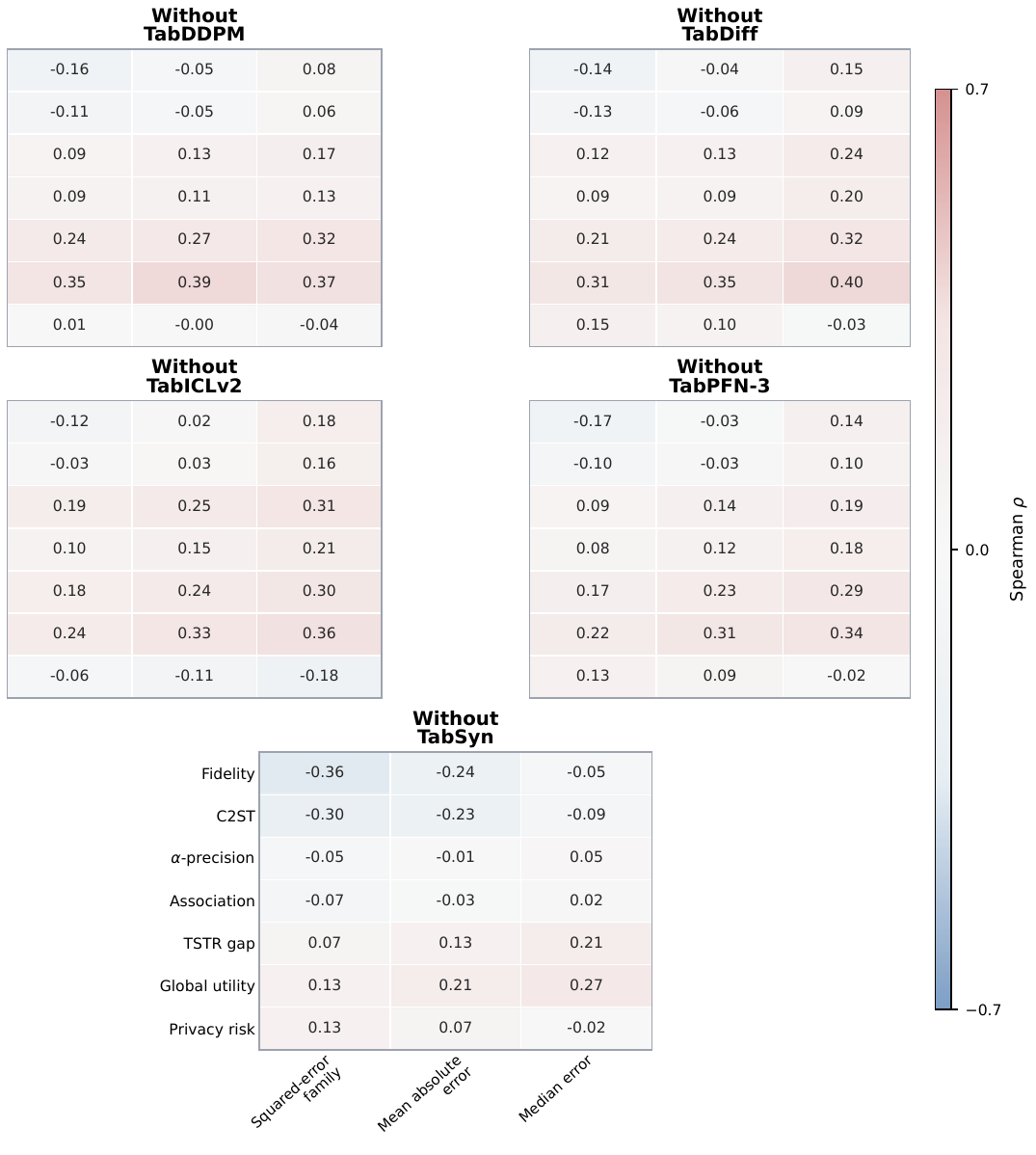}
    \caption{\trdetailcaptionsecond{G $\times$ P-reg}{fig:tra-detail-g-preg-1}}
    \label{fig:tra-detail-g-preg-2}
\end{figure}

Figures~\ref{fig:tra-detail-i-pcls-1}--\ref{fig:tra-detail-g-preg-2} show the complete correlation grids for the full candidate pool and every LOMO recomputation.
Each task pair occupies two pages for legibility.
Each cell contains the pooled correlation coefficient.
The grids identify which method removals account for changes in direction or large LOMO spans.

\clearpage

\clearpage
\clearpage

\clearpage
\clearpage

\clearpage
\clearpage

\clearpage
\clearpage

%% file: sections/e_causal_transfer.tex
\section{Controlled Transfer: Protocol and Additional Diagnostics}
\label{app:causal-transfer}

This section expands the matched-objective study in Section~\ref{sec:transfer}.
It specifies complete-member matching, inference policies, uncertainty,
backbone heterogeneity, conditional interactions, and objective-arm Pareto
frontiers. The implementation and re-evaluation inventory appear in
Appendix~\ref{app:transfer-implementation}.

\subsection{Seven-arm member-complete construction}
The intervention unit is a fixed backbone--dataset--training-seed block, with
arms $\mathcal{C}=\{G,I,P,GI,GP,IP,GIP\}$. JapaneseVowels, cleveland,
credit\_card\_clients\_default, and website\_phishing are classification
datasets; MiamiHousing2016 is the only regression dataset. Table~\ref{tab:transfer-datasets}
gives their OpenML identifiers and fixed splits. Within each atomic setting,
we orient every metric member and assign average-tie ranks across exactly the
seven arms, normalized to $[0,1]$. Members receive equal weight within an
axis, which is formed only when every declared member is finite for every
arm. Imputation mechanism and rate remain separate through ranking and
matching; setting-level contrasts are then averaged with equal weight.
Structural applicability follows the task contract. Complete-member support
is required within a backbone, not across an identical rectangle of all
backbones, datasets, and seeds.

The causal population comprises Gaussian Copula, Forest-VP Joint, TabDiff,
TabSyn, and TabDDPM. Forest-VP Joint uses native query-masked trees: unqueried
tasks contribute neither gradients nor Hessians. GReaT is excluded from the
factorial contrasts. For the four diffusion-based backbones, I and P
recipients use clean observed features ($x_0$) when their objective is enabled
and the $x_t$ conditioning path otherwise. Gaussian Copula uses its native
Gaussian conditional query in every arm. Recipient-conditioned contrasts
hold the relevant rule fixed on both sides.
Generation inference is fixed across arms. I/P self-inclusion and cross-arm
Pareto comparisons additionally involve the policy switch in the four
selector-based adapters. In $G\times I\mid P$, generation and prediction
rules remain fixed, but the I rule changes in those adapters when I is
toggled; I-side interactions therefore cannot be attributed solely to
changes in the training loss.

\subsection{Uncertainty and robustness}
Intervals use 2,000 crossed-bootstrap draws~\citep{efron1979bootstrap,owen2012bootstrapping},
resampling backbones and datasets and then seeds within selected blocks,
with all arms paired. Two-sided Wilcoxon diagnostics~\citep{wilcoxon1945individual}
use backbone--dataset means and receive Holm adjustment~\citep{holm1979simple}
within each analysis family. A robust marker requires a bootstrap interval
excluding zero, adjusted $p<.05$, magnitude at least $.05$ rank units,
all five backbones represented, at least two datasets, and non-zero
backbone-specific means sharing one sign. Dataset-sign and
leave-one-backbone summaries are diagnostics rather than additional gate
conditions.

The analysis reports 40 off-diagonal estimates, of which 28 meet the
support requirement; none passes the full gate. Of 20 self-inclusion
estimates, 14 are support-eligible and 11 are robustly positive. Generation
passes on fidelity, C2ST, $\alpha$-precision, association, TSTR, and global
utility. Imputation passes on association, categorical point accuracy,
categorical distribution, numerical point accuracy, and downstream utility.
Generation self-inclusion holds inference fixed. I/P self-inclusion also
selects the inference policy in the four selector-based adapters and is a
system-level comparison. Probability-based classification axes have only four
backbones because TabDiff lacks probability outputs, while regression has
only one dataset; neither group meets the full confirmatory support rule.

\subsection{Utility scope and backbone heterogeneity}
For P$\rightarrow$G, retained TSTR is $0.092\;[-0.041,0.225]$ and global
utility is $-0.008\;[-0.048,0.036]$. A separate scope-matched per-column
family yields target, all-column, and non-target effects of $.069$, $-.019$,
and $-.017$. The direct paired localization contrasts remain unresolved:
target-minus-all is $0.088\;[-0.072,0.260]$, and target-minus-non-target is
$0.086\;[-0.110,0.278]$; both have Holm-adjusted $p=.133$.
Figure~\ref{fig:causal-utility-scope} shows the backbone variation behind pooled
estimates rather than treating target alignment as an established mechanism.

\begin{figure}[htbp]
    \centering
    \includegraphics[width=\textwidth]{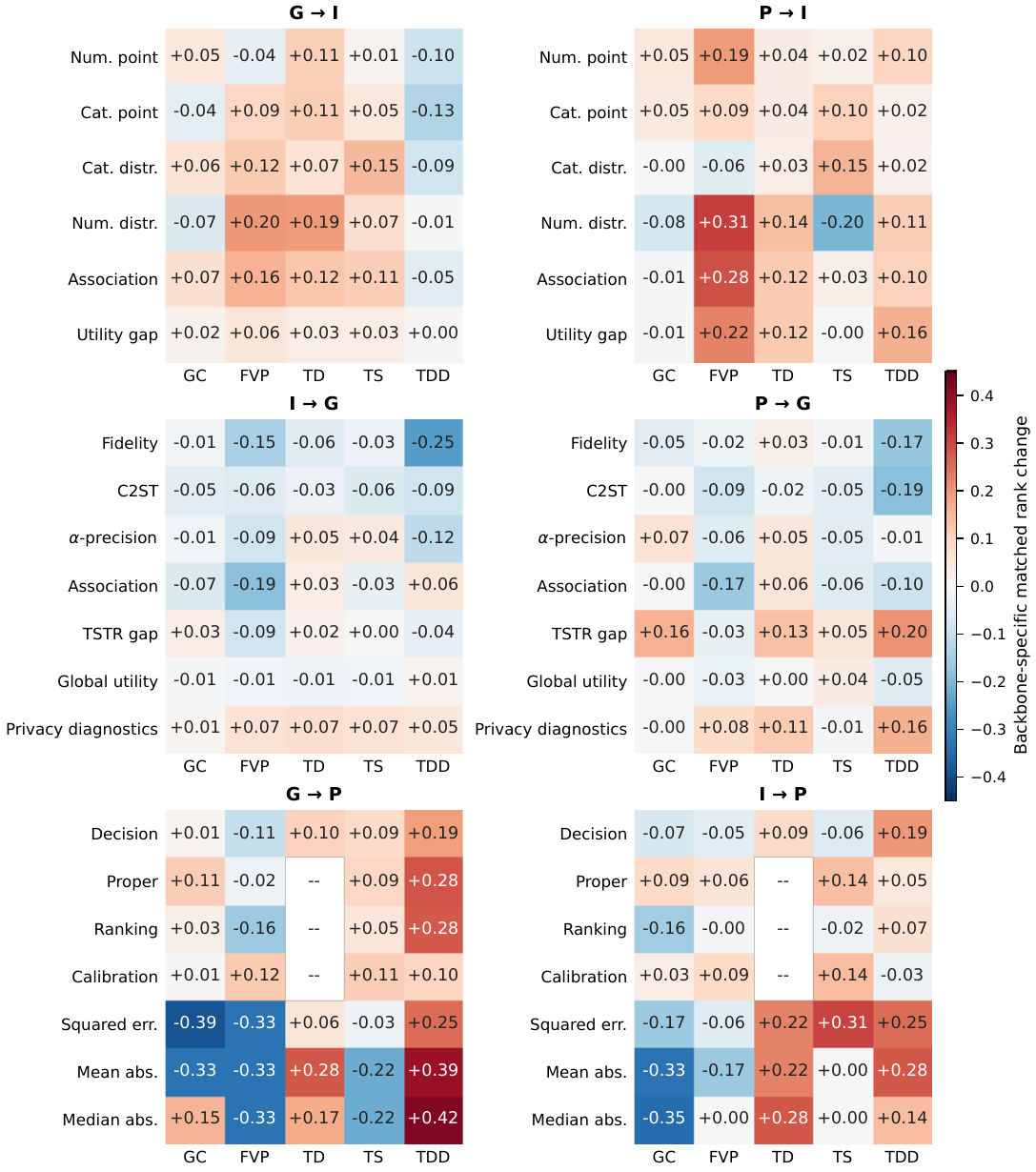}
    \caption{\textbf{Backbone heterogeneity in recipient-conditioned
    transfer.} Each cell is a backbone-specific matched rank change for one
    source--recipient axis. Columns GC, FVP, TD, TS, and TDD denote Gaussian
    Copula, Forest-VP Joint, TabDiff, TabSyn, and TabDDPM, respectively.
    These descriptive panels show variation behind
    pooled estimates and do not define an extra family of primary claims.}
    \label{fig:app-transfer-heterogeneity}
\end{figure}

Imputation mechanism and rate remain atomic throughout ranking and matching.
All 144 setting-specific off-diagonal estimates fail the full robustness
gate. Blank cells denote unavailable evaluations rather than zero effects.

\begin{figure}[htbp]
    \centering
    \includegraphics[width=\textwidth]{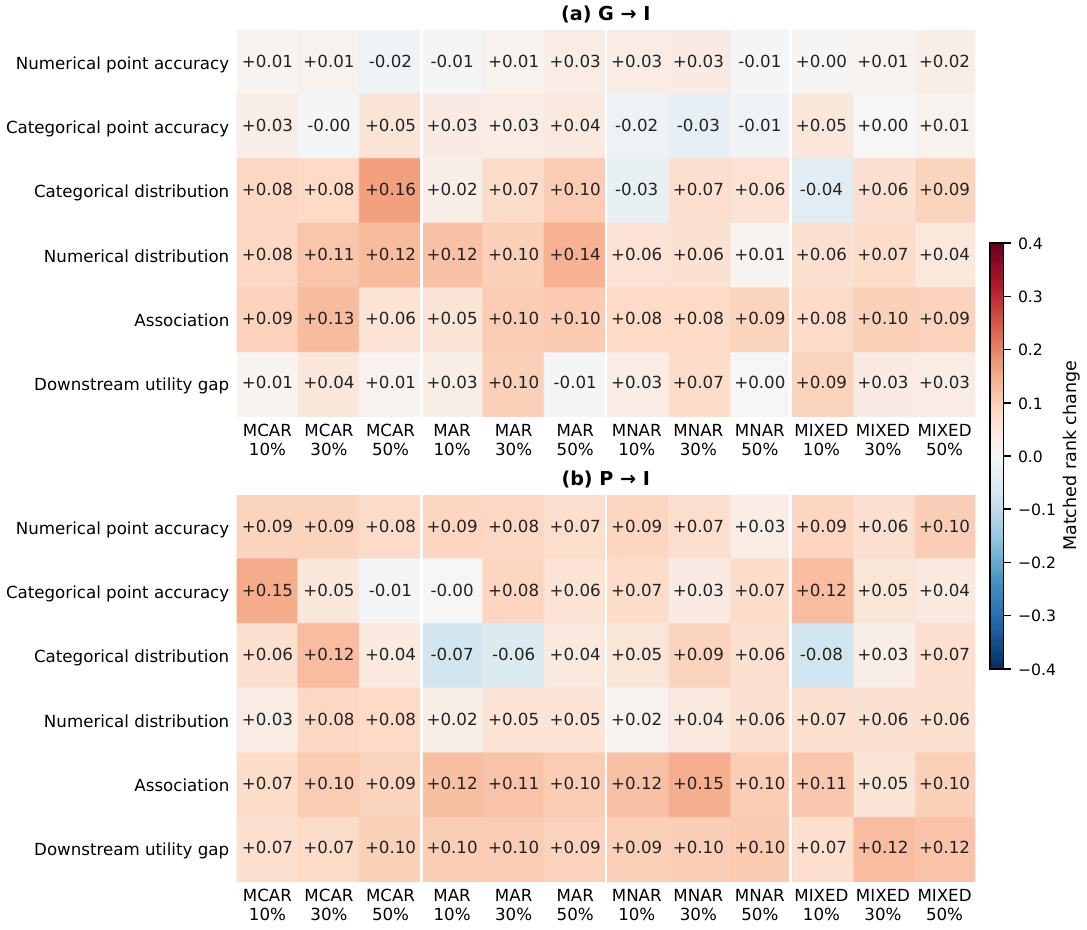}
    \caption{\textbf{Transfer into imputation across missingness settings.}
    Panels add G or P to an I recipient under its fixed clean-observation
    protocol. Each mechanism--rate setting is matched separately before
    equally weighted aggregation. Missing outcomes are left blank.}
    \label{fig:app-imputation-transfer}
\end{figure}

\subsection{Conditional interactions}
Of 60 anchor-conditional interaction estimates, 42 meet the support rule and
seven pass the gate, all negative $G\times I\mid P$ effects:
generation association ($-.315$), C2ST ($-.393$), fidelity ($-.468$), and
global utility ($-.154$), and imputation association ($-.391$), numerical
point accuracy ($-.475$), and utility gap ($-.284$).
These describe sub-additivity under the implemented anchored systems, not
independent harm from G or I or an identified optimization mechanism.
Generation-side comparisons hold their evaluation rule fixed; I-side
interactions additionally include the change in conditioning policy.
None of the 14 support-eligible GIP-versus-singleton gaps (20 reported
estimates) establishes a robust universal advantage or tax.

\subsection{Complete-block Pareto analysis}
The candidates are the seven arms. A task coordinate averages its declared
applicable semantic axes equally; the privacy-risk diagnostic is not part of
the G quality coordinate. I first averages all twelve mechanism--rate scores
within each axis. The primary block analysis retains a backbone's
training-seed block only when every applicable G/I/P axis and all twelve I
settings are available for all seven arms. It retains 35 classification
blocks from four backbones and 12 regression blocks from five backbones.
TabDiff classification is omitted from this full-axis comparison because its
probability axes are unavailable. Their joint summary has 47 blocks, with
seeds, then datasets within each backbone, then backbones equally weighted;
it is not a common cross-backbone rectangle. Regression conclusions concern
MiamiHousing2016 alone.

Exact dominance requires no loss on any task and a strict gain on one;
material dominance requires no loss and a gain of at least $\epsilon=.05$ on
one coordinate. The margin does not permit a $.05$ loss on another task.
Frontier membership is computed once in three dimensions, not separately in
its pairwise projections. For point estimates, the exact and material sets
coincide: joint $\{GI,GP,GIP\}$, classification
$\{G,GI,GP,GIP\}$, and regression $\{G,GI,GP,IP,GIP\}$.
The joint exact-frontier bootstrap frequencies are $89.35\%$ for GI,
$95.25\%$ for GP, and $98.55\%$ for GIP. These describe stability under
resampling, not posterior optimality probabilities or significance levels.
The Pareto bootstrap resamples backbones, then each selected backbone's own
datasets and seeds, preserving all-arm pairing.

\begin{figure}[htbp]
    \centering
    \begin{subfigure}[t]{\textwidth}
        \centering
        \includegraphics[width=\linewidth]{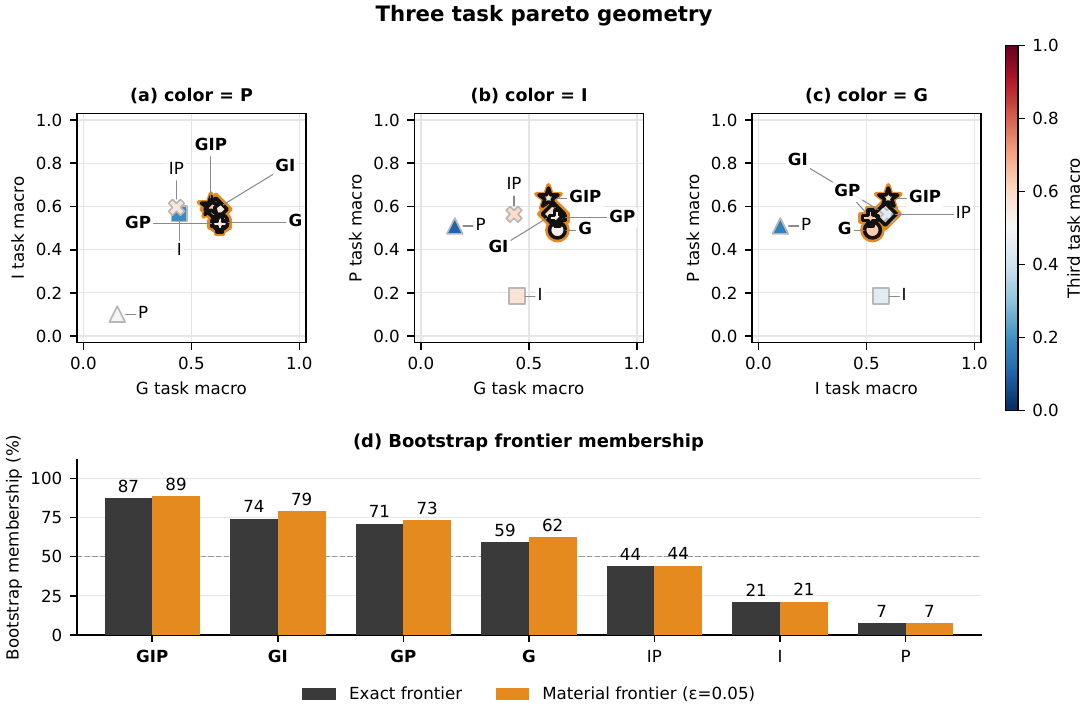}
        \caption{Complete-axis classification: four backbones, 35 blocks.}
    \end{subfigure}
    \vspace{0.7em}
    \begin{subfigure}[t]{\textwidth}
        \centering
        \includegraphics[width=\linewidth]{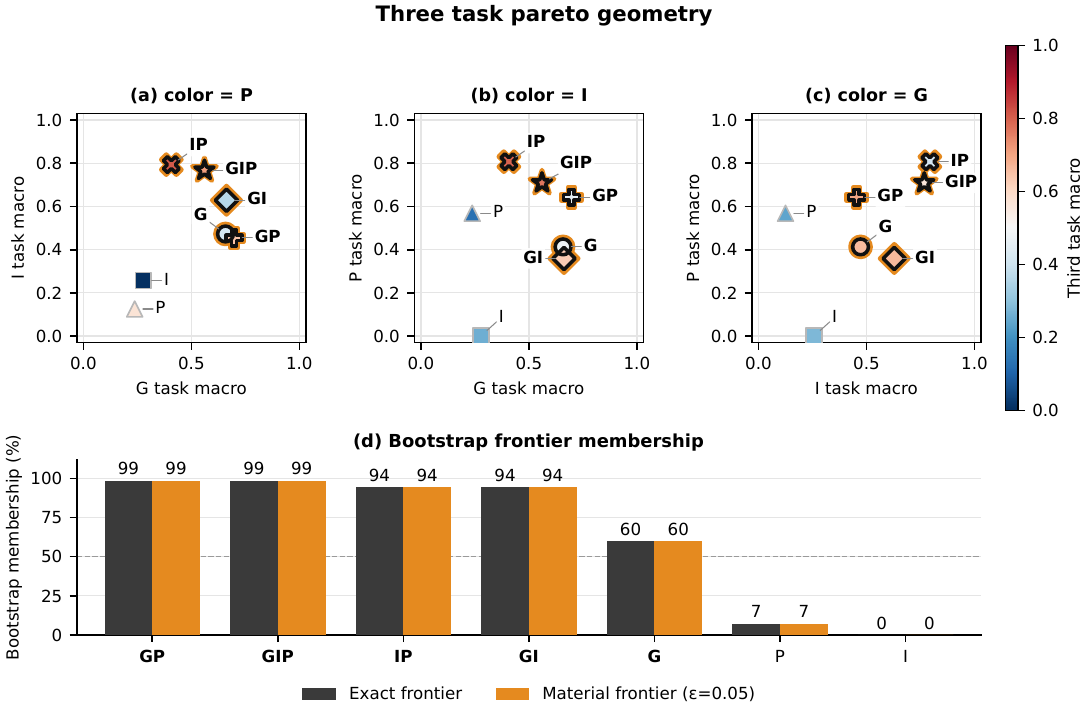}
        \caption{Regression: five backbones, 12 blocks, one dataset.}
    \end{subfigure}
    \caption{\textbf{Objective-arm Pareto decompositions.} Each subfigure
    shows three projections of one joint $(G,I,P)$ frontier and its bootstrap
    membership frequencies below. Black outlines mark exact-frontier arms;
    orange halos mark the $\epsilon=.05$ material frontier.}
    \label{fig:app-objective-pareto}
\end{figure}

\subsection{Additional support checks}
The available-coordinate summary in Section~\ref{sec:objective-pareto}
retains an axis--setting coordinate when every arm is scored, without
requiring the remaining axes or settings in that block. This permits
TabDiff's classification decision axis without its unavailable probability
axes. A five-backbone classification check restricted to hard decisions
retains 44 complete blocks and the frontier $\{G,GI,GP,GIP\}$; allowing
available coordinates additionally retains IP. The regression frontier is
unchanged. These checks explain why the main-text view and the stricter
complete-block analysis need not select identical arms.

%% file: sections/f_method_landscape.tex
\section{Fixed-Method Generality: Protocol and Additional Diagnostics}
\label{app:method-landscape}

This section expands the fixed-identity deployment study in
Section~\ref{sec:methods}.  Following the main-text order, it presents strict
estimands and specialist references, cross-task profile geometry, and portfolio
and operational diagnostics.

\subsection{Strict estimands and fixed-specialist baselines}

The method analysis uses the 20 semantic axes and 42 metric members specified in
Table~\ref{tab:metric-groups}.  Replicates are aggregated first; each member is
oriented and converted to percentile ranks with average ties within its
protocol instance; and members receive equal weight under the member-complete construction.
Target-observed and feature-only imputation are never placed in one
denominator.  The primary imputation estimand remains separate for every
mechanism--rate setting; the cross-task display coordinate is their equally
weighted mean, without reranking.

A fixed specialist keeps one identity across every dataset and axis in its
profile.  Exact dominance is evaluated jointly in the full profile, and
practical dominance permits an $\epsilon=0.05$ tolerance on every coordinate.
Uncertainty is estimated using 2,000 bootstrap draws of whole datasets.  Pairwise
tests remove numerical ties at $10^{-12}$, use a two-sided Wilcoxon signed-rank
test, and apply Holm correction within each declared comparison family.
Frontier frequency is a stability diagnostic, not a probability that a method
is optimal. The historical per-cell oracle is retained only as an unattainable
reference.

The fixed-specialist pool includes ARF~\citep{arf},
MissForest~\citep{missforest}, DiffPuter~\citep{diffputer}, KNN
imputation~\citep{knnimpute}, MICE~\citep{mice}, ReMasker~\citep{remasker},
CatBoost~\citep{catboost}, TabM~\citep{tabm}, LightGBM~\citep{lightgbm}, and
Random Forest~\citep{randomforest}.

\input{tables/specialist_frontier_summary}

\begin{figure}[htbp]
    \centering
    \includegraphics[width=\textwidth]{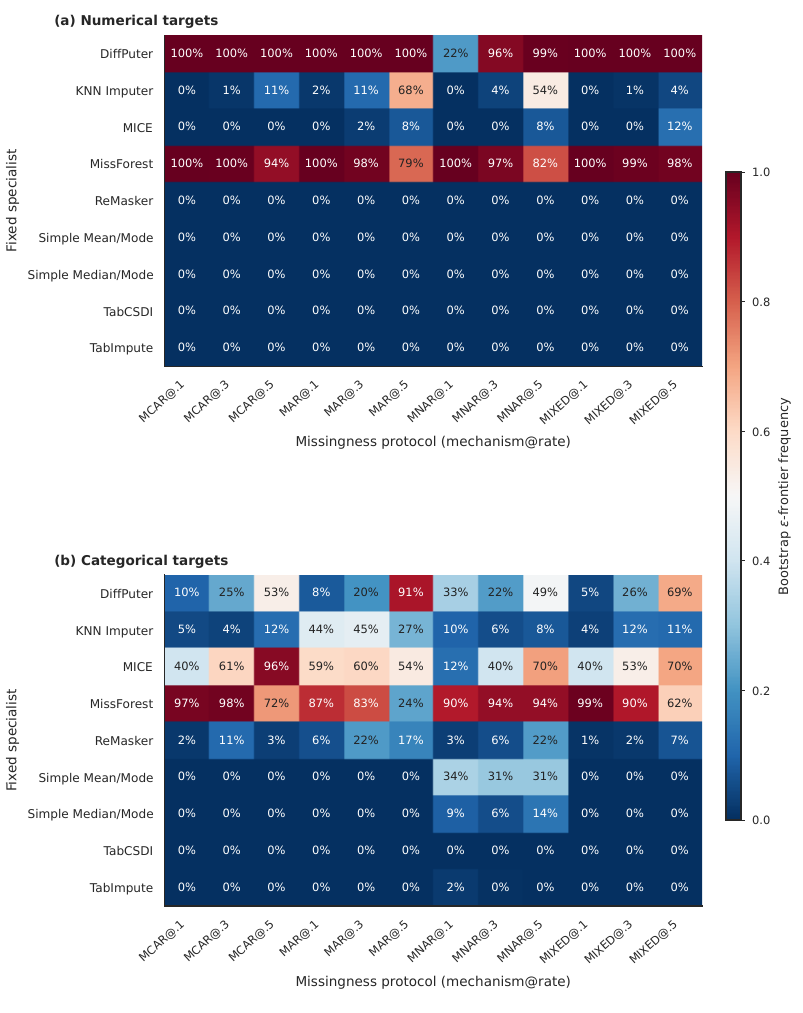}
    \caption{\textbf{Strict imputation specialist-frontier stability.}
    The top panel evaluates numerical targets and the bottom panel categorical
    targets.  Columns are the twelve target-observed mechanism--rate settings;
    cells give the frequency of membership in the $\epsilon=0.05$
    fixed-specialist frontier under whole-dataset bootstrap resampling.
    Settings are not pooled when computing the frontier.}
    \label{fig:strict-imputation-frontiers}
\end{figure}

MissForest is practically stable across all numerical settings, while
DiffPuter is nearly as persistent; their advantage is therefore not an
artifact of one missingness regime.  Categorical imputation is less uniform:
MissForest remains the most persistent reference, but MICE and DiffPuter enter
the frontier in setting-dependent regions.  Generation results involving the
normalized median Gower DCR use the provisional target-one transformation
$-|x-1|$ and are excluded from confirmatory privacy claims.

\subsection{Cross-task profile geometry and controls}

\begin{figure}[htbp]
    \centering
    \includegraphics[width=\textwidth]{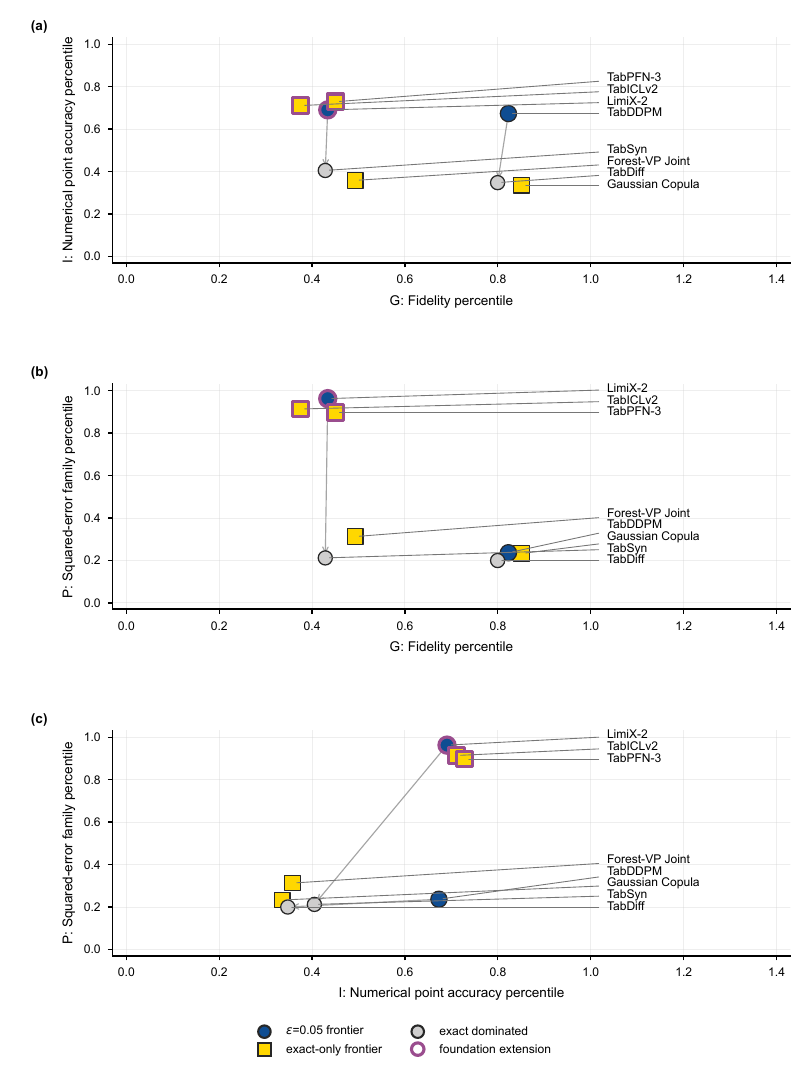}
    \caption{\textbf{Pairwise projections of the numerical-fidelity profile.}
    From top to bottom, the panels show the G--I, G--P, and I--P projections;
    dominance is always computed jointly in all three coordinates.  Blue
    circles are on the practical frontier, yellow squares are exact-only
    frontier points, gray circles are exactly dominated, and arrows show one
    representative dominance certificate.}
    \label{fig:generalist-numeric}
\end{figure}

\begin{figure}[htbp]
    \centering
    \includegraphics[width=\textwidth]{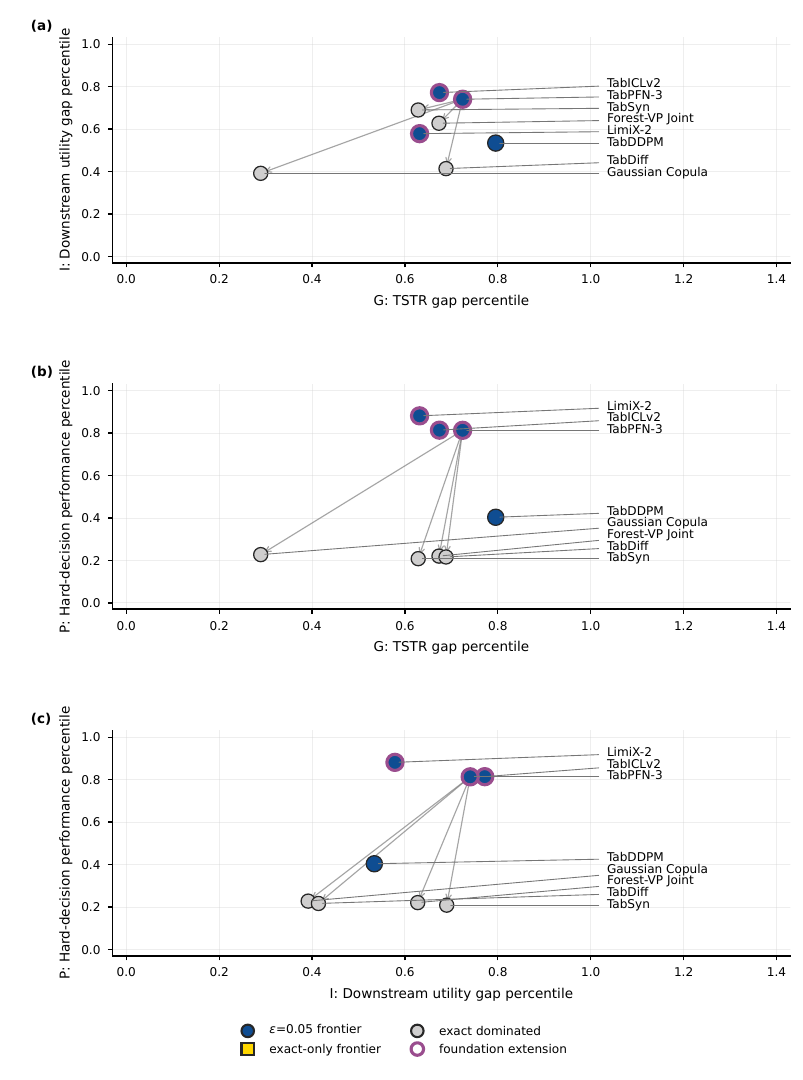}
    \caption{\textbf{Pairwise projections of the downstream-utility profile.}
    From top to bottom, the panels show the G--I, G--P, and I--P projections;
    symbols follow Figure~\ref{fig:generalist-numeric}, and dominance is
    computed in the full three-coordinate profile.}
    \label{fig:generalist-utility}
\end{figure}

The numerical-fidelity exact frontier contains Forest-VP Joint, Gaussian
Copula, LimiX-2, TabDDPM, TabICLv2, and TabPFN-3, but only LimiX-2 and TabDDPM
remain after the $\epsilon=0.05$ practical margin.  The downstream-utility
exact and practical frontiers both contain LimiX-2, TabDDPM, TabICLv2, and
TabPFN-3.  These sets are computed in three dimensions; the projections above
are explanatory views rather than separate frontier calculations.

\begin{figure}[htbp]
    \centering
    \includegraphics[width=\textwidth]{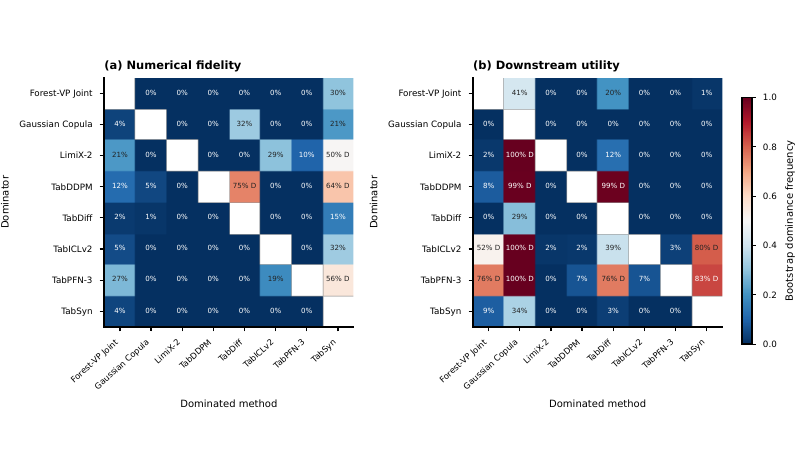}
    \caption{\textbf{Stability of named cross-task dominance relations.}
    Panel (a) is numerical fidelity and panel (b) downstream utility.  Rows
    are candidate dominators, columns are dominated methods, and cells give
    the whole-dataset bootstrap frequency of exact
    three-dimensional dominance.  ``D'' marks dominance in the frozen point
    estimate.}
    \label{fig:within-generalist-dominance}
\end{figure}

\begin{figure}[htbp]
    \centering
    \includegraphics[width=\textwidth]{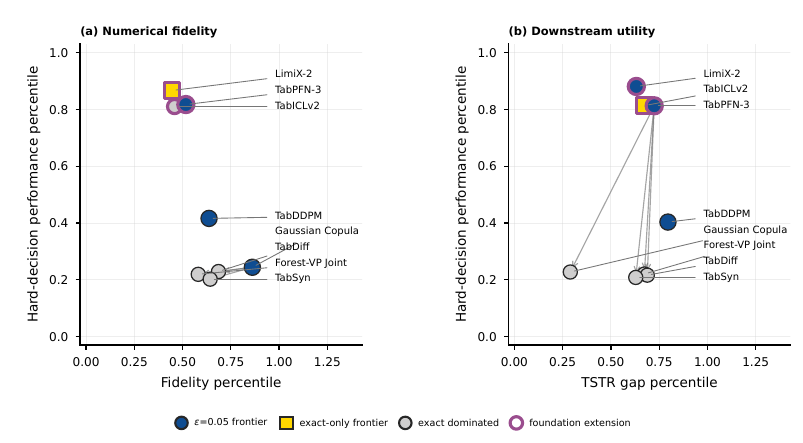}
    \caption{\textbf{Generation--prediction controls without an imputation
    coordinate.}  Panel (a) compares generation fidelity with classification
    hard decisions; panel (b) compares TSTR utility with the same prediction
    axis.  These two-dimensional controls avoid the
    aggregated imputation coordinate.}
    \label{fig:gp-frontiers}
\end{figure}

\subsection{Portfolio and operational diagnostics}

\begin{figure}[htbp]
    \centering
    \includegraphics[width=\textwidth]{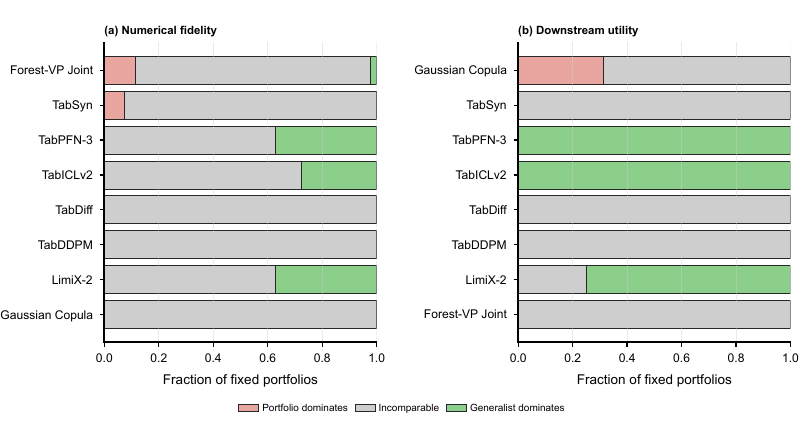}
    \caption{\textbf{In-sample relation between fixed generalists and fixed
    specialist portfolios.}  Panel (a) is numerical fidelity and panel (b)
    downstream utility.  Bars partition prespecified specialist triples into
    those that dominate the generalist, are dominated by it, or remain
    incomparable in the joint three-task geometry.  This maps available
    trade-offs; it is not a held-out selection result.}
    \label{fig:portfolio-relations}
\end{figure}

\paragraph{Specialist portfolios.}
\label{app:fixed-portfolios}
The specialist portfolios in the bottom row of
Figure~\ref{fig:fixed-identity-deployment} are the following fixed
generation--imputation--prediction triples: ARF, MissForest, and LightGBM for
numerical fidelity; ARF, MissForest, and CatBoost for categorical fidelity;
and ARF, DiffPuter, and Random Forest for downstream utility.  Each profile
uses one declared triple on every dataset in its common cohort.
Bootstrap intervals summarize the paired gaps over those datasets.

\begin{figure}[htbp]
    \centering
    \includegraphics[width=\textwidth]{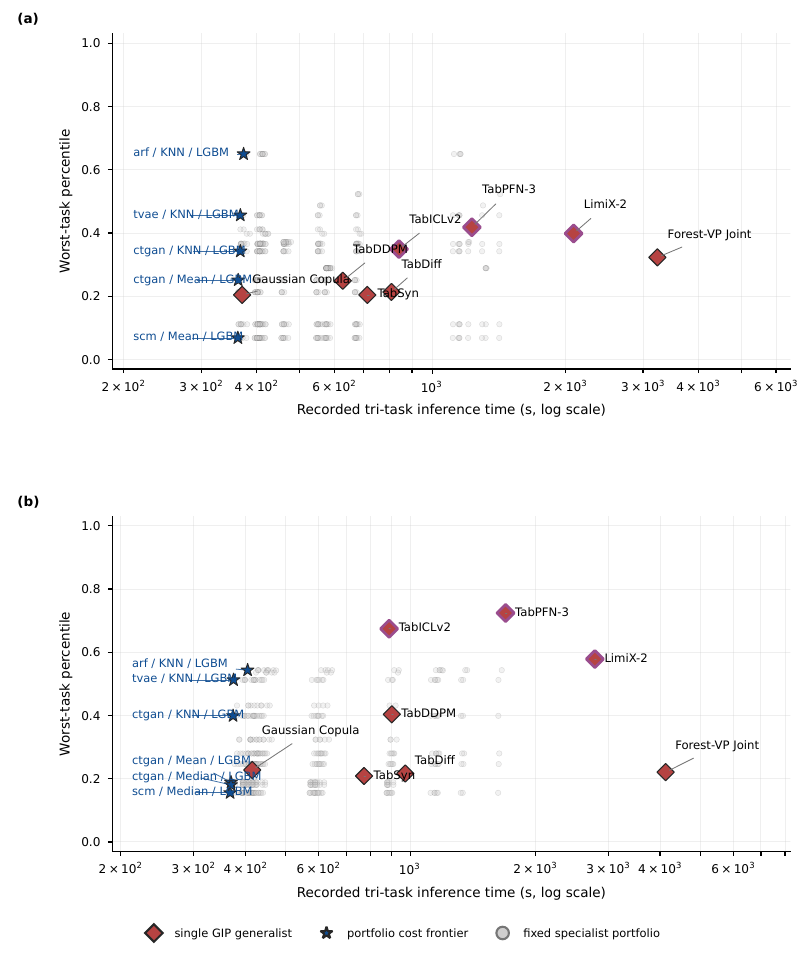}
    \caption{\textbf{Diagnostic quality--cost view.}  The top panel is
    numerical fidelity and the bottom panel downstream utility.  The vertical
    coordinate is worst-task percentile and the horizontal coordinate is
    tri-task inference time under a balanced workload.  Stars mark the
    fixed-portfolio cost frontier and diamonds mark generalists.  Costs refer to
    the displayed inference workload and serve as an operational diagnostic.}
    \label{fig:portfolio-cost}
\end{figure}

\subsection{Interpretation rules}

We use ``associated'' only for observational rank correlations, ``transfers''
only for matched objective contrasts, ``dominates'' only for a declared Pareto
profile, and ``portfolio comparison'' only for the declared fixed specialist
triples above.  Criterion-overlap rows are retained as controls but
do not count as independent cross-task evidence.

%% file: tables/specialist_frontier_summary.tex
\begin{table}[H]
\centering
\caption{\textbf{Fixed specialist reference frontiers.}  For imputation,
$x/12$ is the number of separately evaluated mechanism--rate settings in which
the method belongs to the indicated frontier; it is not a dataset-support
count.  The provisional generation privacy composite is excluded.}
\label{tab:specialist-frontiers}
\small
\setlength{\tabcolsep}{4pt}
\begin{tabularx}{\textwidth}{@{}l >{\raggedright\arraybackslash}X >{\raggedright\arraybackslash}X@{}}
\toprule
Task profile & Exact frontier & $\epsilon=0.05$ frontier \\
\midrule
Generation, six specified axes & ARF & ARF \\
Imputation, numerical targets & MissForest $12/12$; DiffPuter $12/12$;
KNN Imputer $4/12$; MICE $1/12$ & MissForest $12/12$; DiffPuter $11/12$;
KNN Imputer $2/12$ \\
Imputation, categorical targets & MissForest $11/12$; MICE $10/12$;
DiffPuter $6/12$; KNN Imputer $2/12$; ReMasker $2/12$ &
MissForest $11/12$; MICE $7/12$; DiffPuter $3/12$ \\
Prediction, classification & CatBoost; TabM & CatBoost; TabM \\
Prediction, regression & LightGBM; LightGBM Quantile; Random Forest &
LightGBM Quantile; Random Forest \\
\bottomrule
\end{tabularx}
\end{table}

%% file: sections/g_implementation.tex
\section{Implementation Details}
\label{app:implementation}

This appendix specifies the baseline settings, randomization conventions,
controlled-transfer configurations, and computational environment.
For benchmark implementations, method configurations, evaluation code, and
installation and execution instructions, see the public \ours{} repository:
\url{https://github.com/limix-ldm-ai/OTBench}.

\input{sections/g_1_baseline_configurations}

\input{sections/g_2_controlled_transfer}

%% file: sections/g_1_baseline_configurations.tex
\subsection{Baseline configurations and evaluation protocol}
\label{app:baseline-implementation}

\paragraph{Configuration policy.}
Each baseline starts from a method-specific TOML configuration, with
experiment-specific overrides applied before fitting.  The benchmark runners
instantiate these settings directly rather than running a shared external
hyperparameter search.  Table~\ref{tab:baseline-configurations} summarizes the
principal default settings for architecture, optimization, and inference; the
full method configurations are available in the
\href{https://github.com/limix-ldm-ai/OTBench}{public \ours{} repository}. HyperImpute retains
its native model selection for each column and iteration.  Its classifier
candidates are random forest, logistic regression, XGBoost, CatBoost, and neural
networks; its regressor candidates are random forest, linear regression,
XGBoost, CatBoost, and neural networks.  It uses ten inner iterations, lazy
selection, and an optimization threshold of 1,000.  It fits each incomplete
evaluation matrix through its transductive \texttt{fit\_transform} interface.

\paragraph{Training and validation.}
The data store supplies the fixed train--test split described in
Appendix~\ref{app:dataset-construction}.  TabDDPM, TabDiff, and TabSyn use a
deterministic 10\% validation subset of the training rows, selected with seed 0.
Their training-data view redirects any upstream training-time ``test'' access to
this validation subset.  XGBoost, LightGBM, CatBoost, NGBoost,
LightGBM-Quantile, FT-Transformer, TabM, and CARD's guidance network use a
20\% internal validation split determined by the training seed.  FT-Transformer
and TabM use AdamW and retain the state with the lowest validation loss;
their patience is 16 epochs.  MLP uses scikit-learn's internal early-stopping
split.  Ridge, logistic regression, and random forest fit their training rows
without an additional benchmark validation split.

\paragraph{Randomization and evaluation settings.}
The seed set is $\{0,1,2\}$. Generation uses the same seed for
training and sampling. Imputation uses paired seed tuples
$(s_{\rm train},s_{\rm selection},s_{\rm mask},s_{\rm impute})=(s,s,s,s)$ for
$s\in\{0,1,2\}$, across MCAR, MAR, MNAR, and mixed MCAR/MAR/MNAR at rates
$\{0.1,0.3,0.5\}$. Prediction is evaluated separately for each training seed.
Training seeds and conditional-sampling draws are distinct randomization levels.

\begingroup
\footnotesize
\setlength{\tabcolsep}{4pt}
\renewcommand{\arraystretch}{1.10}
\begin{longtable}{@{}p{0.21\linewidth}p{0.75\linewidth}@{}}
\caption{Principal default baseline settings. Learning rates are denoted by LR, weight
decay by WD, and batch size by $B$. Budgets are expressed in the units used by
each implementation. Full per-method TOML configurations are available in the
\href{https://github.com/limix-ldm-ai/OTBench}{public \ours{} repository}.}\label{tab:baseline-configurations}\\
\toprule
Method & Configuration \\
\midrule
\endfirsthead
\multicolumn{2}{l}{\small Table~\ref{tab:baseline-configurations} (continued)}\\
\toprule Method & Configuration \\
\midrule
\endhead
\bottomrule
\endfoot
\multicolumn{2}{@{}l}{\textit{Methods with generation or conditional-query interfaces}}\\
TabDDPM & MLP widths $[1024,2048,2048,1024]$; time embedding 128; 1,000 diffusion steps; cosine schedule; 100,000 updates, LR $10^{-3}$, WD $5\!\times\!10^{-4}$, $B=4096$; quantile normalization.\\
TabDiff & Two layers, token width 4, one head, factor 32, time width 1,024; 50 sampling steps; 8,000 training epochs, LR $10^{-3}$, EMA 0.997, $B=4096$; main and target-only prediction-guidance models.\\
TabSyn & Two-layer, width-4 VAE tokens; latent diffusion width 1,024; VAE 4,000 epochs and diffusion 10,001 epochs, LR $10^{-3}$, $B=4096$, patience 500; 50 sampling steps, 20 imputation resampling rounds.\\
Forest-VP / Joint & 50 time points, 100 noisy duplicates, 100 trees, depth 7, shrinkage 0.3; histogram trees; RePaint $r=5$, $j=0.1$; inference row batches 256 (I) and 1,024 (G).\\
Gaussian copula & At most 1,000 marginal quantiles; 100 conditional samples for regression and 200 class-conditional samples.\\
GReaT & Training iteration parameter 5; at most 3,000 training rows; completion temperature 0.7, batch 100, maximum length 1,000, and at most 15 retries.\\
CTGAN & Generator/discriminator: two hidden layers of width 500; 100 training iterations, one discriminator step, LR $10^{-3}$, WD $10^{-3}$.\\
TVAE & Encoder/decoder: three hidden layers of width 500; embedding width 500; 100 training iterations, LR $10^{-3}$, WD $10^{-5}$.\\
NFlow & One hidden layer of width 100; rational-quadratic autoregressive transform with permutation; dropout 0.1; 1,000 iterations, LR $10^{-3}$; at most 2,000 training rows.\\
ARF & 30 trees, minimum node size 5, at most ten adversarial iterations, early stopping enabled.\\
Bayesian network & Tree structure search with K2 score; structure-learning iteration parameter 1,000; at most 10,000 training rows; 100-bin numerical discretization.\\
SCM generator & DirectLiNGAM causal-discovery backend.\\
NRGBoost & 200 trees, shrinkage 0.15, at most 256 leaves; 80,000 model samples, 16 chains, burn-in 100.\\
TabEBM & 200 SGLD steps, step size 0.1, noise standard deviation 0.01; at most 10,000 context rows; chunked one-hot inference.\\
SMOTE / SMOTE-NC / SMOTER & Five neighbors; task-appropriate numerical, mixed-type, or regression interpolation.\\
Independent marginals / training-row bootstrap & Empirical per-column sampling / empirical row sampling with replacement.\\
\midrule
\multicolumn{2}{@{}l}{\textit{Imputation specialists}}\\
Mean/mode; median/mode; kNN & Numerical mean or median with categorical mode; kNN uses five neighbors.\\
MICE / MissForest & Ten chained iterations / at most ten forest iterations with 100 trees and tolerance $10^{-3}$.\\
HyperImpute & Five classifier and five regressor candidates; ten inner iterations; model selection by column and iteration; lazy selection threshold 1,000.\\
ReMasker & Encoder/decoder depths 6/4, width 32, four heads, MLP ratio 4; 600 epochs, $B=64$, mask ratio 0.5, 40 warmup epochs, base LR $10^{-3}$, minimum LR $10^{-5}$.\\
TabCSDI & Four diffusion layers, 64 channels, four heads; 50 quadratic-schedule steps; 250 epochs, LR $5\!\times\!10^{-4}$, $B=64$.\\
DiffPuter & Hidden width 1,024; ten EM rounds, 20 internal samples, 50 diffusion steps; at most 10,000 epochs per M-step, LR $10^{-4}$, $B=4096$, training-loss patience 500; warm start.\\
TabImpute & V2 pretrained weights; context cap 2,048 rows; test-row batch 256; 5,000 distribution bins; one point completion.\\
\midrule
\multicolumn{2}{@{}l}{\textit{Prediction specialists}}\\
Ridge / logistic regression & Ridge penalty $\alpha=1$ / inverse regularization $C=1$; logistic regression uses at most 1,000 solver iterations.\\
Random forest & 300 trees, minimum leaf size 1.\\
XGBoost & 500 estimators, depth 6, LR 0.1; row/column fractions 1; early-stopping patience 50.\\
LightGBM & 500 estimators, 31 leaves, LR 0.05, minimum child size 20; early-stopping patience 50.\\
CatBoost & 500 iterations, depth 6, LR 0.05, leaf $L_2=3$; early-stopping patience 50.\\
MLP & ReLU widths $[256,128]$, $L_2=10^{-4}$, initial LR $10^{-3}$; at most 500 iterations, automatic batch size, patience 20.\\
FT-Transformer / TabM & Three blocks; 100 epochs, $B=256$, patience 16. FT-Transformer: LR $10^{-4}$, WD $10^{-5}$. TabM: width 512, 32 ensemble members, dropout 0.1, LR $2\!\times\!10^{-3}$, WD 0.\\
CARD & 1,000 linear-schedule diffusion steps; backbone width 128; guidance widths $[100,50]$; guidance/diffusion budgets 1,000 epochs, LR $10^{-3}$; diffusion $B=32$, gradient clip 1; guidance patience 50.\\
NGBoost & Normal predictive distribution; 500 estimators, LR 0.01, full minibatches, early-stopping patience 50.\\
LightGBM-Quantile & 500 estimators, 31 leaves, LR 0.05; quantiles $\{.025,.05,.1,.25,.5,.75,.9,.95,.975\}$; early-stopping patience 50.\\
\midrule
\multicolumn{2}{@{}l}{\textit{Pretrained generalists}}\\
TabPFN-3 + unsupervised & V3 default classifier/regressor checkpoints; one estimator/permutation; context cap 50,000; generation temperature 1 and imputation temperature $10^{-9}$; cached fitting.\\
TabICLv2 & Classifier/regressor checkpoints dated 2026-02-12; eight estimators, batch 8, context cap 50,000; imputation temperature 1 and two iterations.\\
LimiX-2 & \texttt{LimiX-2.ckpt}; v2 no-retrieval classification, regression, and MVI configurations; context cap 50,000; prediction temperatures 0.9 (classification) and 1 (regression).\\
\end{longtable}
\endgroup

\paragraph{Foundation-model query interfaces.}
We evaluate TabPFN-3, TabICLv2, and LimiX-2 as pretrained inference systems
without dataset-specific weight fine-tuning~\citep{tabpfn3,tabiclv2,limix2}.
TabPFN-3 uses the \texttt{tabpfn-extensions} conditional sampler for
generation and imputation~\citep{tabpfn_extensions}, whereas TabICLv2 uses
\texttt{TabICLUnsupervised}. LimiX-2 combines native feature imputation with
sequential conditional generation. Prediction follows each adapter's declared
pretrained classification/regression route. A fixed method identity denotes
the backbone together with its query adapters; it does not imply a shared
output head. The query implementations are available in the
\href{https://github.com/limix-ldm-ai/OTBench}{public \ours{} repository}.

\paragraph{Resource allocation and environment.}
\label{app:implementation-environment}
The sharded launcher allocates one
independent method--dataset worker to each visible GPU and balances datasets
by $(n_{\rm train}+n_{\rm test})p$.  Its OpenMP, MKL, and OpenBLAS thread
defaults are two per worker.
We use NVIDIA GeForce RTX 5090 GPUs with 32\,GB of device memory per GPU.

%% file: sections/g_2_controlled_transfer.tex
\subsection{Controlled-transfer implementation}
\label{app:transfer-implementation}

For the underlying benchmark implementations, see the
\href{https://github.com/limix-ldm-ai/OTBench}{public \ours{} repository}.
The controlled study uses the five OpenML datasets in
Table~\ref{tab:transfer-datasets}, training seeds $\{0,1,2\}$, five
backbones, and seven objective arms $\{G,I,P,GI,GP,IP,GIP\}$.
Axis scores follow the seven-arm member-complete construction in
Appendix~\ref{app:causal-transfer}; availability is assessed within each
backbone rather than by freezing a common cross-backbone subset.

\begin{table}[htbp]
\centering
\caption{Controlled-transfer datasets and train--test splits.  Feature counts
exclude the target. C and R denote classification and regression.}
\label{tab:transfer-datasets}
\small
\begin{tabular}{lrrrrr}
\toprule
Dataset & OpenML ID & Type & Features & Train & Test\\
\midrule
JapaneseVowels & 976 & C & 14 & 7,968 & 1,993\\
MiamiHousing2016 & 43093 & R & 16 & 11,145 & 2,787\\
cleveland & 786 & C & 13 & 242 & 61\\
credit\_card\_clients\_default & 46919 & C & 23 & 24,000 & 6,000\\
website\_phishing & 46963 & C & 9 & 1,082 & 271\\
\bottomrule
\end{tabular}
\end{table}

\paragraph{Interventions and fitting.}
An enabled query type is sampled uniformly for each neural-training row,
duplicated forest exposure, or Gaussian-copula optimizer update. G scores
all variables; P observes the features and scores the target; I selects each
feature for reconstruction with probability $0.3$, enforces a nonempty query,
and hides the target without scoring it. Observed, non-query variables form the conditioning inputs.
The neural losses are masked to the queried coordinates; the forest objective
sets unqueried gradients and Hessians to zero. Configurations in
Table~\ref{tab:transfer-training} are fixed within each backbone across
datasets and arms. The three vendored neural pipelines use export protocol v6:
a deterministic 10\% internal holdout is drawn from the outer training
split using seed 0, and training-time validation uses this holdout. TabSyn
shares one objective-independent VAE per dataset--seed block across all seven arms.

\begin{table}[htbp]
\centering
\caption{Training configurations for the controlled-transfer study. Budgets are backbone-specific and matched across objective arms
within a dataset--seed block. LR denotes learning rate; WD denotes weight decay.}
\label{tab:transfer-training}
\small
\setlength{\tabcolsep}{4pt}
\begin{tabularx}{\textwidth}{@{}lXX@{}}
\toprule
Backbone & Architecture and loss & Optimizer and configured budget\\
\midrule
Gaussian Copula & Empirical marginals with at most 1,000 stored quantiles;
joint/conditional Gaussian likelihood for the selected query. &
Adam, LR $0.03$, batch 1,024, 2,000 updates; gradient norm clipped at 10.\\[2pt]
Forest-VP Joint & 50 diffusion times; 100 duplicated exposures; query-masked
multi-output squared-error trees. &
100 boosting rounds per time, depth 7, shrinkage $0.3$, histogram trees;
one job and one XGBoost thread per worker.\\[2pt]
TabDDPM & MLP widths $(1024,2048,2048,1024)$, time embedding 128, no dropout;
1,000 diffusion times, cosine noise schedule; masked numerical MSE plus
multinomial loss. &
AdamW, LR $10^{-3}$ with linear decay, WD $5\!\times\!10^{-4}$,
batch 4,096, 100,000 updates.\\[2pt]
TabDiff & Two layers, token width 4, one head, factor 32, time width 1,024;
learnable power-mean numerical and log-linear categorical schedules.
Discrete loss weight 1; continuous weight $1-e/8000$ at epoch $e$. &
AdamW, fixed LR $10^{-3}$, WD 0, batch 4,096, 8,000 epochs;
EMA $0.997$ and final EMA checkpoint; no separate target-only guidance model.\\[2pt]
TabSyn & Shared VAE: two layers, token width 4, one head, factor 32;
reconstruction plus $\beta$-KL. Latent EDM: width 1,024 and query-masked loss. &
Adam, LR $10^{-3}$, WD 0, batch 4,096; VAE 4,000 epochs;
latent diffusion 10,001 epochs at fixed LR, saving the final model.\\
\bottomrule
\end{tabularx}
\end{table}

TabSyn's shared VAE selects the checkpoint with the lowest categorical
cross-entropy on the internal validation set, uses ReduceLROnPlateau
(factor $0.95$, patience 10), and initializes $\beta=10^{-2}$ with decay factor
$0.7$ and threshold $10^{-5}$.
The causal branches bypass TabDiff's configured plateau LR scheduler and
TabSyn's diffusion plateau scheduler and early stopping. TabDiff's configured
\texttt{steps=8000} denotes epochs; TabDDPM's \texttt{steps=100000} denotes
optimizer updates.

\paragraph{Evaluation draws and analysis.}
Each checkpoint is reused for all three tasks. Generation requests as many
synthetic rows as the outer training split, with evaluation seed equal to the
training seed. Imputation uses MCAR, MAR, MNAR, and their mixed mechanism at
rates $\{0.1,0.3,0.5\}$, with the target hidden. Each mechanism--rate--training-seed
setting produces three completed-table draws under one fixed,
method-independent mask; the evaluation trial count is this draw count,
not an additional repetition factor. Selection, masking, and imputation seeds
equal the training seed. One checkpoint evaluation aggregates three
conditional target draws for Forest-VP Joint, TabDDPM, TabDiff, and TabSyn;
the \texttt{trials} and \texttt{n\_samples} arguments do not introduce an
additional nested repetition. Gaussian Copula uses native Monte Carlo
budgets of 100 draws for regression and 200 for classification.
Conditional sampling and training randomization are specified separately.
TabDiff and TabDDPM use one imputation resampling round; TabSyn uses 50 EDM
steps and 20 resampling rounds; Forest-VP uses RePaint with $r=5$ and $j=0.1$.

For Forest-VP Joint, TabDDPM, TabDiff, and TabSyn, the conditioning selector
follows the training arm: prediction uses clean observed features ($x_0$)
for $P,GP,IP,GIP$ and the $x_t$ conditioning path
for $G,I,GI$; imputation uses clean observed features for $I,GI,IP,GIP$
and the $x_t$ path for $G,P,GP$. Gaussian Copula does not use this selector:
it retains native Gaussian conditioning in all arms, including its Cleveland
imputation re-evaluation. Recipient-conditioned contrasts therefore hold the
relevant recipient protocol fixed. Generation inference is fixed across
arms. I/P self-inclusion, imputation-side interactions that toggle I, and
cross-arm Pareto profiles additionally compare the inference-policy switch
in the four selector-based adapters.

The updated analysis reuses the completed prediction evaluations and replaces
imputation outcomes with checkpoint-only re-evaluations. Of 270 planned
imputation cells, 264 completed all twelve mechanism--rate settings,
producing 3,168 metric artifacts. Six GPU-memory failures remain unavailable:
Forest-VP Joint GI/credit\_card\_clients\_default/seed 0;
IP/JapaneseVowels/seeds 1 and 2; IP/MiamiHousing2016/seed 0;
IP/cleveland/seed 0; and TabDDPM IP/credit\_card\_clients\_default/seed 1.
The old imputation outcomes for these six cells are excluded rather than
mixed with the updated protocol. Other unmodified imputation baselines are
retained. Re-evaluation does not update checkpoints or invoke fitting.

Uncertainty is estimated with 2,000 bootstrap draws and base random seed
20260924; 95\% intervals are the 2.5th and 97.5th
percentiles. Point estimates average available matched settings within each seed block,
then seeds within each backbone--dataset pair, and finally the pair means.
Pareto point aggregation instead weights seeds, then datasets within each
backbone, then backbones equally; its bootstrap resamples backbones and each
selected backbone's datasets and seeds, preserving the seven-arm pairing.
Wilcoxon diagnostics use the backbone--dataset means, with Holm adjustment
within each analysis family.

%% file: sections/h_related_work.tex
\section{Related Work}
\label{app:related-work}

\paragraph{Task-specific models and benchmarks.}
Generative tabular models span adversarial and variational approaches such
as CTGAN and TVAE~\citep{ctgan}, language-model generators such as
GReaT~\citep{great}, and diffusion methods such as TabDDPM, TabSyn, and
TabDiff~\citep{tabddpm,tabsyn,tabdiff}. Imputation ranges from chained
equations and random forests~\citep{mice,missforest} to learned conditional
models~\citep{vaeac,forestdiffusion}. Prediction combines tree ensembles
with neural and pretrained models~\citep{xgboost,catboost,tabpfn}.
Benchmark suites including Why Trees, TabZilla, TALENT, and TabArena broaden
systematic comparisons for prediction~\citep{whytrees,tabzilla,talent,talent_closer,tabarena}.
These task-wise evaluations establish performance within each problem,
while leaving open how capability rankings relate across generation,
imputation, and prediction.

\paragraph{Generalist and foundation models.}
Joint-distribution models can support conditional sampling, while
arbitrary-conditioning approaches accommodate varied observation
patterns~\citep{vaeac,forestdiffusion}. LimiX further motivates multiple
tabular queries within a shared modeling framework~\citep{limix}.
Recent tabular foundation models, including LimiX-2, TabICLv2, and
TabPFN-3~\citep{limix2,tabiclv2,tabpfn3}, extend the range of pretrained
systems considered in our benchmark. Conditional-query extensions also
expose generation and imputation routes~\citep{tabpfn_extensions}.
These developments motivate distinguishing interface coverage from
empirical quality under fixed model identities and task contracts.
\ours{} evaluates this distinction through capability-specific scores and
joint task profiles, rather than introducing a new model architecture.

\paragraph{Evaluation and task relations.}
Synthetic-data quality depends on which aspects of fidelity, dependence,
and downstream utility are measured, and metrics can favor conflicting
properties~\citep{structuredeval,shadesnull}. Multi-task learning also
studies cooperation and competition among training objectives when deciding
which tasks to learn together~\citep{standley2020tasks}.
\ours{} connects these issues through three complementary experiments.
Task-rank analysis studies capability co-occurrence on aligned support;
SAE and CAE measure objective additions and combinations under declared
training--inference protocols; and fixed-method profiles compare joint task
performance with specialist portfolios. This separates observational task
relationships, controlled objective effects, and practical model choice,
while preserving information-contract and metric distinctions.